\documentclass[runningheads]{llncs}

\usepackage{eccv}

\usepackage{eccvabbrv}

\usepackage{graphicx}
\usepackage{booktabs}

\usepackage[accsupp]{axessibility}  

\usepackage[pagebackref,breaklinks,colorlinks,citecolor=eccvblue]{hyperref}

\usepackage{orcidlink}

\usepackage{cite}
\usepackage{amsmath,amssymb,amsfonts}
\usepackage{algorithmic}
\usepackage{graphicx}
\usepackage{textcomp}

\usepackage{pifont}
\usepackage{tabularx}

\usepackage[table,dvipsnames]{xcolor} 
\usepackage{color}

\definecolor{Fraunhofergreen}{RGB}{23,156,125}
\definecolor{Fraunhofersteelblue}{RGB}{0,91,127}
\definecolor{Fraunhofersilvergrey}{RGB}{166,187,200}
\definecolor{Fraunhoferorange}{RGB}{245,130,32}

\definecolor{Fraunhofergraphit}{RGB}{28,63,82} 
\definecolor{Fraunhofersand}{RGB}{211,199,174} 
\definecolor{Fraunhoferpetrol}{RGB}{0,133,152}
\definecolor{Fraunhoferaqua}{RGB}{57,193,205}
\definecolor{Fraunhoferlime}{RGB}{178,210,53}

\definecolor{Fraunhoferblue}{RGB}{31,130,192} 
\definecolor{Fraunhoferred}{RGB}{187,0,86} 
\definecolor{Fraunhoferweinrot}{RGB}{124,21,77} 

\definecolor{Fraunhofersteelblue}{RGB}{0,91,127}
\definecolor{Fraunhofersilvergrey}{RGB}{166,187,200}

\definecolor{KITred}{RGB}{160,30,40}
\definecolor{KITblue}{RGB}{ 70,100,170} 
\definecolor{KITblue70}{RGB}{125,146,195} 
\definecolor{KITblue50}{RGB}{162,177,212} 
\definecolor{KITblue30}{RGB}{199,208,229} 
\definecolor{KITblue15}{RGB}{227,232,242} 

\definecolor{Set2-01}{RGB}{102, 194, 165}
\definecolor{Set2-02}{RGB}{252, 141,  98}
\definecolor{Set2-03}{RGB}{141, 160, 203}
\definecolor{Set2-04}{RGB}{231, 138, 195}

\definecolor{Set3-01}{RGB}{141, 211, 199}
\definecolor{Set3-02}{RGB}{255, 255, 179}
\definecolor{Set3-03}{RGB}{251, 128, 114}
\definecolor{Set3-04}{RGB}{ 128, 177, 211}

\definecolor{Set3-05}{RGB}{179, 222, 105}
\definecolor{Set3-06}{RGB}{252, 205, 229}
\definecolor{Set3-07}{RGB}{188, 128, 189}
\definecolor{Set3-08}{RGB}{ 204, 235, 197}

\definecolor{IBM-Blue}{RGB}{ 100, 143, 255}
\definecolor{IBM-Yellow}{RGB}{ 255, 176, 0}
\definecolor{IBM-Red}{RGB}{ 220, 38, 127}

\definecolor{coolred}{rgb}{0.8, 0.0, 0.0}
\definecolor{truepositives}{RGB}{68, 119, 170}
\definecolor{localizationerrors}{RGB}{238, 102, 119}
\definecolor{duplicates}{RGB}{34, 136, 51}
\definecolor{alldetections}{RGB}{204, 187, 68}
\definecolor{topkdetections}{RGB}{102, 204, 238}
\definecolor{falsepositives}{RGB}{170, 51, 119}

\colorlet{colorFst}{Green!20}       %
\colorlet{colorSnd}{SpringGreen!45} %
\colorlet{colorTrd}{Yellow!30}      %

\usepackage{tikz}
\usetikzlibrary{%
	automata,%
	arrows,%
	backgrounds,%
	calc,%
	chains,%
	decorations,%
	decorations.markings,%
	decorations.pathmorphing,%
	external,%
	fadings,%
	fit,%
	matrix,%
	positioning,%
	scopes,%
	shadows,%
	shapes.geometric,%
	shapes.misc,%
	shapes.multipart,%
	through,%
	mindmap, 
	backgrounds, 
	topaths}

\usetikzlibrary{arrows.meta}
\usetikzlibrary{decorations.pathreplacing}
\tikzset{
	pics/square/.default={1},
	pics/square/.style = {
		code = {
			\draw[pic actions, ultra thick] (0,0) rectangle (#1,#1);
		}
	}
}

\tikzset{
	icon/.style={
		draw, rounded corners=2pt, minimum width=9mm, minimum height=6mm,
		font=\footnotesize\sffamily, align=center
	},
	src/.style={icon, fill=blue!10},
	eff/.style={icon, fill=green!10},
	grp/.style={icon, fill=orange!20},
}

\usepackage{amsmath}

\usepackage{indentfirst}
\usepackage{amsfonts}
\usepackage{amssymb}

\usepackage{comment}
\usepackage{enumitem}
\usepackage[font=footnotesize,skip=3pt,subrefformat=parens]{subcaption}

\definecolor{darkspringgreen}{rgb}{0.09, 0.45, 0.27}
\definecolor{bistre}{rgb}{0.24, 0.17, 0.12}
\definecolor{auburn}{rgb}{0.43, 0.21, 0.1}
\def\projectpage/{}
\usepackage{pdflscape}
\usepackage{threeparttable}
\usepackage{multirow}
\usepackage{booktabs}
\usepackage{csquotes}
\definecolor{britishracinggreen}{rgb}{0.0, 0.26, 0.15}

\usepackage{textpos}

\usepackage{adjustbox}
\usepackage[]{todonotes}
\usepackage{arydshln} 

\usepackage{utfsym}

\usepackage[linesnumbered,ruled,vlined]{algorithm2e}

\newcommand{\smallsec}[1]{\vspace{0.75em}\noindent\textbf{#1}\;}

\newcommand{\subw}{0.15\textwidth} 

\newcommand{\degfunc}[1]{{'#1'}}
\newcommand{\degeff}[1]{\emph{#1}}

\newcommand{\repo}[2]{\href{#2}{\texttt{#1}}}

\usepackage{siunitx}

\begin{document}

\title{A Causally Grounded Taxonomy for \\ Image Degradation Robustness Evaluation}

\titlerunning{Taxonomy for Degradation Robustness Evaluation}

\author{
	Stefan Becker\inst{1}\orcidlink{0000-0001-7367-2519} \and
	Simon Weiss\inst{1}\orcidlink{0009-0008-8171-9397} \and
	Wolfgang H\"ubner\inst{1}\orcidlink{0000-0001-5634-6324} \and
	Michael Arens\inst{1}\orcidlink{0000-0002-7857-0332}
}

\authorrunning{S.~Becker et al.}

\institute{Fraunhofer IOSB, Ettlingen, Germany\\
	Fraunhofer Center for Machine Learning \\
	\email{\{firstname.lastname\}@iosb.fraunhofer.de}
}

\maketitle

\begin{abstract}
Image degradations arise throughout acquisition, processing, and transmission, altering both visual appearance and downstream machine-perception performance. They are studied across synthetic corruption benchmarks, perceptual image quality assessment (IQA), low-level vision, and physically grounded imaging-system analysis. However, these communities rely on incompatible grouping schemes and backend-specific severity definitions, making results difficult to compare across datasets, degradation models, and tasks.

We propose a causally grounded framework for organizing and interpreting image degradations. Rather than introducing new degradations or redefining existing benchmarks, we provide an interpretive representation and measurement layer that makes implicit assumptions explicit. Each degradation is characterized by two orthogonal axes: its dominant causal source in the imaging pipeline and its dominant perceptual effect. This dual-axis abstraction yields a compact taxonomy spanning algorithmic corruptions, perceptual distortions, low-level imaging artifacts, and physically motivated camera or system failures.

To make severity semantics observable without modifying existing implementations, we quantify each native severity level using full-reference image quality metrics: PSNR, SSIM, and LPIPS. This separates causal origin, perceptual appearance, measured distortion strength, and downstream task impact while preserving native parameterizations. We demonstrate the framework through COCO-Degradation, a taxonomy-aligned benchmark for evaluating object detector robustness under diverse low-level imaging conditions.

By connecting causal source, perceptual effect, and measured severity, the proposed framework supports more interpretable and reproducible evaluation of how low-level visual degradations affect machine perception.

Source code is available at \href{https://github.com/beckerio/imdeg}{Code}.
\keywords{image degradation \and robustness evaluation  \and object detection \and corruption benchmarks \and image quality assessment \and taxonomy \and severity calibration}
\end{abstract}

\section{Introduction}
\label{sec:intro}

Object detection systems are increasingly deployed under imaging conditions that deviate substantially from the clean, well-exposed data typically used for training and evaluation. Environmental effects such as fog or rain, sensor and optical limitations, image signal processing (ISP) artifacts, and downstream transmission or system faults all alter image appearance and can severely affect detection performance. As modern detectors approach saturation on standard benchmarks, understanding and characterizing robustness under degraded imaging conditions has become a central concern. We view this problem as a low-level vision evaluation problem for machine perception. Image formation, restoration, enhancement, compression, and quality assessment all operate on low-level visual signals, yet their effects are often evaluated using heterogeneous degradation definitions and incompatible severity conventions. This makes it difficult to determine whether performance changes arise from the causal source of a degradation, its perceptual manifestation, its measured distortion strength, or the downstream model’s sensitivity to that degradation.

Robustness to image degradation has been studied across three largely separate research traditions. Synthetic corruption benchmarks, most prominently ImageNet-C~\cite{Hendrycks_ICLR_2019} and COCO-C~\cite{Michaelis_NeurIPSW_2019}, evaluate models under algorithmic stress tests designed to disrupt semantic cues. In parallel, the image quality assessment (IQA) community has developed datasets and metrics that model perceptual distortions with respect to human visual quality. Finally, real-camera and imaging-system studies investigate degradations arising from sensor defects, optical failures, ISP malfunctions, or board-level system faults. Although these lines of work address closely related phenomena, they rely on different definitions, grouping schemes, and severity parameterizations. As a consequence, robustness results are difficult to compare across benchmarks, and the relationship between performance degradation and the underlying imaging process often remains unclear.

Two limitations are particularly restrictive. First, degradations are typically grouped in ad hoc or domain-specific ways, which obscures conceptual relationships between algorithmic corruptions, perceptual distortions, and physically grounded imaging artifacts. Second, severity levels are usually defined implicitly through backend-specific parameters. A severity index of ``3'' may correspond to fundamentally different distortion magnitudes depending on the degradation type or its source, even within the same benchmark. This lack of semantic alignment complicates interpretation and weakens cross-benchmark analysis.

\begin{figure*}[t]
	\centering
	\begin{subfigure}{0.95\textwidth}
		\centering
		\includegraphics[width=\textwidth]{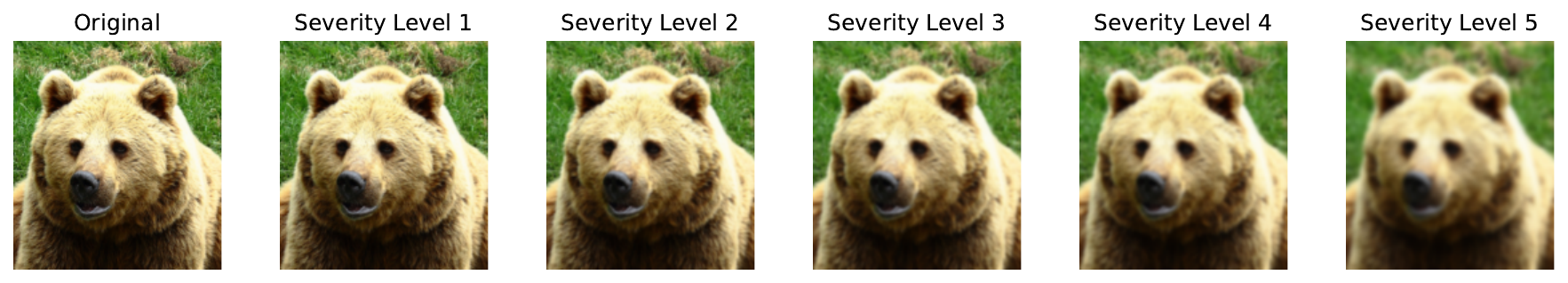}
		\caption{\degfunc{Gaussian Blur} as defined in ImageNet-C~\cite{Hendrycks_ICLR_2019} at native severities 1--5.}
	\end{subfigure}
	
	\vspace{0.8em}
	
	\begin{subfigure}{0.95\textwidth}
		\centering
		\includegraphics[width=\textwidth]{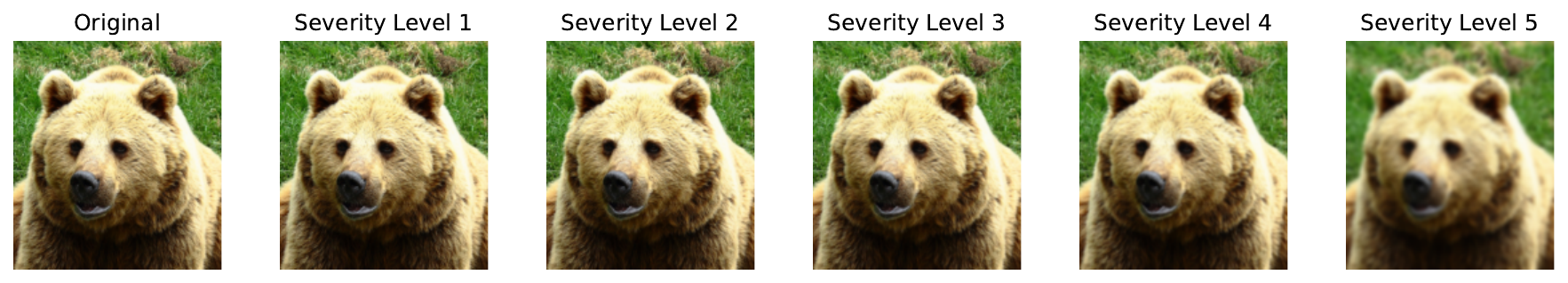}
		\caption{\degfunc{Gaussian Blur} as implemented in the IQA-oriented degradation set used by ARNIQA~\cite{Agnolucci_WACV_2024} at native severities 1--5.}
	\end{subfigure}
	\caption{
		Backend ambiguity for identically named degradations. Two implementations labeled \degfunc{Gaussian Blur} are applied to the same COCO image~\cite{Lin_ECCV_2014}: a corruption-benchmark operator from ImageNet-C~\cite{Hendrycks_ICLR_2019} (top) and an IQA-oriented distortion used by ARNIQA~\cite{Agnolucci_WACV_2024} (bottom). While both produce the perceptual effect of \degeff{blur}, they rely on different modeling assumptions and native severity schedules. Consequently, the ordinal severity indices 1--5 are not directly comparable across backends.
	}
	\label{fig:coco_c_examples}
\end{figure*}

Grouping degradations only by perceptual appearance is often insufficient for meaningful analysis. Many degradations produce similar visual effects while originating from fundamentally different stages of the imaging pipeline. These differences in causal origin are not merely terminological: they influence how degradations arise, how their severity scales with physical or algorithmic parameters, how they interact with other processing stages, and how likely they are to occur in real systems.

Blur is a representative example. From a purely perceptual perspective, defocus blur, motion blur, and ISP-induced smoothing may all appear as visually similar forms of blur. Yet their causal origins differ substantially. Defocus blur is governed by optical properties such as aperture, depth, and focus control; motion blur depends on relative motion during exposure and shutter timing; and ISP-induced smoothing is algorithmic and often spatially uniform, reflecting the processing pipeline rather than scene geometry. Although these effects share a common perceptual label, they differ in spatial structure, temporal behavior, and correlation with other imaging artifacts. A detector that is robust to the operator \degfunc{Gaussian Blur} as defined in a synthetic corruption benchmark may therefore still fail under motion-induced blur or optical defocus, despite comparable visual appearance.

Figure~\ref{fig:coco_c_examples} illustrates this issue for \degfunc{Gaussian Blur}. Although identically named, the ImageNet-C implementation~\cite{Hendrycks_ICLR_2019} and the IQA-oriented variant used by~\cite{Agnolucci_WACV_2024} instantiate distinct parameterizations that reflect different objectives---algorithmic stress testing versus perceptual scaling. As a result, native severity indices are neither visually nor quantitatively aligned. Similar inconsistencies arise across many degradation families, including noise, compression, and color shifts, obscuring how nominal severity relates to actual distortion strength and downstream detection performance.

In this work, we introduce a \emph{causally grounded framework} for organizing and analyzing image degradations. We describe each degradation along two orthogonal axes: its dominant \emph{causal source} within the imaging pipeline---environment, sensor/optics, ISP/renderer/codec, or transfer/system---and its dominant \emph{perceptual effect}, such as noise, blur, compression artifacts, or illumination changes. This dual-axis abstraction yields a compact taxonomy that unifies synthetic corruptions, IQA distortions, and real-camera or system-level artifacts within a common conceptual space.

Rather than redefining native severity schedules, we further introduce a lightweight \emph{severity measurement layer}. For each degradation and each native severity level defined by a backend, we explicitly quantify degradation strength using full-reference image quality metrics, namely PSNR, SSIM \cite{Wang_TIP_2004}, and LPIPS \cite{Zhang_CVPR_2018}. This makes severity semantics observable and comparable across degradation sources without modifying the original implementations. Importantly, the proposed layer is descriptive rather than prescriptive: it exposes how native severity indices relate to measurable distortion magnitude, without imposing a single artificial severity scale.

We demonstrate the practical utility of this framework through \emph{COCO-Degradation}, a taxonomy-aligned benchmark for evaluating object detector robustness under diverse imaging conditions. COCO-Degradation integrates degradation functions from synthetic corruption benchmarks, IQA-motivated distortions, and real-camera or system-level failure models. All degradations are annotated according to the proposed taxonomy, and their native severity levels are accompanied by measured degradation strengths, enabling more principled robustness analysis while preserving compatibility with established benchmarks. 

Our goal is interpretability and structural clarification rather than exhaustive empirical coverage: the framework is intended to make degradation assumptions, severity semantics, and robustness results easier to compare, not to replace existing task-specific benchmarks. In particular, \emph{COCO-Degradation} is a taxonomy-aligned case study built from existing degradation backends rather than a claim of a new corruption family.

Throughout the paper, \emph{framework} refers to the combination of the dual-axis taxonomy and the severity measurement layer, while \emph{representation} refers to the resulting cause/effect annotations together with measured distortion strength. We use \emph{causal} to denote degradation provenance within the imaging pipeline, not causal-effect identification, intervention analysis, counterfactual reasoning, or a structural causal model.

To support reproducibility and reuse, we release an open-source implementation, \texttt{imdeg}, which provides degradation functions, taxonomy mappings, severity measurement tools, and benchmark-generation utilities in a unified library.

In summary, our contributions are:
\begin{itemize}[leftmargin=*]
	\item A causally grounded, dual-axis taxonomy that provides a consistent organizational structure for image degradations across synthetic corruption benchmarks, IQA datasets, and real-camera or system-level studies.
	\item A lightweight severity measurement layer that makes native severity semantics explicit and supports cross-source interpretation without modifying existing degradation definitions.
	\item COCO-Degradation and the \texttt{imdeg} library, providing a practical benchmark and tooling framework for structured machine-perception robustness evaluation under heterogeneous imaging conditions.
\end{itemize}
\section{Related Work}
\label{sec:relatedwork}

Research on image degradations spans several largely independent communities. In this section, we position our work with respect to three major lines: synthetic corruption benchmarks developed for robustness evaluation, perceptual image quality assessment (IQA), and physically grounded imaging-system or real-camera degradation studies. While all three investigate degraded imagery, they differ substantially in how degradations are defined, grouped, and parameterized, which complicates comparison and interpretation across datasets and tasks.

Synthetic corruption benchmarks evaluate model robustness under controlled algorithmic image perturbations. ImageNet-C established the common benchmark paradigm with five native severities per degradation~\cite{Hendrycks_ICLR_2019}, and subsequent work transferred this setting to object detection on COCO and related datasets~\cite{Michaelis_NeurIPSW_2019,Oksuz_CVPR_2023}. These benchmarks are effective stress tests, but their degradations are typically grouped by benchmark design rather than by a causal view of image formation. In addition, benchmark variants often differ not only in the degradation sets they use, but also in their dataset construction protocols, which further complicates comparison across studies.

Image quality assessment studies degradations from the perspective of perceived image quality. Classical full-reference metrics such as SSIM~\cite{Wang_TIP_2004} and learned perceptual metrics such as LPIPS~\cite{Zhang_CVPR_2018} are evaluated on datasets whose severity schedules are designed for perceptual separability rather than for robustness stress testing. Recent IQA models and datasets continue this line, including ARNIQA and KADID-based distortions~\cite{Agnolucci_WACV_2024,Lin_QoMEX_2019}. While many IQA distortions overlap with corruption benchmarks at the perceptual level, their severity semantics differ, since equal severity steps are intended to reflect perceptual change rather than fixed backend-parameter increments. IQA also includes no-reference approaches that estimate image quality without a pristine reference. This setting is complementary to the present benchmark construction, where degradations are generated from clean images and paired references are therefore available for direct distortion measurement.

Recent low-level vision work increasingly evaluates restoration, enhancement, compression, and degradation-aware preprocessing not only by perceptual quality, but also by their effect on downstream machine-perception tasks. This trend highlights that low-level image quality and task robustness are coupled but not interchangeable: improving perceptual quality does not necessarily imply improved detection or recognition, and equal nominal degradation levels may affect models differently. Our work follows this evaluation-oriented perspective, focusing on how heterogeneous degradation definitions and severity conventions can be interpreted consistently across low-level vision, IQA, and robustness evaluation.

Physically grounded imaging-system studies model degradations through environment, optics, sensor behavior, ISP, or board-level faults. Recent robustness work has begun to incorporate such real-camera failures into detection benchmarks~\cite{Liu_IJCV_2024}. These approaches offer stronger causal grounding, but usually do not provide a common organizational layer that connects them to synthetic corruptions and IQA distortions, nor a severity interpretation layer that aligns their effects with existing benchmark conventions.

Across these communities, degradations are defined with different intents, grouped using different schemes, and parameterized by severity schedules whose semantics are rarely aligned. Our goal is not to replace these benchmarks, but to provide a common interpretive layer that jointly accounts for causal origin, perceptual effect, and measurable severity.

Additional robustness-evaluation context, including detector families, corruption benchmark variants, weather datasets, and cross-detector results, is provided in the supplementary material.
\section{Causally Grounded Taxonomy and Severity Measurement}
\label{sec:taxonomy_severity}

To enable consistent reasoning about image degradations across synthetic corruption benchmarks, perceptual IQA datasets, and real-camera or system-level studies, we describe each degradation along two orthogonal axes: its dominant \emph{causal source} within the imaging pipeline and its dominant \emph{perceptual effect} in the final image. This separation captures both \emph{where} a degradation originates and \emph{how} it manifests visually, providing a compact representation that generalizes across existing benchmark traditions.

We treat each degradation as a backend-defined operator applied to a clean reference image with native parameters and a native severity index. Each resulting degradation instance is described by a dominant causal source and a dominant perceptual effect, while its realized strength is measured separately using full-reference measures. This separates what the degradation is, where it originates, how it appears, and how strong it is under a particular implementation.

\begin{figure*}[t]
	\centering
	\resizebox{0.8\textwidth}{!}{%
		\includegraphics[width=\linewidth]{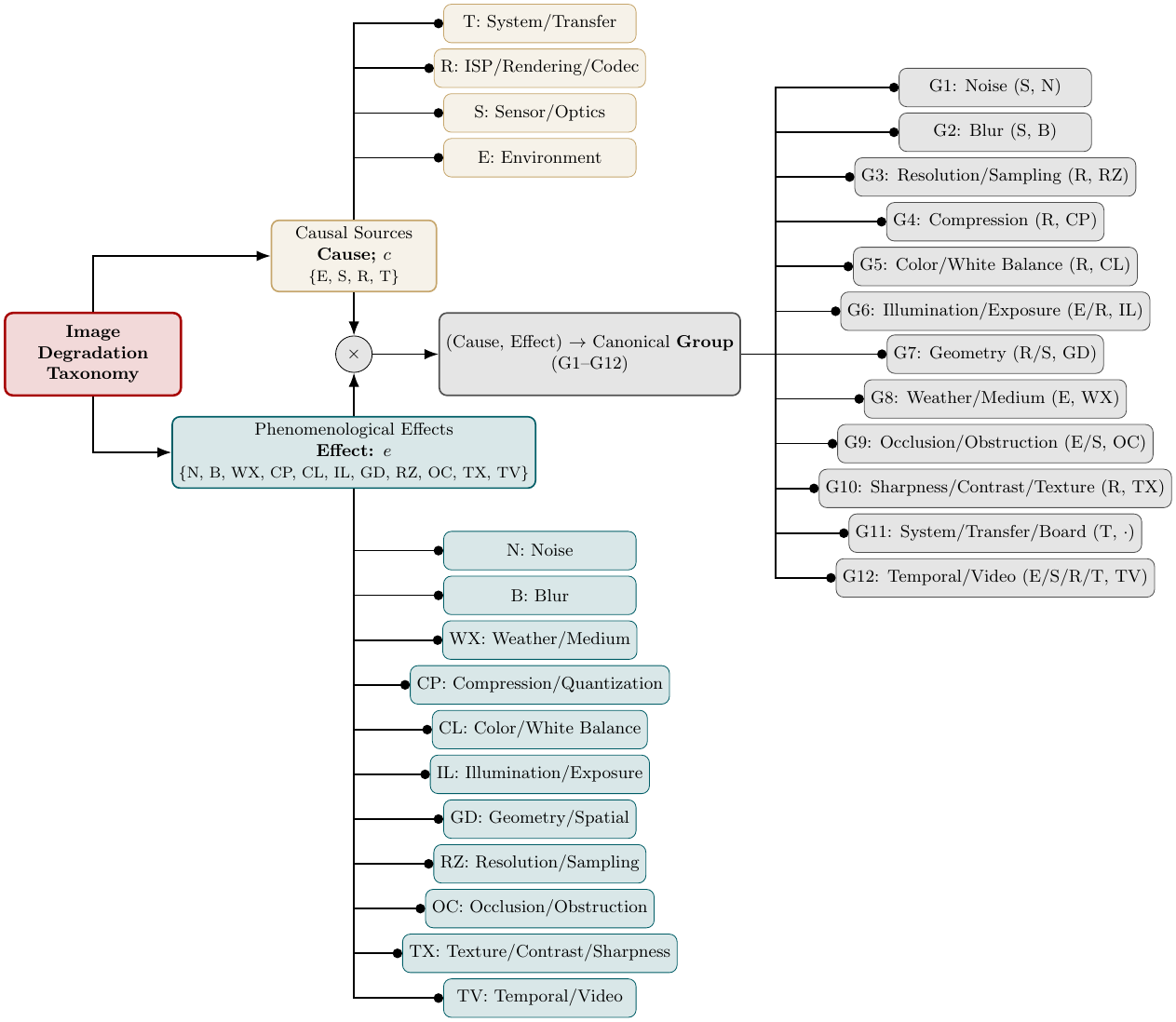}
	}
	\caption{
		Schematic overview of the proposed image degradation taxonomy.
		Degradations are described by a causal source (\emph{cause}: $c$) in the imaging pipeline and a perceptual manifestation (\emph{effect}: $e$). Their combination yields twelve canonical degradation groups (G1--G12), providing a compact organizational layer across corruption benchmarks, IQA distortions, and real-camera or system-level artifacts.
	}
	\label{fig:taxonomy_overview}
\end{figure*}

\textbf{Cause axis.}
The causal source variable identifies the primary stage of the imaging pipeline at which a degradation arises. Following the causal image-formation perspective of Drenkow et al.~\cite{Drenkow_arxiv_2022}, we adopt a compact four-stage decomposition of the imaging pipeline into
\emph{Environment (E)}, \emph{Sensor/Optics (S)}, \emph{ISP/Renderer/Codec (R)}, and \emph{Transfer/System (T)}.
This decomposition is not introduced as a new physical image-formation model; rather, we use it as an interpretive abstraction for robustness evaluation. It separates, for example, weather-induced visibility loss from sensor-originated noise, ISP-induced color errors, and downstream board-level or transfer failures. This positioning makes explicit that the causal-source axis builds on prior causal views of imaging pipelines, while our contribution lies in combining this axis with an orthogonal perceptual-effect axis and a severity measurement layer.

\textbf{Effect axis.}
Orthogonal to cause, the perceptual effect variable describes the dominant visual manifestation in the final image. We use the categories
\emph{Noise (N)}, \emph{Blur (B)}, \emph{Weather/Medium (WX)}, \emph{Compression/Quantization (CP)}, \emph{Color/White Balance (CL)}, \emph{Illumination/Exposure (IL)}, \emph{Geometry/Spatial (GD)}, \emph{Resolution/Sampling (RZ)}, \emph{Occlusion/Obstruction (OC)}, \emph{Texture/\-Sharpness/\-Contrast (TX)}, and \emph{Temporal/Video (TV)}.
These categories are intentionally perceptual rather than algorithmic: distinct implementations may share the same visual effect while differing substantially in physical origin and parameterization.

\textbf{Taxonomy assignment.}
Assignments are derived from the documented mechanism and semantics of each backend operator: we first identify the dominant imaging-pipeline stage responsible for the degradation and then its dominant visual manifestation in the output. Complete per-backend mappings are reported in the supplementary material. Ambiguous or mixed cases are therefore treated as an explicit annotation convention rather than as a claim of a unique ontology.

Combining cause and effect yields twelve canonical degradation groups, illustrated in Figure~\ref{fig:taxonomy_overview}. The representation is deliberately compact. It does not attempt to model all compound degradations explicitly; instead, it provides an interpretable abstraction layer that is broad enough to cover the major degradation families used in robustness benchmarking, IQA, and imaging-system studies.

The value of this separation is that perceptually similar degradations need not be treated as conceptually identical. Blur is a representative example: optical defocus, motion blur, and ISP-induced smoothing may all appear as \degeff{blur}, yet they differ in origin, spatial structure, severity scaling, and system relevance. Organizing degradations only by appearance therefore obscures information that matters when interpreting robustness results.

A second source of inconsistency is \emph{severity}. Most benchmarks expose severity as a small ordinal scale, but the semantics of that scale are tied to backend-specific parameter choices such as blur kernel size, noise variance, or compression quality. As a result, identical severity indices are often not comparable across degradation types or sources.

Rather than redefining native severity schedules, we introduce a lightweight \emph{severity measurement layer}. Let $x^{(n)}$ denote a clean reference image and $x^{(n)}_{deg;D,l}$ its degraded counterpart produced by degradation $\degfunc{deg}$ from backend $D$ at native severity level $l$. We estimate average degradation strength as
\begin{equation}
	d_{D,deg,l}
	=
	\frac{1}{N}
	\sum_{n=1}^{N}
	d\!\left(x^{(n)},x^{(n)}_{deg;D,l}\right),
\end{equation}
where $d(\cdot,\cdot)$ is a full-reference measure.

In this work, we report degradation strength using \emph{PSNR}, \emph{SSIM}, and \emph{LPIPS} as a compact complementary set spanning pixel-level fidelity, structural change, and learned perceptual difference. This choice is not intended to be exhaustive: the measurement layer is metric-extensible and can incorporate additional full-reference measures without changing the taxonomy or the native backend definitions. We treat the reported measures as \emph{descriptive} rather than predictive, that is, as observable characterizations of distortion strength rather than proxies for semantic task difficulty.

\begin{figure}[t]
	\captionsetup[subfigure]{justification=centering}
	\centering
	\begin{subfigure}{0.48\linewidth}
		\centering
		\includegraphics[width=\linewidth]{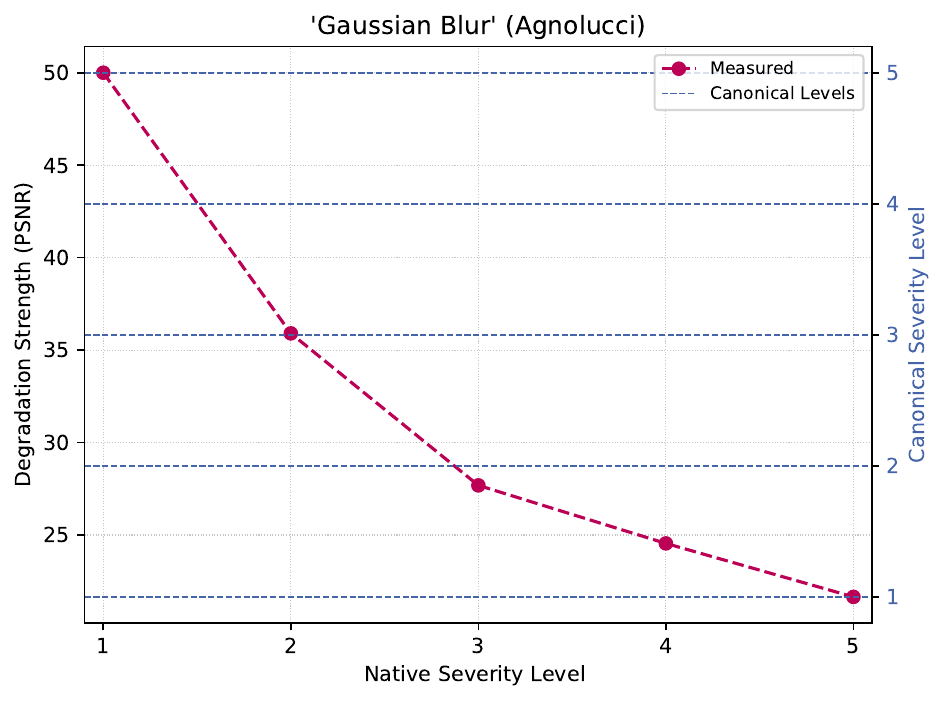}
		\caption*{PSNR}
	\end{subfigure}
	\hfill
	\begin{subfigure}{0.48\linewidth}
		\centering
		\includegraphics[width=\linewidth]{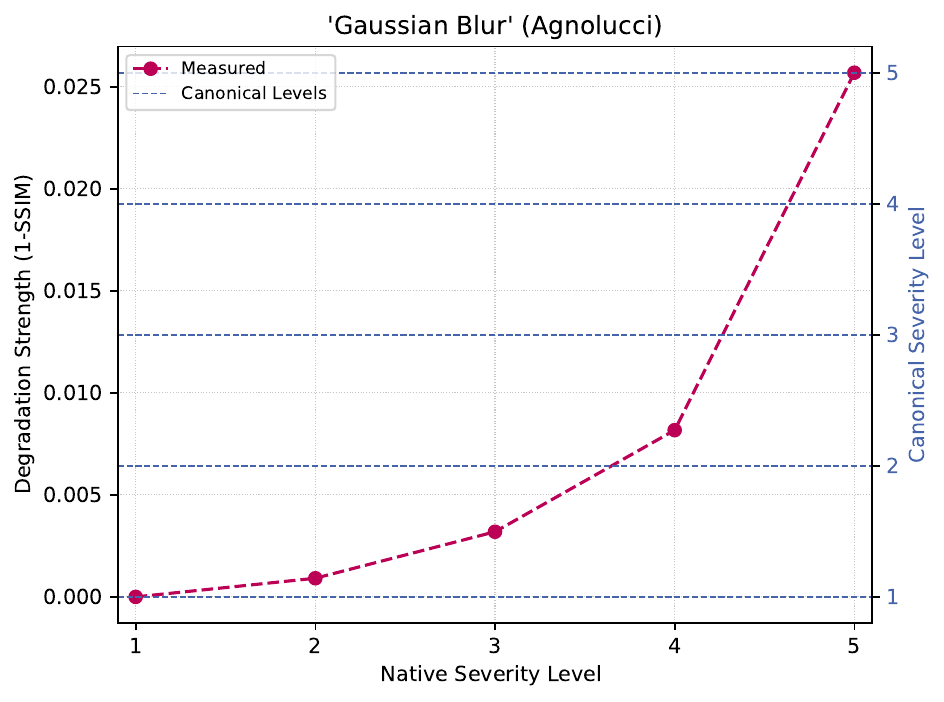}
		\caption*{$(1-\mathrm{SSIM})$}
	\end{subfigure}
	\caption{
		Severity calibration for \degfunc{Gaussian Blur} from~\cite{Agnolucci_WACV_2024}. Measured degradation strengths at native severity levels define a severity--strength curve, from which metric-space reference levels can be derived for analysis.
	}
	\label{fig:can_sev_level}
\end{figure}

For each degradation, this produces a severity--strength curve that makes native severity semantics explicit. Figure~\ref{fig:can_sev_level} shows this for \degfunc{Gaussian Blur}. The calibration is performed in metric space rather than in backend parameter space, allowing native severities from different implementations to be analyzed against a shared set of distortion measurements without modifying the original degradation functions.

Table~\ref{tab:degradation_strengths_all_metrics_hendrycks} provides a concrete view of this measurement layer for the ImageNet-C style degradation family. Two observations are immediate. First, native severity levels are not uniformly spaced in distortion space, even within a single backend. Second, some degradations exhibit weakly monotonic or irregular metric behavior, indicating that ordinal severity alone is insufficient for principled interpretation. The proposed measurement layer therefore does not replace native severity definitions; it makes their semantics observable while preserving compatibility with established benchmark implementations.

In the remainder of the paper, we use this taxonomy and measurement layer as an analysis framework for \emph{COCO-Degradation}. All degradations remain evaluated at their native severity levels, while PSNR, SSIM, and LPIPS provide an explicit characterization of how strong those native levels actually are.
\section{COCO-Degradation}
\label{sec:coco-degradation}
We instantiate the proposed framework in \emph{COCO-Degradation}, a low-level degradation evaluation case study for object detection on COCO \texttt{val2017}~\cite{Lin_ECCV_2014}. The benchmark applies single degradations to all 5{,}000 validation images using five native severity levels per operator and reports detector performance relative to the pristine baseline. This preserves the standard benchmark setting while isolating the effect of degradation severity.

\textbf{Backends.}
COCO-Degradation integrates three complementary degradation families:
(i) 19 ImageNet-C style operators used in corruption benchmarking~\cite{Hendrycks_ICLR_2019,Michaelis_NeurIPSW_2019,Oksuz_CVPR_2023},
(ii) 24 IQA-motivated distortions used by ARNIQA and derived from KADID-10k~\cite{Agnolucci_WACV_2024,Lin_QoMEX_2019}, and
(iii) 16 real-camera or system-level failures motivated by object detection under real-world corruptions~\cite{Liu_IJCV_2024}.
Each degradation is annotated by the proposed taxonomy, and each native severity level is augmented with measured PSNR, SSIM, and LPIPS values.

\textbf{Evaluation.}
In the main paper, we report results for YOLOv10-m~\cite{Wang_NeurIPS_2024} using standard COCO detection metrics, with emphasis on mAP@0.5:0.95. Complementary experiments with YOLOv9, YOLO11, and RT-DETR are reported in the supplementary material under the same COCO-Degradation protocol to assess whether the observed degradation trends persist across representative one-stage and transformer-based detector families. These experiments provide a cross-architecture consistency check rather than evidence of task generality.

\begin{figure*}[t]
	\captionsetup[subfigure]{justification=centering}
	\centering
	\begin{subfigure}{0.32\textwidth}
		\centering
		\includegraphics[width=\textwidth]{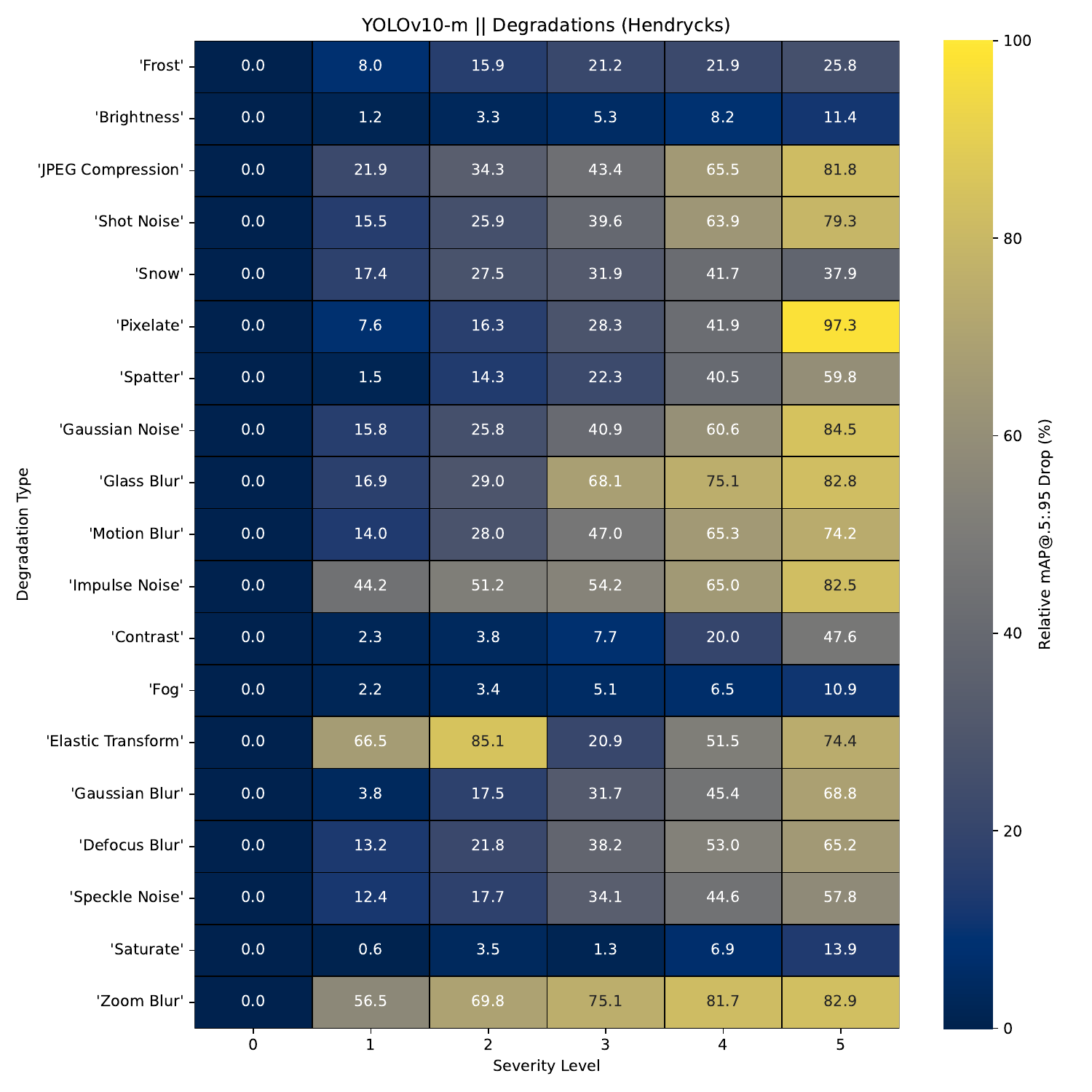}
		\caption*{Corruption benchmarks\\Degradations from~\cite{Hendrycks_ICLR_2019}}
	\end{subfigure}
	\hfill
	\begin{subfigure}{0.32\textwidth}
		\centering
		\includegraphics[width=\textwidth]{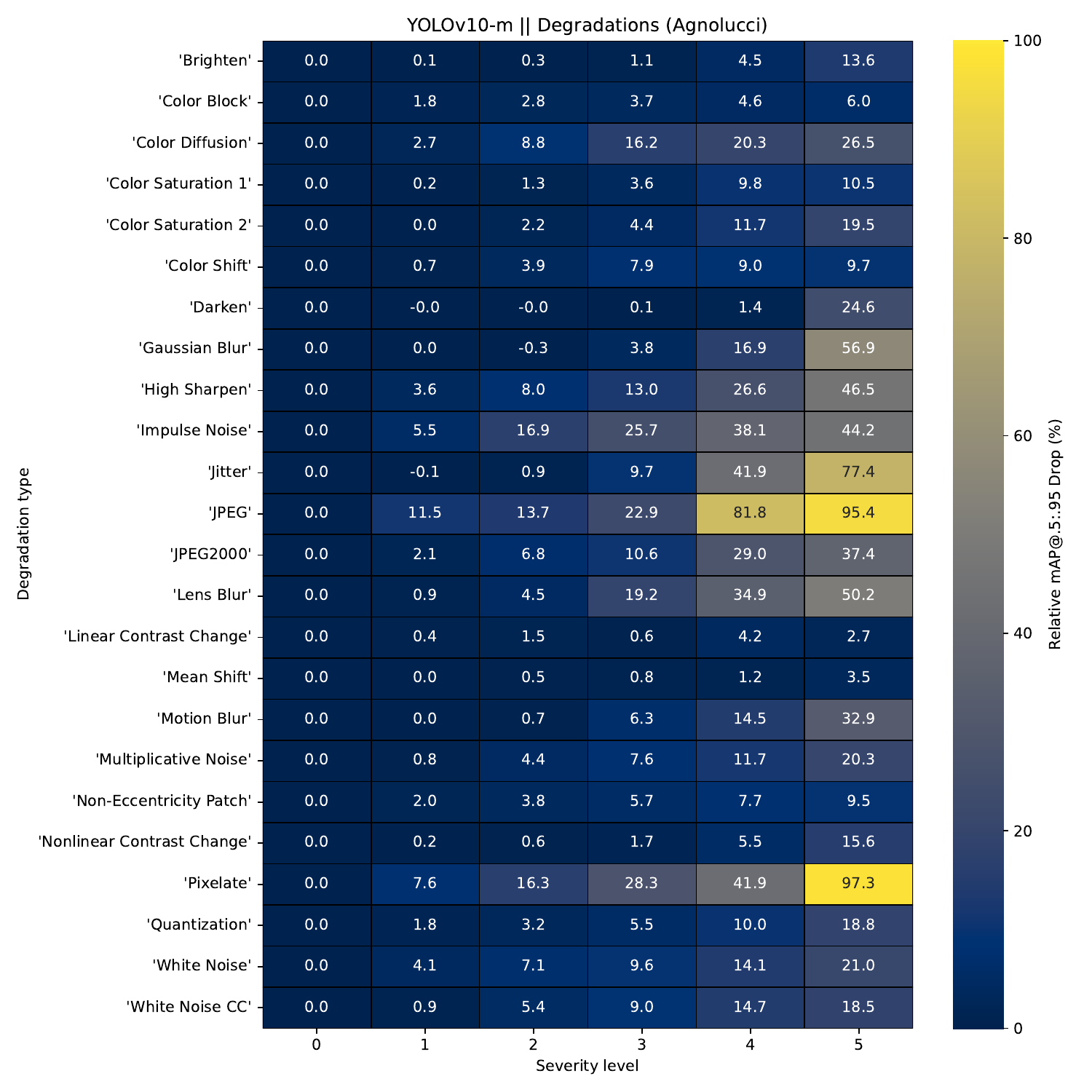}
		\caption*{IQA distortions\\Degradations from~\cite{Agnolucci_WACV_2024}}
	\end{subfigure}
	\hfill
	\begin{subfigure}{0.32\textwidth}
		\centering
		\includegraphics[width=\textwidth]{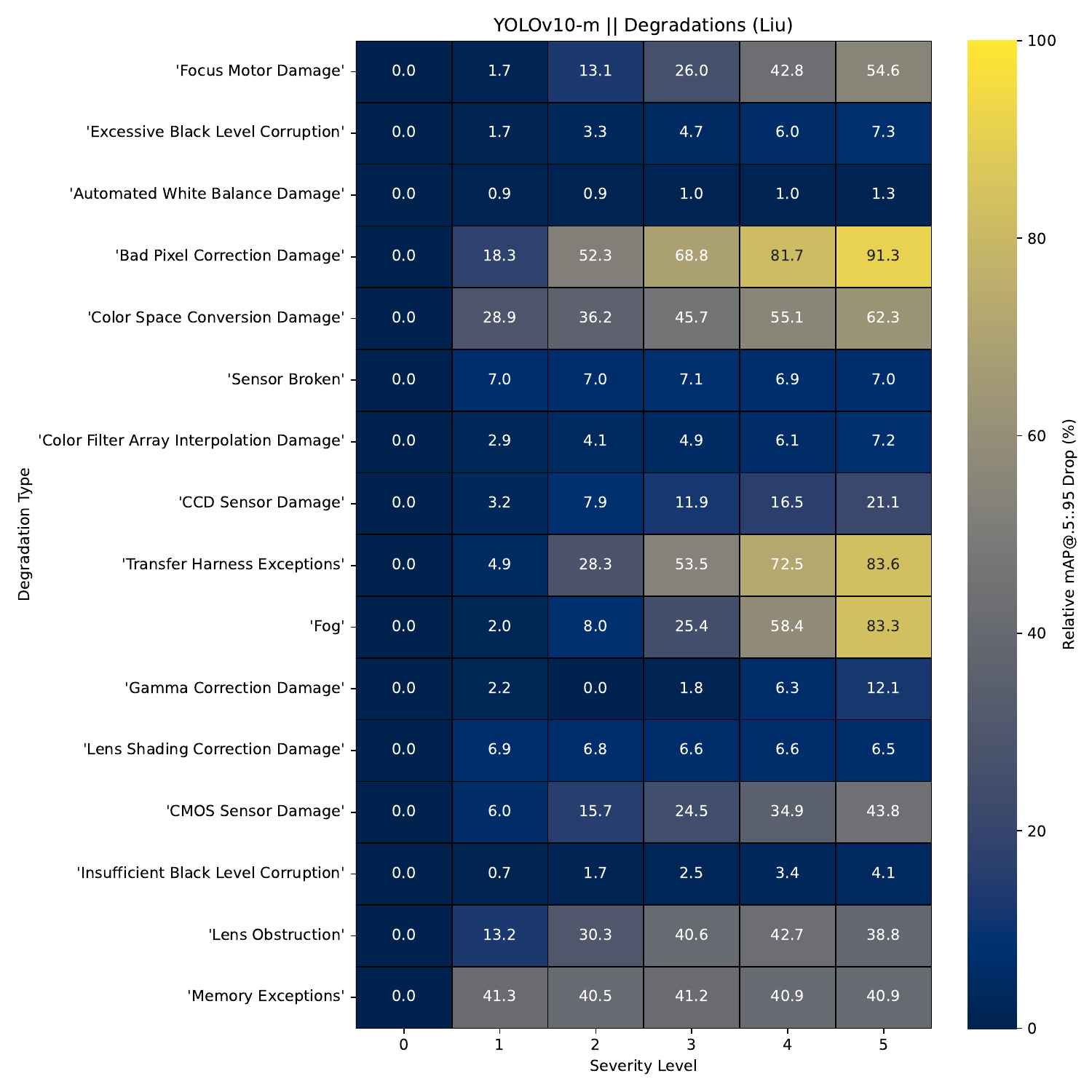}
		\caption*{Camera/System failures\\Degradations from~\cite{Liu_IJCV_2024}}
	\end{subfigure}
	\caption{
		Relative mAP@0.5:0.95 drop of YOLOv10-m under increasing native severity. Measured severity semantics help explain why equal ordinal levels can produce very different robustness behavior across degradation sources.
	}
	\label{fig:map_vs_severity_all}
\end{figure*}

\begin{table*}[t]
	\centering
	\scriptsize
	\setlength{\tabcolsep}{2pt}
	\resizebox{\textwidth}{!}{%
		\begin{tabular}{l|c:c:c:c:c|c:c:c:c:c|c:c:c:c:c|}
			\hline
			& \multicolumn{5}{c|}{(1-SSIM)}
			& \multicolumn{5}{c|}{PSNR / dB}
			& \multicolumn{5}{c|}{LPIPS} \\
			& \multicolumn{5}{c|}{Severity Level}
			& \multicolumn{5}{c|}{Severity Level}
			& \multicolumn{5}{c|}{Severity Level} \\
			Degradation Function \cite{Hendrycks_ICLR_2019}
			& 1 & 2 & 3 & 4 & 5
			& 1 & 2 & 3 & 4 & 5
			& 1 & 2 & 3 & 4 & 5 \\
			\hline
			\rowcolor{Set3-08!20}
			'Gaussian Noise'
			& 0.010 & 0.017 & 0.027 & 0.042 & 0.067
			& 22.345 & 19.016 & 15.811 & 13.094 & 10.628
			& 0.329 & 0.419 & 0.513 & 0.595 & 0.664 \\
			\rowcolor{gray!20}
			'Shot Noise'
			& 0.010 & 0.017 & 0.026 & 0.043 & 0.060
			& 22.072 & 18.487 & 15.593 & 12.325 & 10.527
			& 0.314 & 0.411 & 0.497 & 0.595 & 0.645 \\
			\rowcolor{Set3-08!20}
			'Impulse Noise'
			& 0.011 & 0.019 & 0.027 & 0.045 & 0.069
			& 19.993 & 16.979 & 15.221 & 12.456 & 10.448
			& 0.309 & 0.421 & 0.491 & 0.597 & 0.662 \\
			\rowcolor{gray!20}
			'Speckle Noise'
			& 0.009 & 0.013 & 0.024 & 0.033 & 0.044
			& 23.301 & 20.939 & 16.461 & 14.594 & 12.736
			& 0.259 & 0.314 & 0.435 & 0.491 & 0.549 \\
			\rowcolor{Set3-08!20}
			'Defocus Blur'
			& 0.006 & 0.009 & 0.017 & 0.025 & 0.031
			& 24.578 & 23.613 & 22.239 & 21.383 & 20.694
			& 0.276 & 0.340 & 0.428 & 0.489 & 0.534 \\
			\rowcolor{gray!20}
			'Gaussian Blur'
			& 0.003 & 0.009 & 0.015 & 0.021 & 0.032
			& 27.526 & 24.271 & 22.856 & 21.946 & 20.750
			& 0.171 & 0.316 & 0.402 & 0.462 & 0.544 \\
			\rowcolor{Set3-08!20}
			'Glass Blur'
			& 0.014 & 0.012 & 0.027 & 0.025 & 0.029
			& 23.517 & 23.712 & 20.866 & 21.320 & 20.837
			& 0.249 & 0.309 & 0.418 & 0.441 & 0.491 \\
			\rowcolor{gray!20}
			'Motion Blur'
			& 0.017 & 0.027 & 0.038 & 0.048 & 0.055
			& 23.776 & 21.807 & 20.248 & 19.123 & 18.535
			& 0.197 & 0.286 & 0.364 & 0.424 & 0.458 \\
			\rowcolor{Set3-08!20}
			'Zoom Blur'
			& 0.051 & 0.061 & 0.068 & 0.076 & 0.082
			& 19.132 & 18.137 & 17.807 & 17.237 & 16.924
			& 0.395 & 0.444 & 0.431 & 0.458 & 0.449 \\
			\rowcolor{gray!20}
			'Fog'
			& 0.163 & 0.179 & 0.193 & 0.196 & 0.207
			& 13.943 & 12.919 & 12.286 & 12.176 & 11.782
			& 0.206 & 0.255 & 0.311 & 0.340 & 0.404 \\
			\rowcolor{Set3-08!20}
			'Frost'
			& 0.230 & 0.280 & 0.297 & 0.279 & 0.297
			& 12.356 & 10.562 & 9.711 & 10.349 & 9.813
			& 0.248 & 0.342 & 0.394 & 0.403 & 0.434 \\
			\rowcolor{gray!20}
			'Snow'
			& 0.156 & 0.248 & 0.247 & 0.296 & 0.347
			& 15.301 & 11.269 & 11.200 & 9.566 & 8.467
			& 0.272 & 0.409 & 0.398 & 0.457 & 0.487 \\
			\rowcolor{Set3-08!20}
			'Spatter'
			& 0.002 & 0.016 & 0.028 & 0.033 & 0.053
			& 38.682 & 24.780 & 22.496 & 19.066 & 16.897
			& 0.030 & 0.178 & 0.264 & 0.312 & 0.394 \\
			\rowcolor{gray!20}
			'Brightness'
			& 0.077 & 0.147 & 0.209 & 0.261 & 0.304
			& 21.719 & 16.022 & 12.884 & 10.845 & 9.436
			& 0.042 & 0.095 & 0.153 & 0.209 & 0.263 \\
			\rowcolor{Set3-08!20}
			'Contrast'
			& 0.111 & 0.130 & 0.148 & 0.167 & 0.176
			& 16.917 & 15.578 & 14.419 & 13.396 & 12.926
			& 0.171 & 0.242 & 0.339 & 0.489 & 0.598 \\
			\rowcolor{gray!20}
			'Saturate'
			& 0.051 & 0.066 & 0.050 & 0.115 & 0.159
			& 22.925 & 20.807 & 22.877 & 15.516 & 12.717
			& 0.112 & 0.187 & 0.083 & 0.256 & 0.337 \\
			\rowcolor{Set3-08!20}
			'JPEG'
			& 0.008 & 0.010 & 0.012 & 0.016 & 0.021
			& 28.112 & 27.216 & 26.692 & 25.502 & 24.363
			& 0.215 & 0.258 & 0.286 & 0.354 & 0.416 \\
			\rowcolor{gray!20}
			'Pixelate'
			& 0.003 & 0.003 & 0.008 & 0.006 & 0.008
			& 27.653 & 26.810 & 24.469 & 24.087 & 23.457
			& 0.108 & 0.143 & 0.227 & 0.295 & 0.340 \\
			\rowcolor{Set3-08!20}
			'Elastic Transform'
			& 0.099 & 0.137 & 0.036 & 0.038 & 0.042
			& 14.375 & 12.777 & 18.690 & 18.603 & 18.153
			& 0.424 & 0.511 & 0.207 & 0.332 & 0.404 \\
			\bottomrule
		\end{tabular}
	}
	\caption{
		Measured degradation strengths for COCO images under ImageNet-C style degradations (\cite{Hendrycks_ICLR_2019} and extensions~\cite{Michaelis_NeurIPSW_2019,Oksuz_CVPR_2023}). Columns report $(1-\mathrm{SSIM})$, PSNR (dB), and LPIPS for native severity levels 1--5. For $(1-\mathrm{SSIM})$ and LPIPS, higher indicates stronger degradation; for PSNR, lower indicates stronger degradation.
	}
	\label{tab:degradation_strengths_all_metrics_hendrycks}
\end{table*}

\textbf{Findings.}
Figure~\ref{fig:map_vs_severity_all} shows three consistent observations. First, equal native severity indices are not comparable across backends: a nominal level~3 can be barely measurable under one implementation and severe under another. Table~\ref{tab:degradation_strengths_all_metrics_hendrycks} makes this explicit even within a single backend, where native severity steps are not uniformly spaced in PSNR, $(1-\mathrm{SSIM})$, or LPIPS. This confirms that ordinal severity alone is insufficient for principled interpretation.

Second, ImageNet-C style operators generally produce stronger performance degradation than IQA-motivated distortions at the same nominal level. This is consistent with their design goals: corruption-benchmark operators are constructed as robustness stress tests that aggressively disrupt semantic and structural cues, whereas IQA distortions are designed to span perceptually meaningful quality variations that may preserve object-level structure.

Third, real-camera and system-level failures expose qualitatively different robustness modes from both synthetic corruptions and IQA distortions. In particular, they include abrupt failure patterns and degradation signatures that are not well represented by conventional benchmark operators, highlighting the value of a causally grounded organization beyond purely phenomenological grouping.

The measurement layer is useful precisely because it separates nominal severity from observed distortion strength. It enables robustness results to be interpreted in terms of both \emph{what} a degradation is and \emph{how strong} it actually is under a given backend. This is particularly important for cross-family comparisons and for identifying weakly monotonic, saturated, or irregular severity schedules.

\textbf{Why severity measurement matters.}
Most robustness benchmarks treat severity levels as ordinal indices whose semantics are implicitly defined by backend parameters. Our results show that equal nominal severity levels can correspond to markedly different distortion magnitudes, both across backends and across degradation families within the same backend. By explicitly measuring degradation strength, the proposed framework enables analyses beyond ordinal comparison, including relating performance drop to observable distortion magnitude, identifying weakly monotonic or saturated severity schedules, and separating backend-specific severity conventions from the actual visual strength of a degradation.

Spatially transforming degradations such as \degfunc{Elastic Transform} should be interpreted with care, because ground-truth annotations are not warped jointly with the image. In such cases, performance loss reflects both visual degradation and annotation mismatch.

To keep the main paper focused, full per-backend severity tables, cross-detector comparisons, degradation galleries, taxonomy mappings, and additional application protocols are provided in the supplementary material.
\section{Discussion and Limitations}
\label{sec:discussion}
The proposed framework is an interpretive layer rather than a complete physical model of image formation. The taxonomy provides a compact way to relate degradations across corruption benchmarks, IQA datasets, and real-camera studies, but it necessarily simplifies mixed-cause and compound effects.

The framework is not intended to recover a hidden degradation history from the final image. For known sequential chains, such as noise followed by blur, the applied operators and their order can be retained at the evaluation-protocol level even when one effect visually masks another. The main experiments intentionally use single degradations to isolate severity semantics; ordered compositional protocols are documented in the supplementary material.

Its purpose is to make benchmark assumptions explicit and comparable, not to define a unique ontology of degradations.

Likewise, the severity measurement layer is descriptive rather than predictive. PSNR, SSIM, and LPIPS provide complementary views of distortion magnitude, but they do not directly measure task relevance. We therefore use metric-space severity as an analysis tool, not as a surrogate for semantic difficulty. This distinction is particularly important for low-level vision pipelines: a restoration, enhancement, or compression method may improve one image-quality metric while leaving downstream robustness unchanged, or even reducing task performance. The proposed measurement layer is therefore intended to expose degradation strength, not to collapse perceptual quality and machine-task difficulty into a single score.

The current measurement instantiation assumes paired clean and degraded images, as available when benchmark degradations are generated from a pristine reference. For naturally degraded real-world imagery without a paired reference, the taxonomy remains applicable, but the present full-reference severity measurement cannot be evaluated directly. Reference-free measures, including no-reference IQA models, could be incorporated as an alternative measurement layer.

A further limitation concerns degradations that alter spatial correspondence. For transformations such as \degfunc{Elastic Transform}, detector performance may decrease partly because annotations are no longer geometrically aligned with the image. In such cases, measured robustness conflates visual degradation with evaluation mismatch.

The current experiments focus on frame-wise object detection. The cross-detector experiments in the supplementary material test architectural consistency within detection, not task generality. Extending the framework to classification, semantic or instance segmentation, restoration-aware evaluation, video detection, tracking, and other perception tasks is a natural next step.

\section{Conclusion}
\label{sec:conclusion}
We presented a compact framework for interpreting robustness under diverse image degradations. The proposed dual-axis taxonomy characterizes each degradation operator by its dominant cause and perceptual effect, while the severity measurement layer characterizes its realized strength under the native backend definition.

A central design choice is to preserve native degradation implementations rather than replace them with a universal severity scale. This keeps the framework compatible with existing benchmarks while making their assumptions explicit. COCO-Degradation demonstrates that this additional structure yields more interpretable robustness analysis across corruption benchmarks, IQA distortions, and physically motivated camera or system failures.

Beyond object detection, the same representation can support degradation-aware evaluation of restoration, enhancement, compression, and preprocessing pipelines when their outputs are consumed by machine-perception systems.

We hope the accompanying \texttt{imdeg} library supports more reproducible and better structured robustness studies across datasets, tasks, and imaging pipelines.

\bibliographystyle{splncs04}
\bibliography{reference}

\clearpage
\section*{Supplementary Material}
\addcontentsline{toc}{section}{Supplementary Material}
This supplementary material collects content that was omitted from the main paper for space reasons while remaining important for reproducibility and detailed interpretation. It includes additional cross-detector robustness comparisons, full per-backend severity tables, visual galleries of degradation functions, complete taxonomy mappings, temporal/video subtype definitions, extended related-work notes, licensing information, and example usage of the \texttt{imdeg} library.

\phantomsection
\section*{Supplementary Evaluation: Detection Performance}
\addcontentsline{toc}{subsection}{Supplementary Evaluation: Detection Performance}
\label{app:detection_performance}
To complement the main-paper experiments, we evaluate additional detector families under the same COCO-Degradation protocol. While YOLOv10 serves as the primary reference model in the main paper, the results reported here assess whether the observed degradation trends remain consistent across representative one-stage and transformer-based detectors, namely YOLOv9, YOLOv11, and RT-DETR.

Unless noted otherwise, all models use COCO (\texttt{train2017}) pretrained weights from the Ultralytics implementations, an input size of 640, and a confidence threshold of 0.001. Relative and absolute performance plots are reported using mAP@0.5:0.95 under the same single-degradation, native-severity setting as in the main paper.

\smallsec{Detectors.}
YOLOv9~\cite{Wang_ECCV_2024} is a one-stage detector emphasizing information preservation through PGI and GELAN. YOLO11~\cite{yolo11_ultralytics} is a recent YOLO-family variant with incremental architectural refinements for speed--accuracy trade-offs. RT-DETR~\cite{Zhao_2024_CVPR} is an end-to-end transformer-based detector that performs set prediction without NMS.

\smallsec{Interpretation.}
These supplementary results are intended as a consistency check across architectures rather than as a standalone detector comparison. The central question is whether the degradation trends discussed in the main paper persist across different model families.

\renewcommand{\subw}{0.3\textwidth}
\begin{figure*}[h!]
	\captionsetup[subfigure]{justification=centering}
	\centering
	\begin{subfigure}{\subw}
		\centering
		\includegraphics[width=\textwidth]{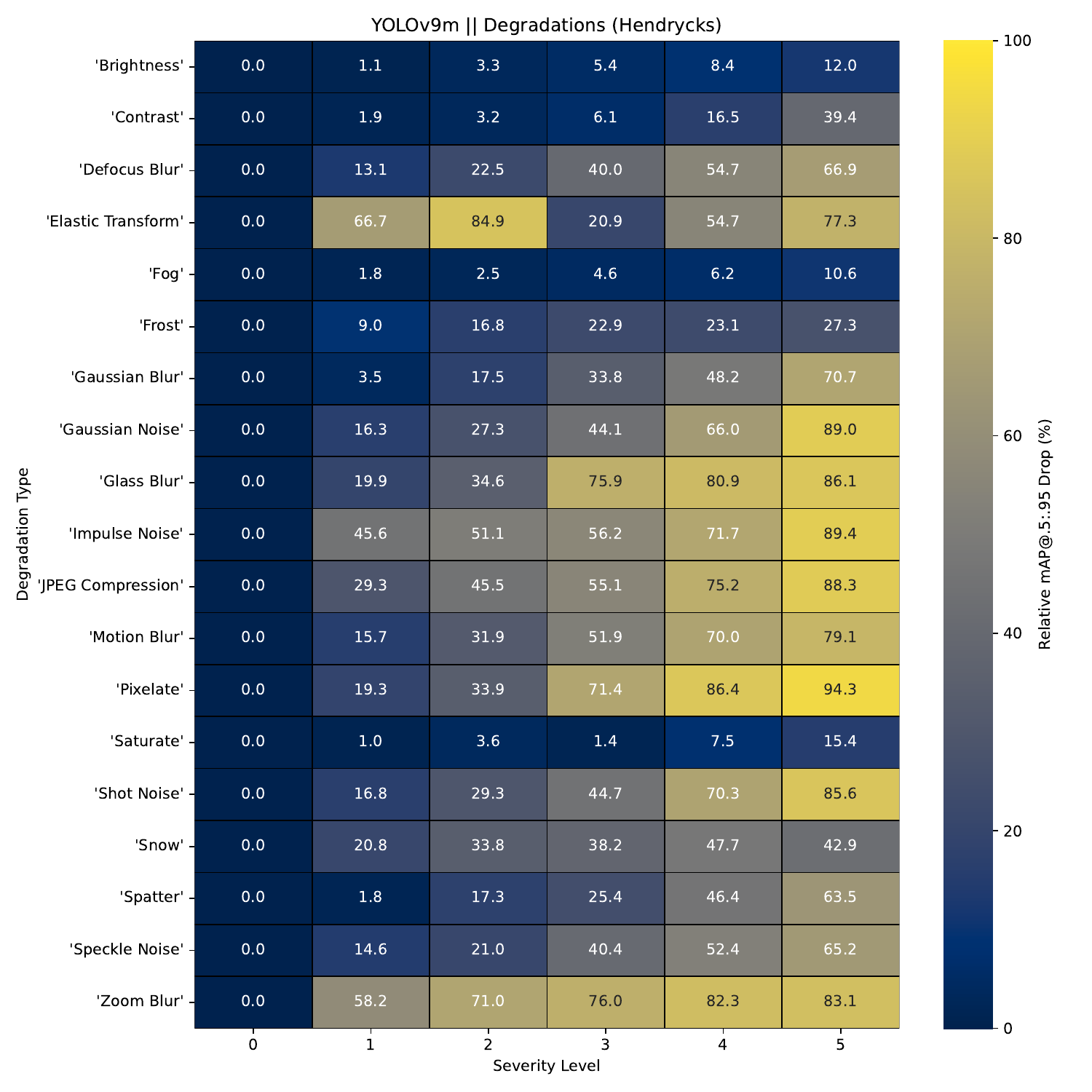}
		\caption*{Degradations from~\cite{Hendrycks_ICLR_2019}}
	\end{subfigure}
	\hfill
	\begin{subfigure}{\subw}
		\centering
		\includegraphics[width=\textwidth]{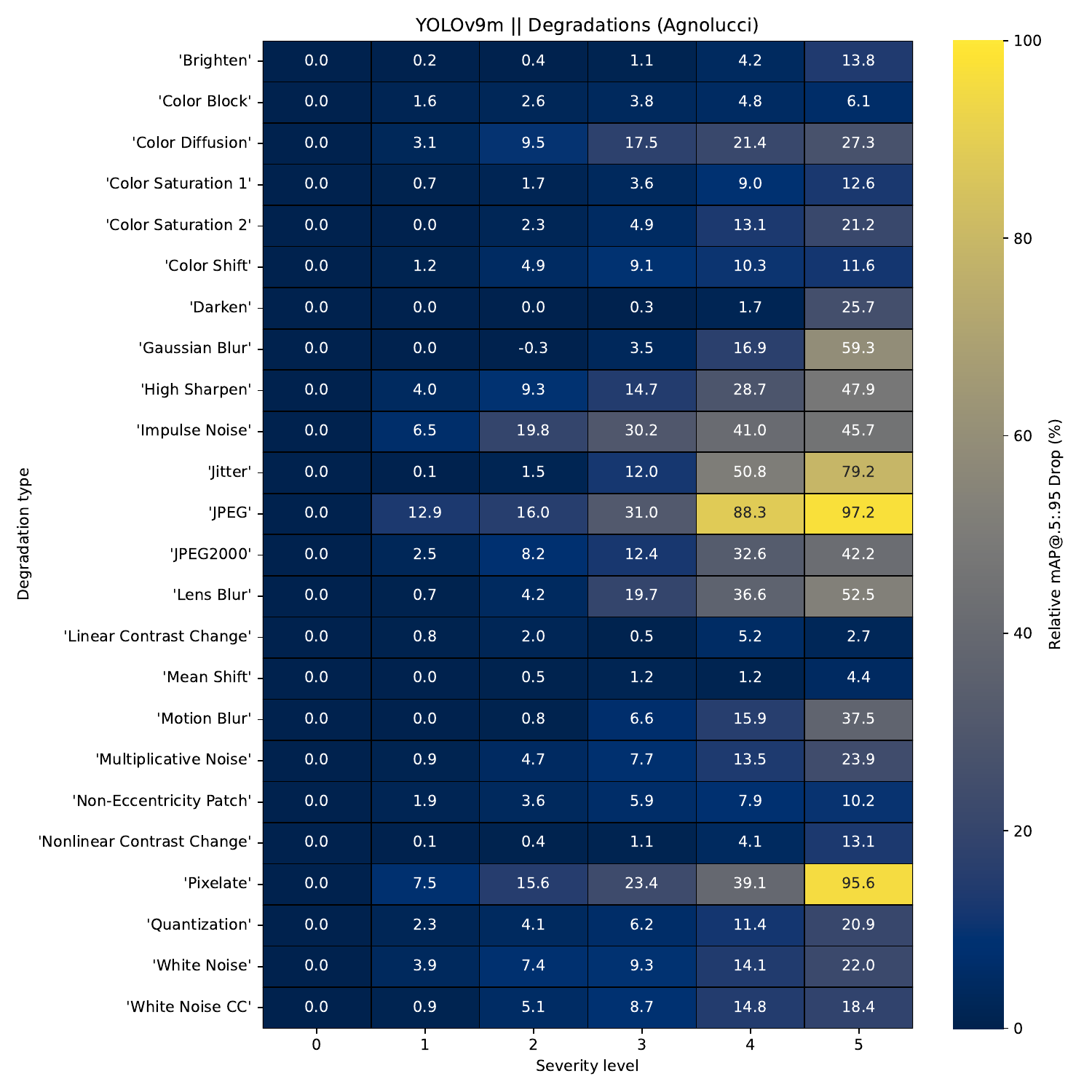}
		\caption*{Degradations from~\cite{Agnolucci_WACV_2024}}
	\end{subfigure}
	\hfill
	\begin{subfigure}{\subw}
		\centering
		\includegraphics[width=\textwidth]{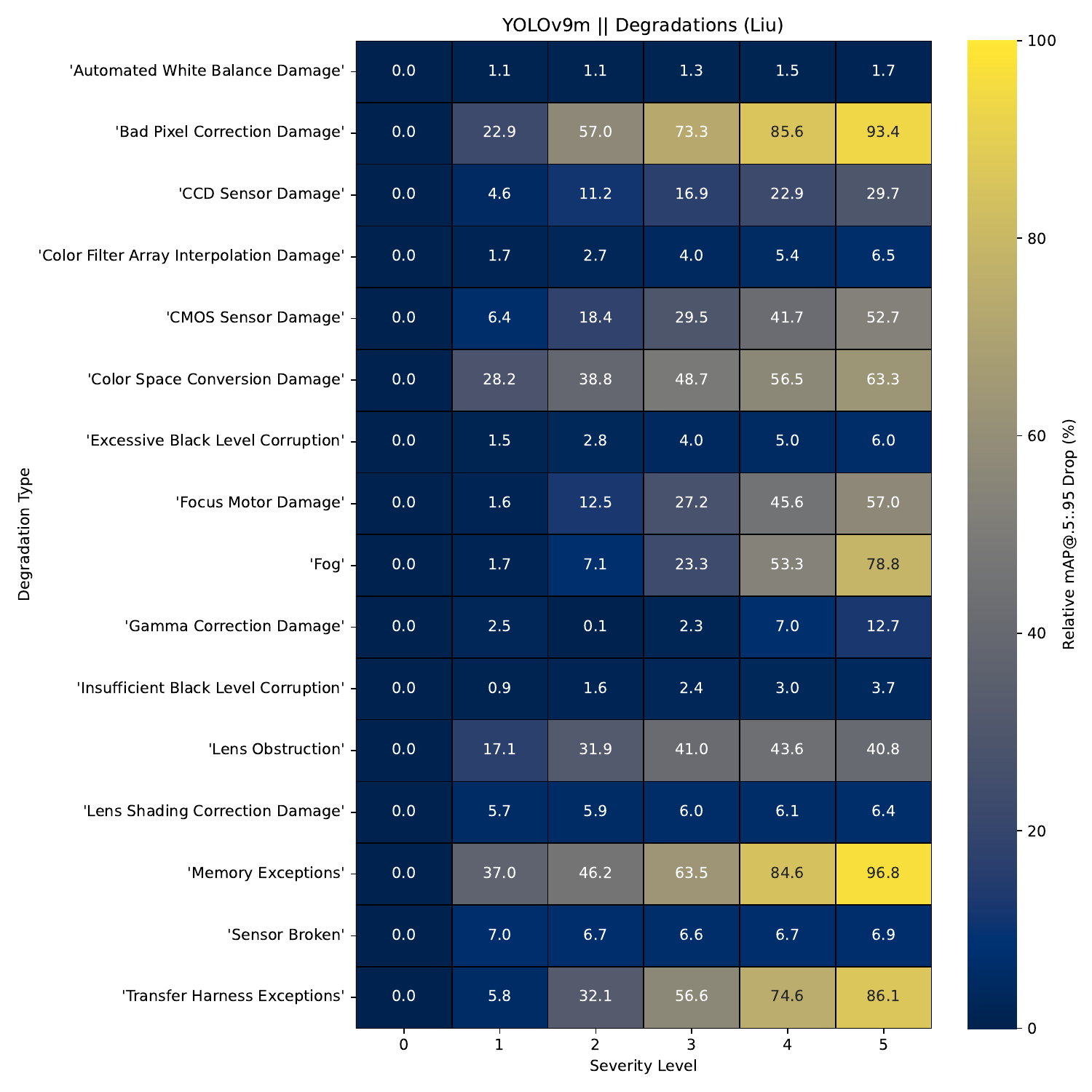}
		\caption*{Degradations from~\cite{Liu_IJCV_2024}}
	\end{subfigure}
	\caption{
		Relative detection performance drop of YOLOv9-m~\cite{Wang_ECCV_2024} on COCO \texttt{val2017} under increasing native severity levels (mAP@0.5:0.95). Each cell shows the percentage drop relative to the pristine COCO baseline.
	}
	\label{fig:map_vs_severity_all_YOLOv9-m}
\end{figure*}

\renewcommand{\subw}{0.3\textwidth}
\begin{figure*}[h!]
	\captionsetup[subfigure]{justification=centering}
	\centering
	\begin{subfigure}{\subw}
		\centering
		\includegraphics[width=\textwidth]{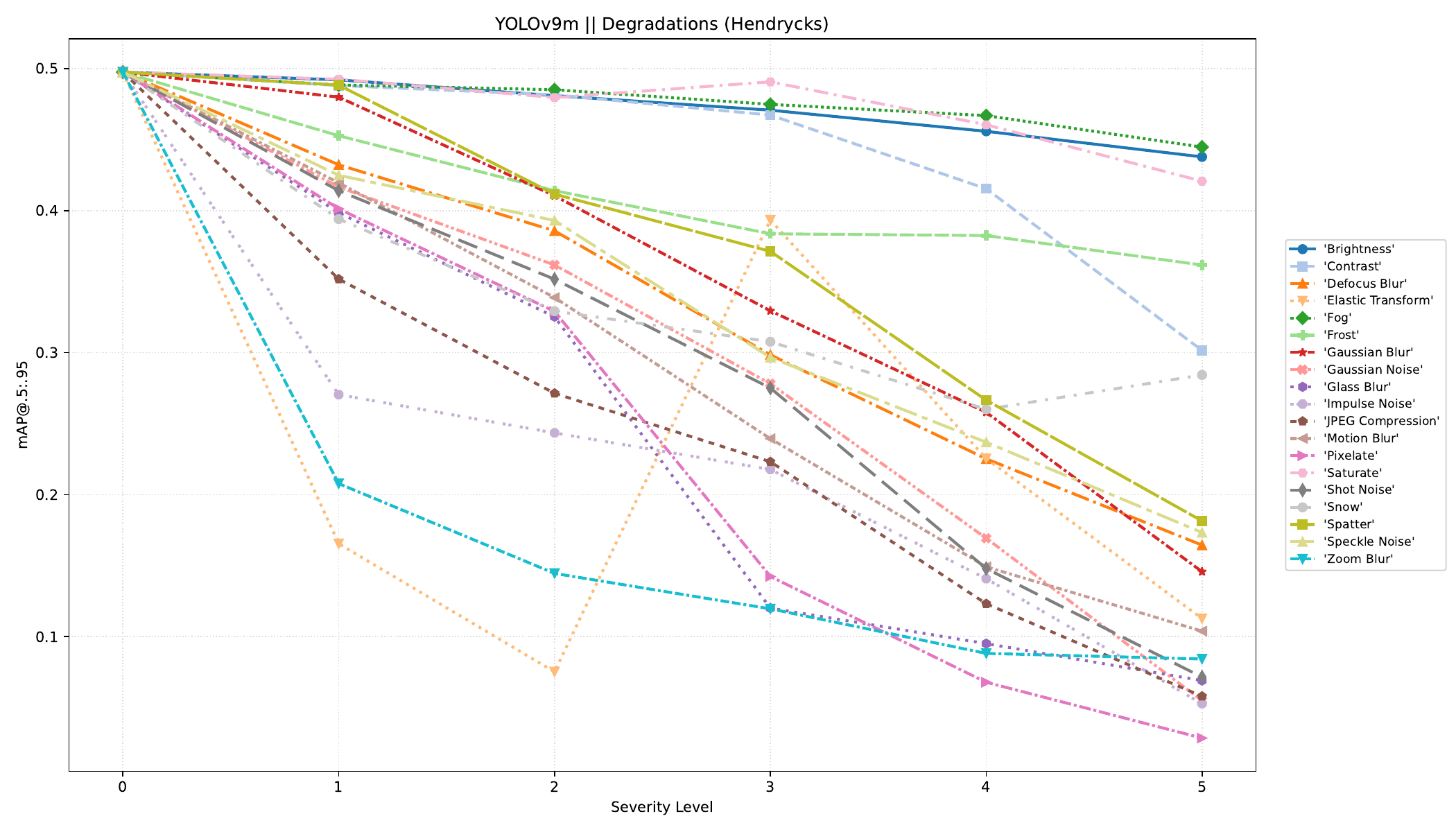}
		\caption*{Degradations from~\cite{Hendrycks_ICLR_2019}}
	\end{subfigure}
	\hfill
	\begin{subfigure}{\subw}
		\centering
		\includegraphics[width=\textwidth]{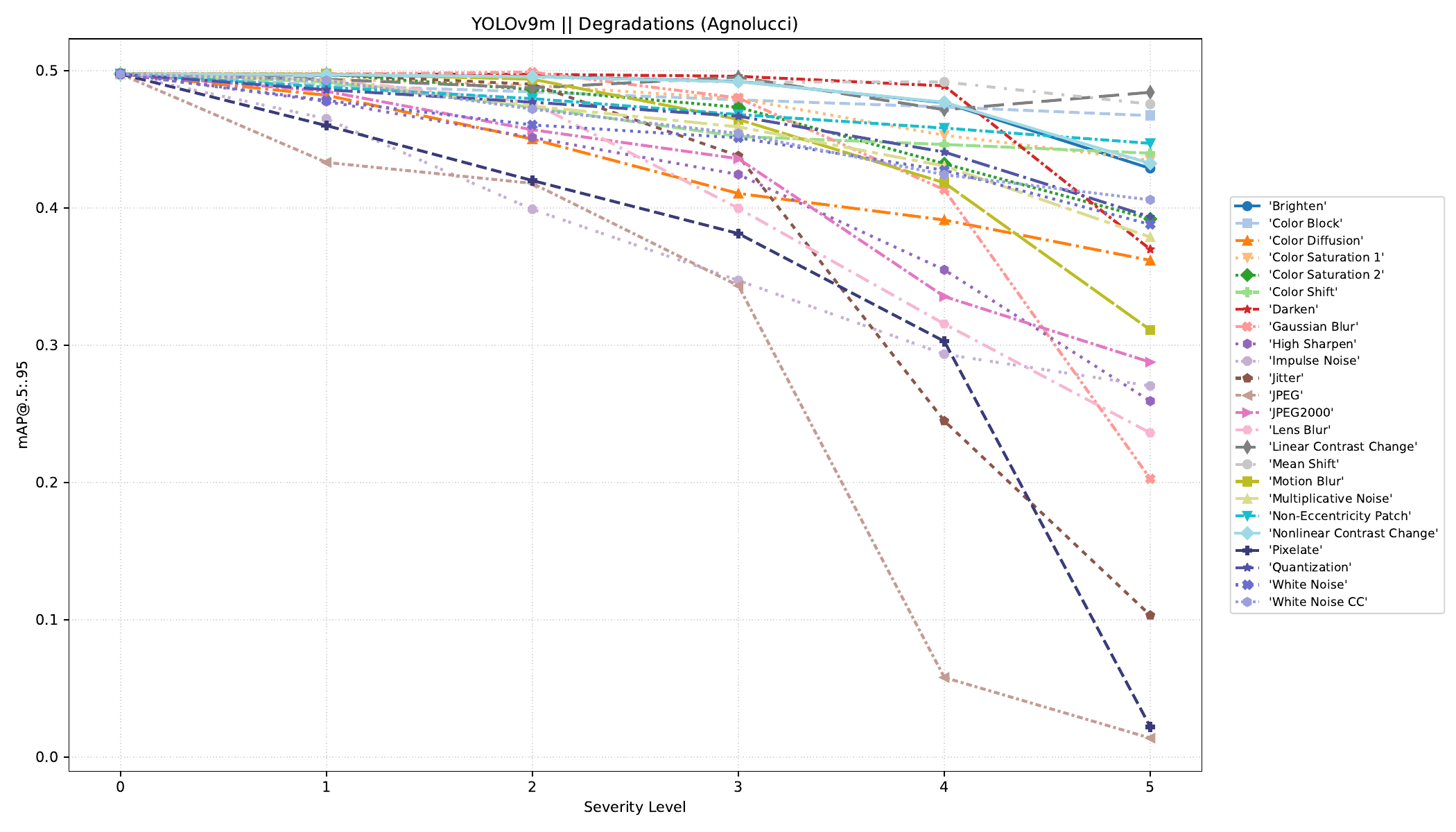}
		\caption*{Degradations from~\cite{Agnolucci_WACV_2024}}
	\end{subfigure}
	\hfill
	\begin{subfigure}{\subw}
		\centering
		\includegraphics[width=\textwidth]{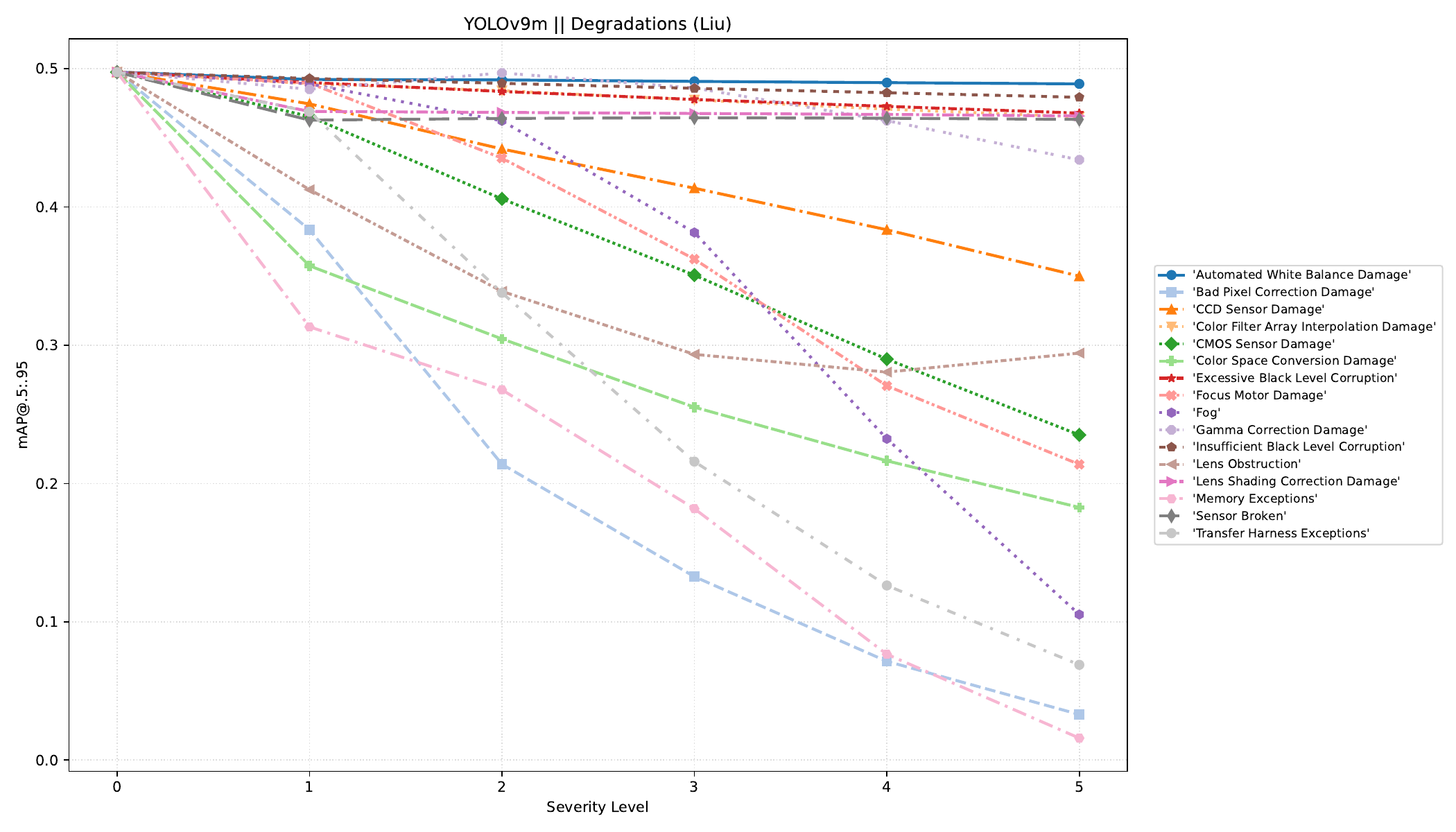}
		\caption*{Degradations from~\cite{Liu_IJCV_2024}}
	\end{subfigure}
	\caption{
		Absolute detection performance of YOLOv9-m~\cite{Wang_ECCV_2024} on COCO \texttt{val2017} across native severity levels (mAP@0.5:0.95).
	}
	\label{fig:map_vs_severity_all_absolute_yolov9m}
\end{figure*}

\renewcommand{\subw}{0.3\textwidth}
\begin{figure*}[h!]
	\captionsetup[subfigure]{justification=centering}
	\centering
	\begin{subfigure}{\subw}
		\centering
		\includegraphics[width=\textwidth]{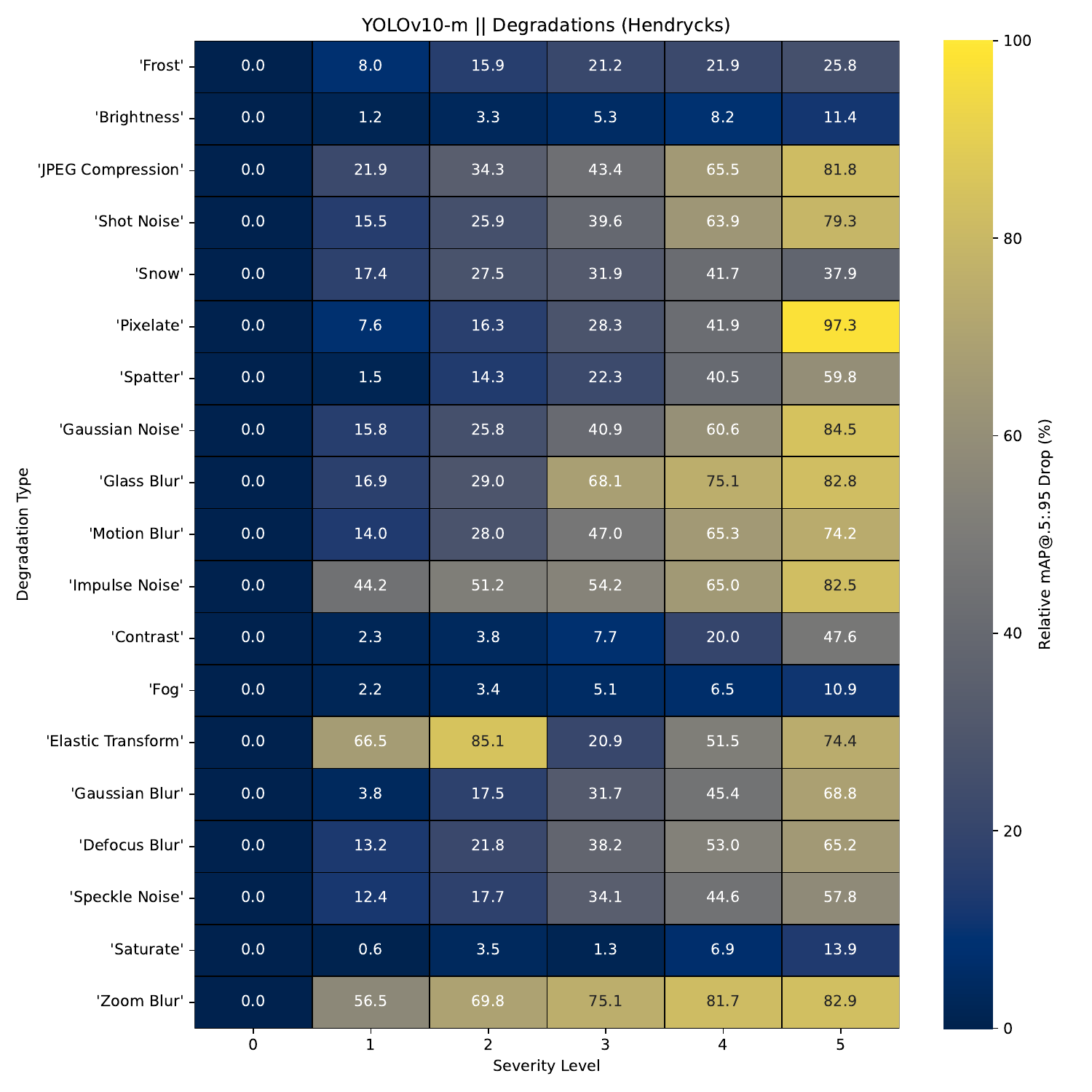}
		\caption*{Degradation functions from~\cite{Hendrycks_ICLR_2019}}
	\end{subfigure}
	\hfill
	\begin{subfigure}{\subw}
		\centering 
		\includegraphics[width=\textwidth]{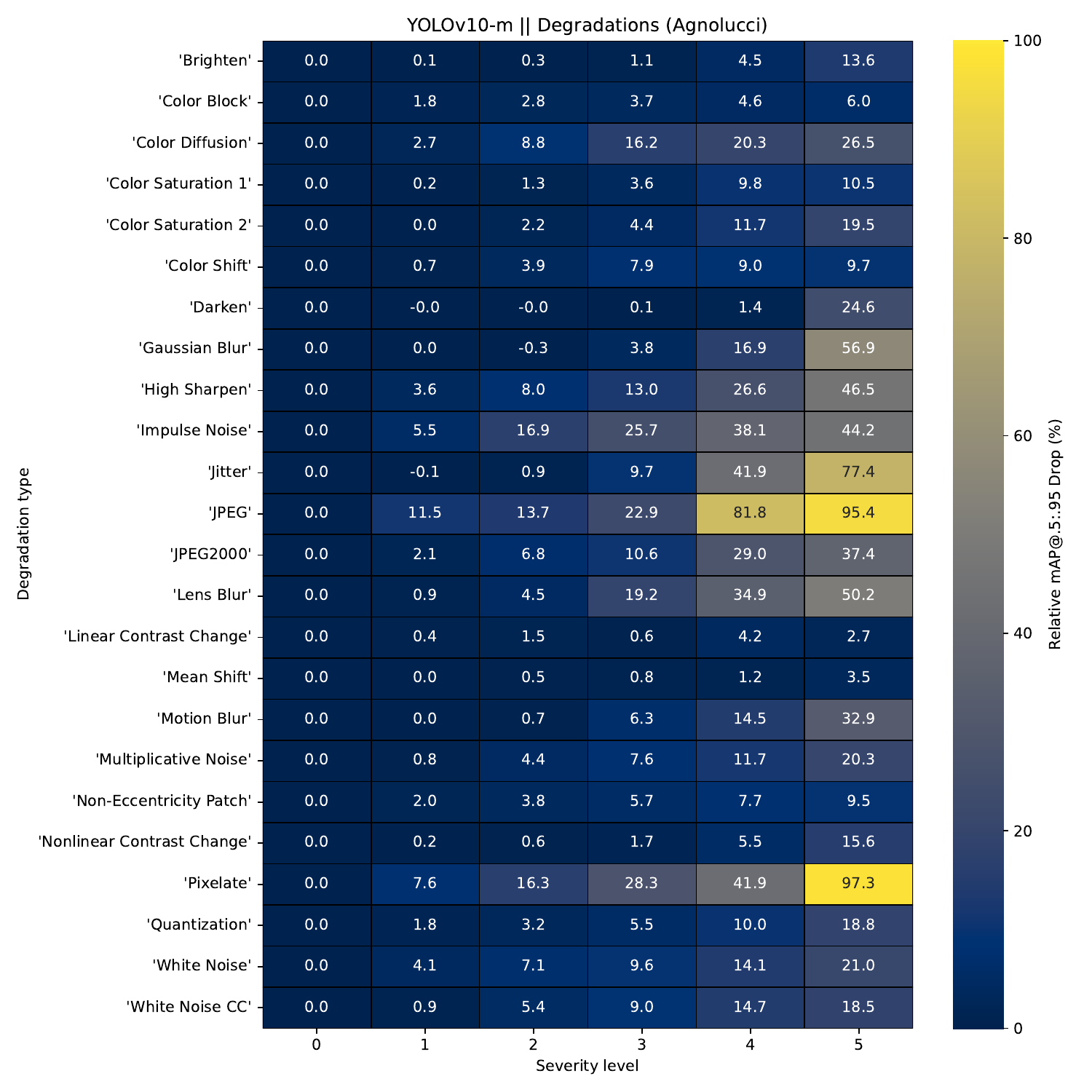}
		\caption*{Degradation functions from~\cite{Agnolucci_WACV_2024}}
	\end{subfigure}
	\hfill
	\begin{subfigure}{\subw}
		\centering 
		\includegraphics[width=\textwidth]{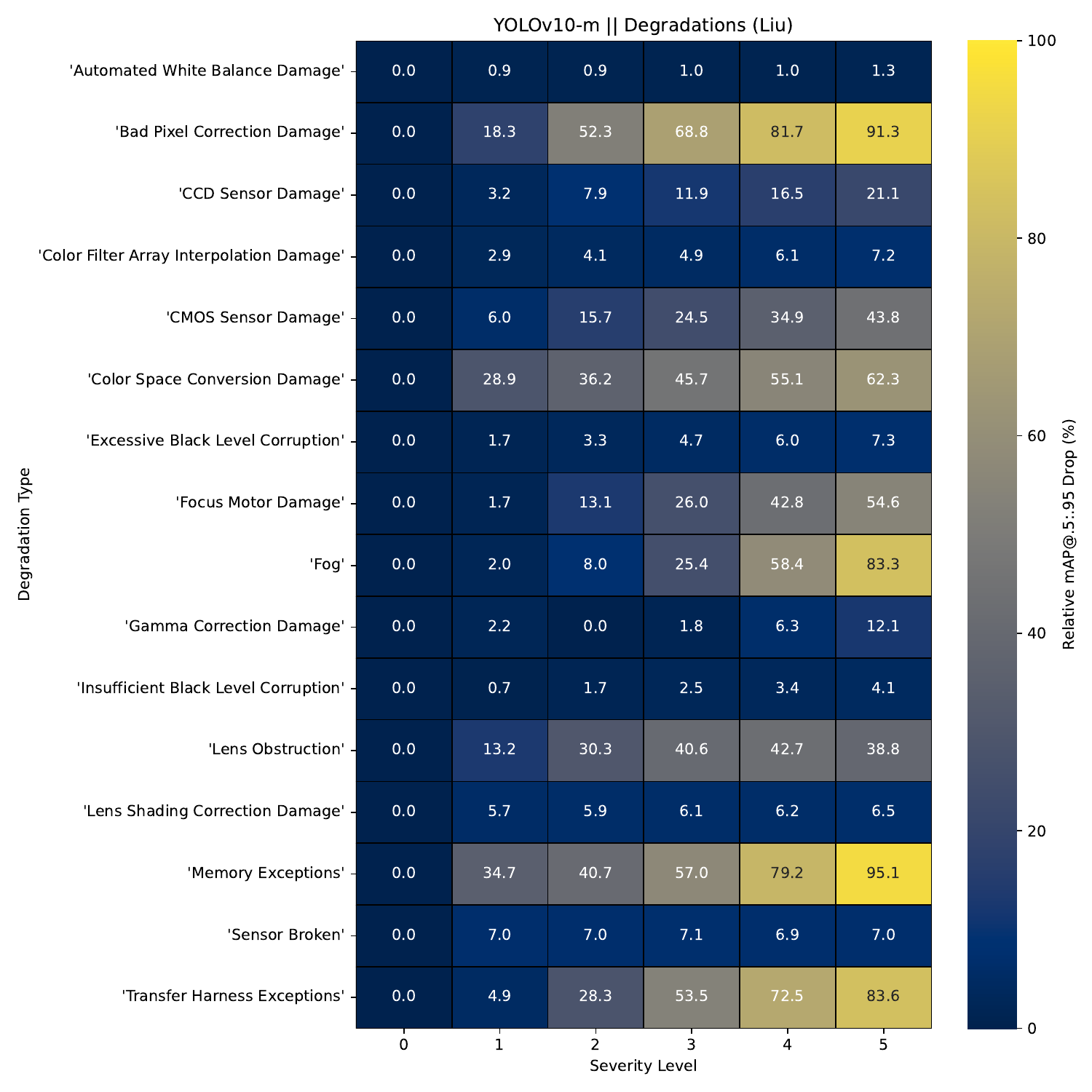}
		\caption*{Degradation functions from~\cite{Liu_IJCV_2024}}
	\end{subfigure}
	\caption{
		Relative detection performance drop of YOLOv10-m~\cite{Wang_NeurIPS_2024} on COCO \texttt{val2017} under increasing \emph{native} severity levels
		(mAP@0.5:0.95). Each cell shows the percentage drop relative to the clean \texttt{val2017} baseline (pristine COCO images, no added degradations).
		Brighter cells indicate larger drops. Left: degradations from~\cite{Hendrycks_ICLR_2019}. Middle: degradations
		from~\cite{Agnolucci_WACV_2024}. Right: degradations from~\cite{Liu_IJCV_2024}.
	}
	\label{fig:map_vs_severity_all_yolov10m}
\end{figure*}
\renewcommand{\subw}{0.3\textwidth}
\begin{figure*}[h!]
	\captionsetup[subfigure]{justification=centering}
	\centering
	\begin{subfigure}{\subw}
		\centering
		\includegraphics[width=\textwidth]{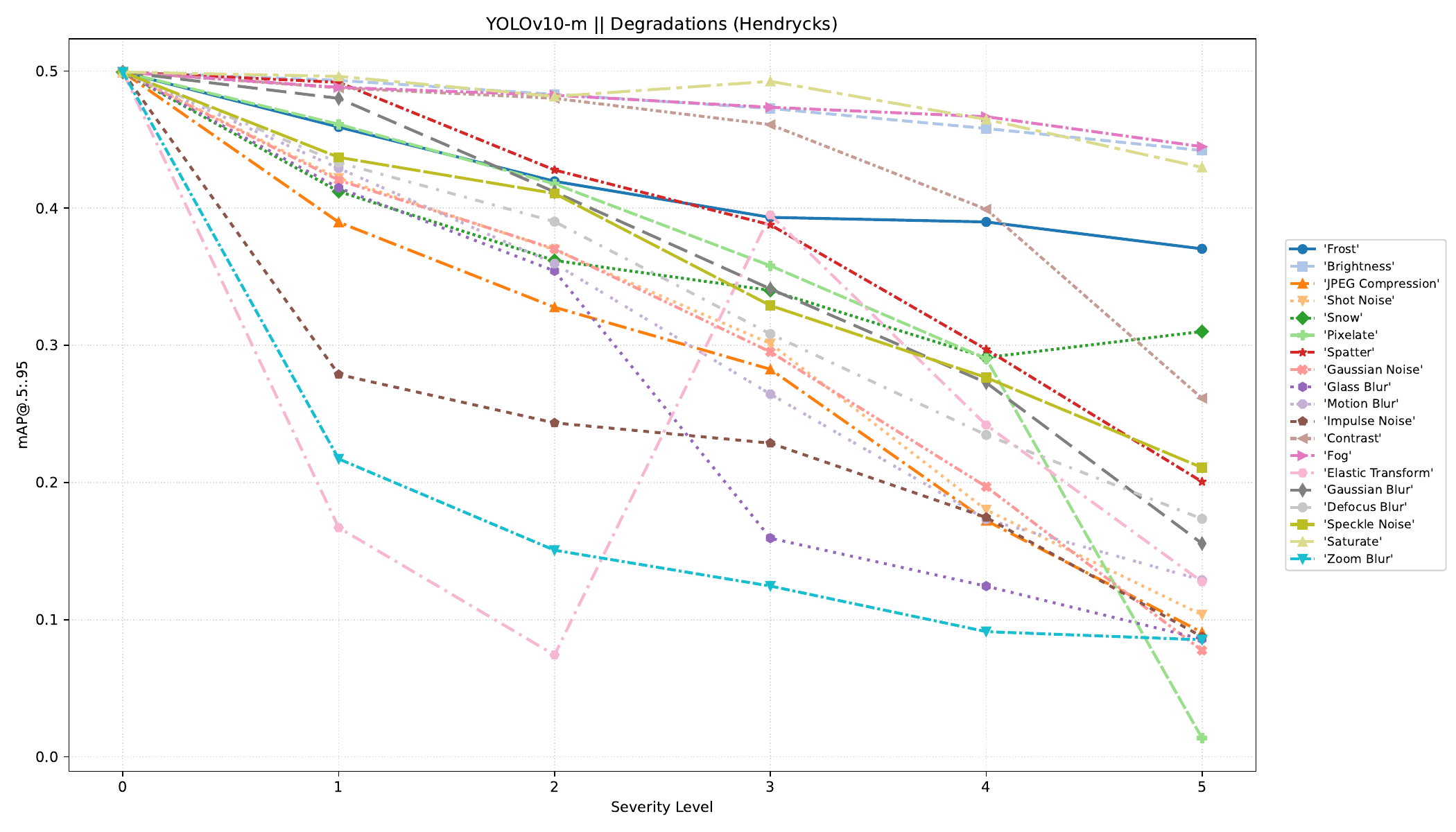}
		\caption*{Degradation functions from~\cite{Hendrycks_ICLR_2019}}
	\end{subfigure}
	\hfill
	\begin{subfigure}{\subw}
		\centering
		\includegraphics[width=\textwidth]{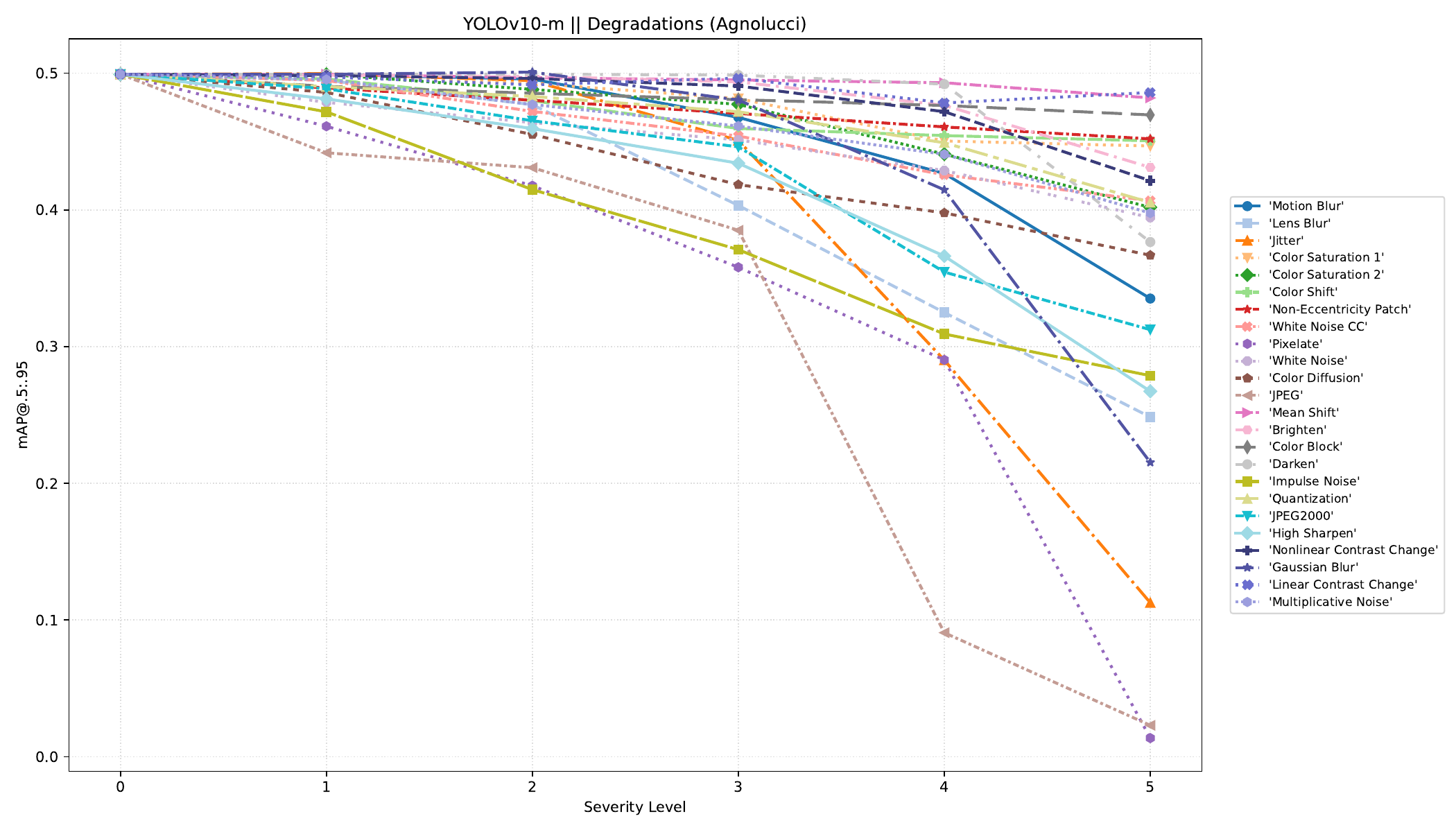}
		\caption*{Degradation functions from~\cite{Agnolucci_WACV_2024}}
	\end{subfigure}
	\hfill
	\begin{subfigure}{\subw}
		\centering
		\includegraphics[width=\textwidth]{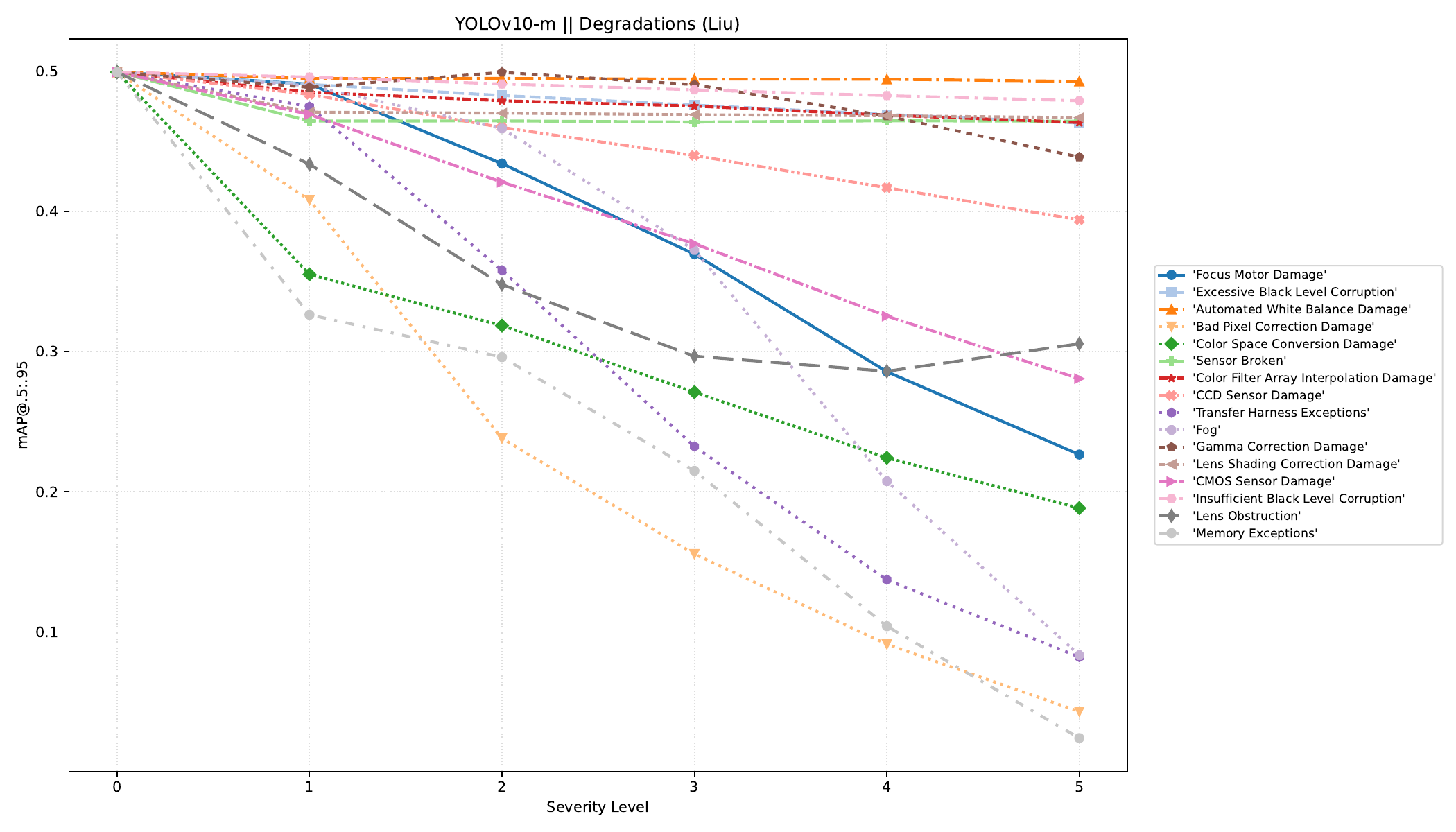}
		\caption*{Degradation functions from~\cite{Liu_IJCV_2024}}
	\end{subfigure}
	\caption{
		Absolute detection performance (mAP@0.5:0.95) of YOLOv10-m~\cite{Wang_NeurIPS_2024} across native severity levels for
		different image degradations. Left: degradations from~\cite{Hendrycks_ICLR_2019}. Middle: degradations from~\cite{Liu_IJCV_2024}. Right: degradations from~\cite{Agnolucci_WACV_2024}. Each curve corresponds to one degradation type.
	}	
	\label{fig:map_vs_severity_all_absolute_yolov10m}
\end{figure*}

\renewcommand{\subw}{0.3\textwidth}
\begin{figure*}[h!]
	\captionsetup[subfigure]{justification=centering}
	\centering
	\begin{subfigure}{\subw}
		\centering
		\includegraphics[width=\textwidth]{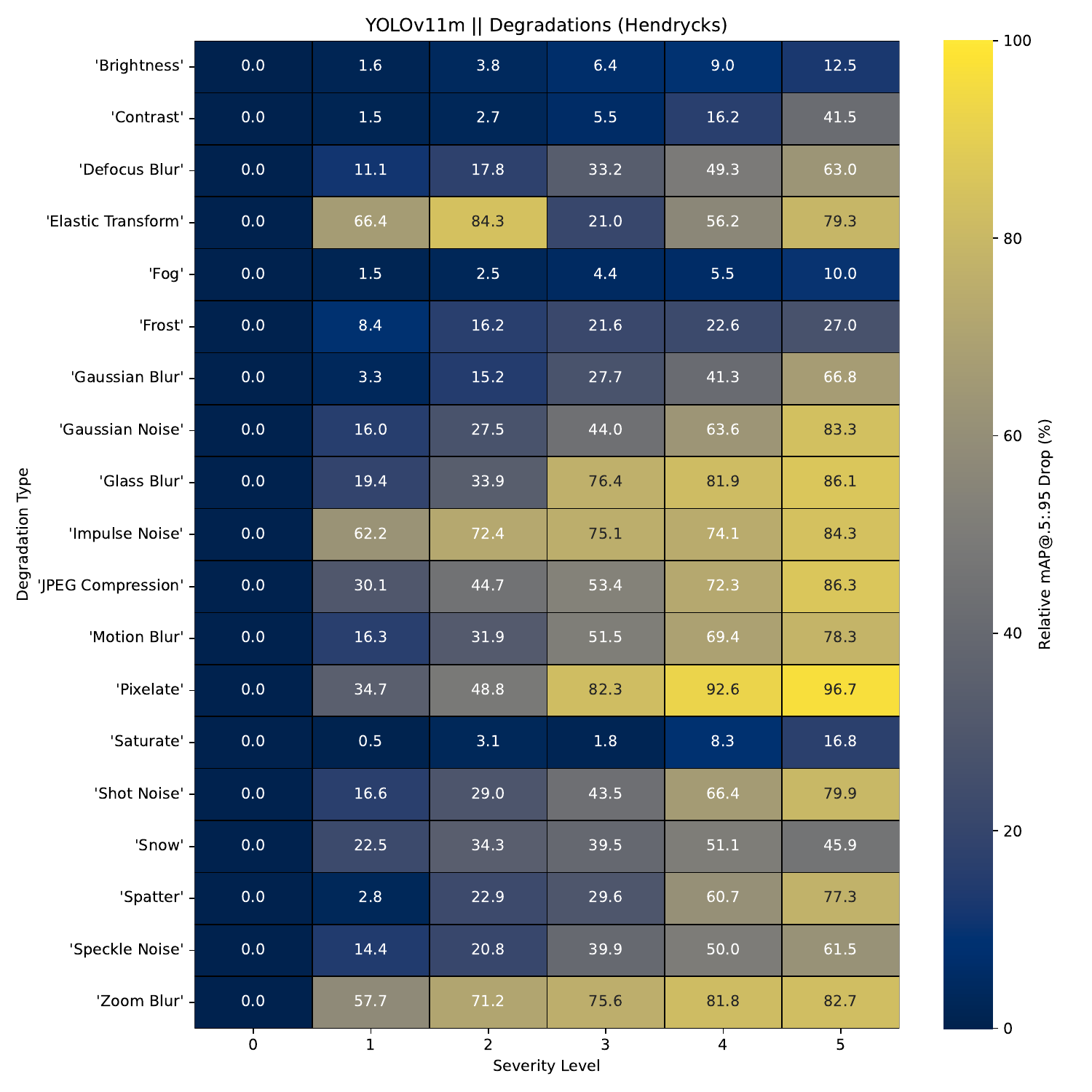}
		\caption*{Degradations from~\cite{Hendrycks_ICLR_2019}}
	\end{subfigure}
	\hfill
	\begin{subfigure}{\subw}
		\centering
		\includegraphics[width=\textwidth]{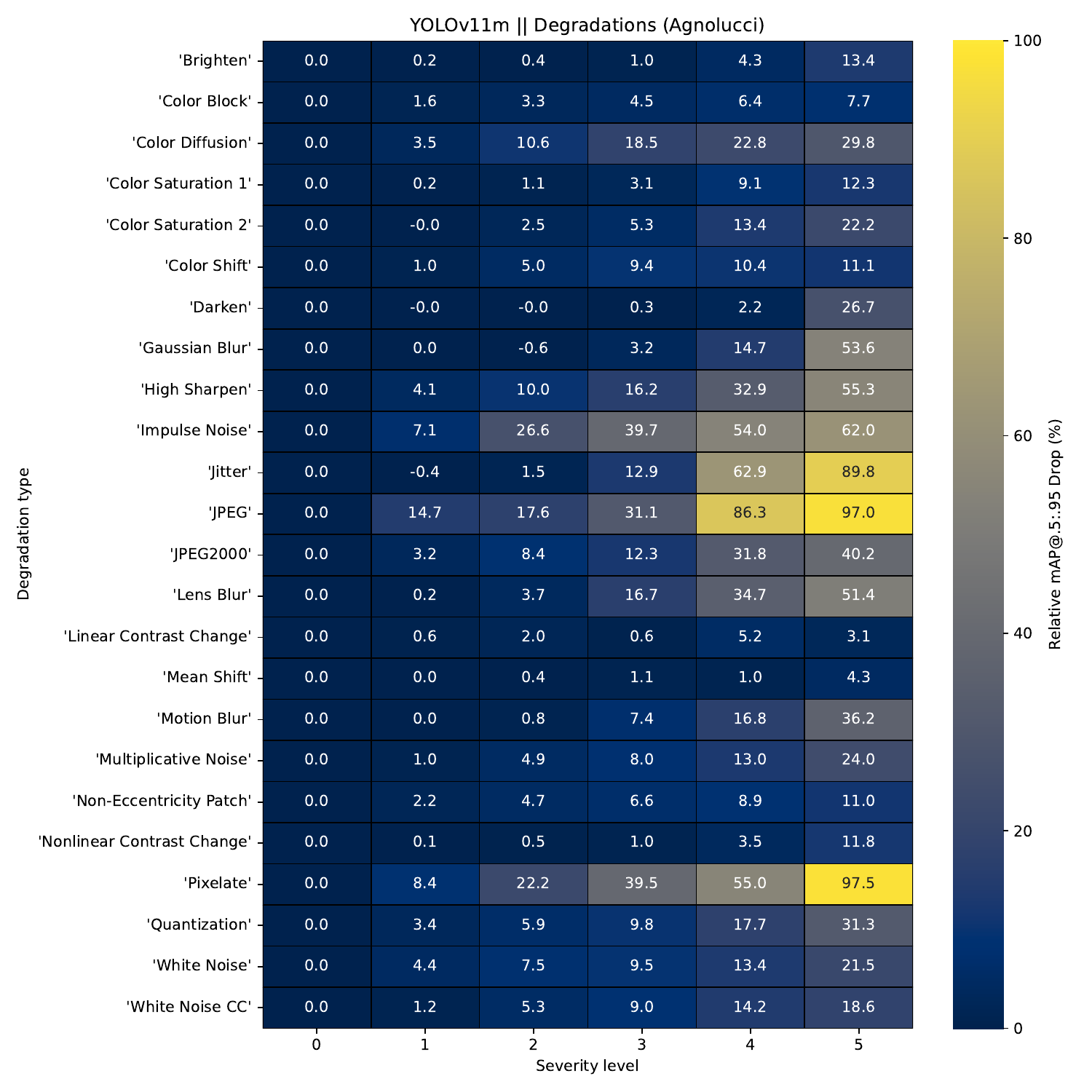}
		\caption*{Degradations from~\cite{Agnolucci_WACV_2024}}
	\end{subfigure}
	\hfill
	\begin{subfigure}{\subw}
		\centering
		\includegraphics[width=\textwidth]{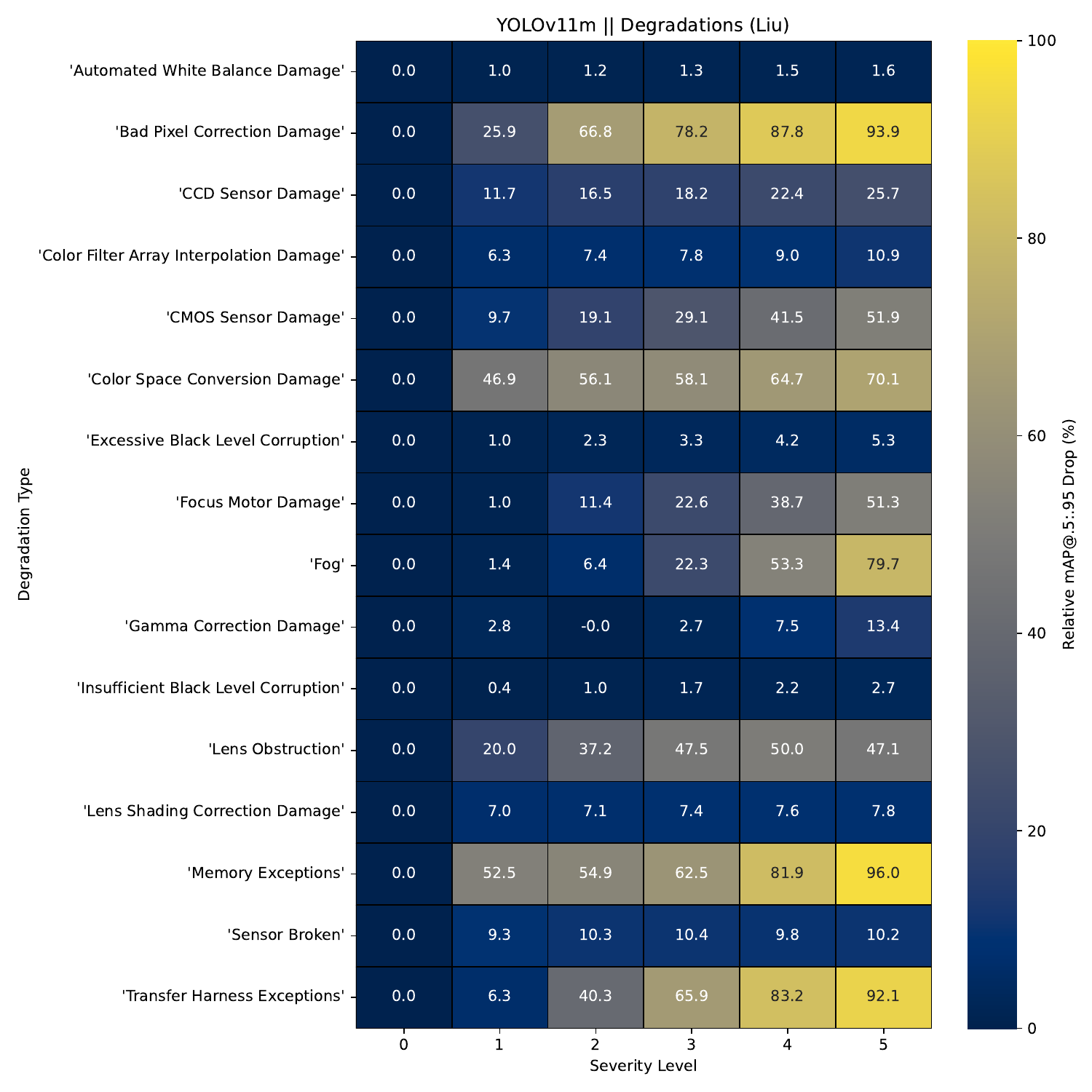}
		\caption*{Degradations from~\cite{Liu_IJCV_2024}}
	\end{subfigure}
	\caption{
		Relative detection performance drop of YOLO11-m~\cite{yolo11_ultralytics} on COCO \texttt{val2017} under increasing native severity levels (mAP@0.5:0.95).
	}
	\label{fig:map_vs_severity_all_YOLOv11-m}
\end{figure*}

\renewcommand{\subw}{0.3\textwidth}
\begin{figure*}[h!]
	\captionsetup[subfigure]{justification=centering}
	\centering
	\begin{subfigure}{\subw}
		\centering
		\includegraphics[width=\textwidth]{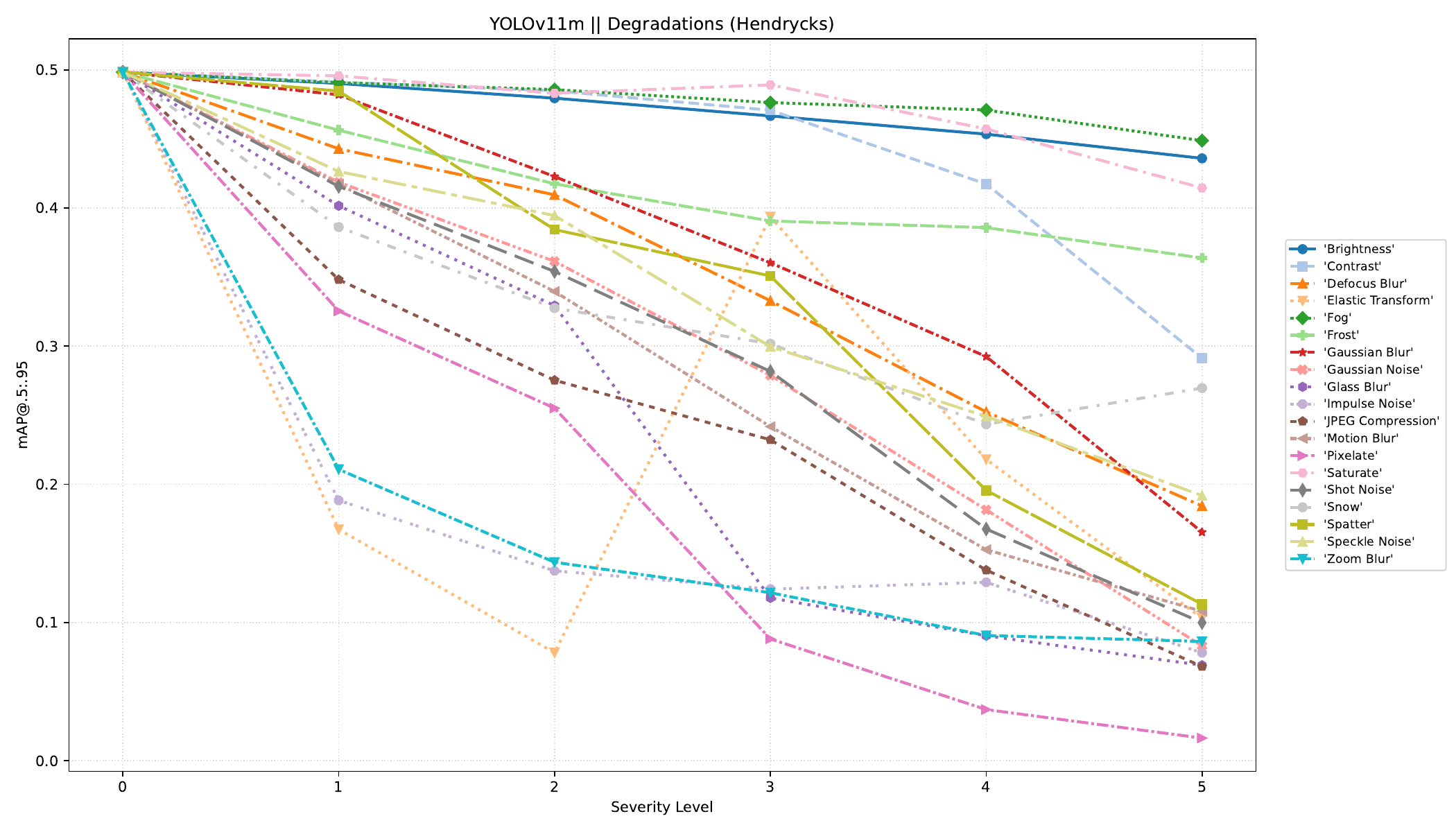}
		\caption*{Degradations from~\cite{Hendrycks_ICLR_2019}}
	\end{subfigure}
	\hfill
	\begin{subfigure}{\subw}
		\centering
		\includegraphics[width=\textwidth]{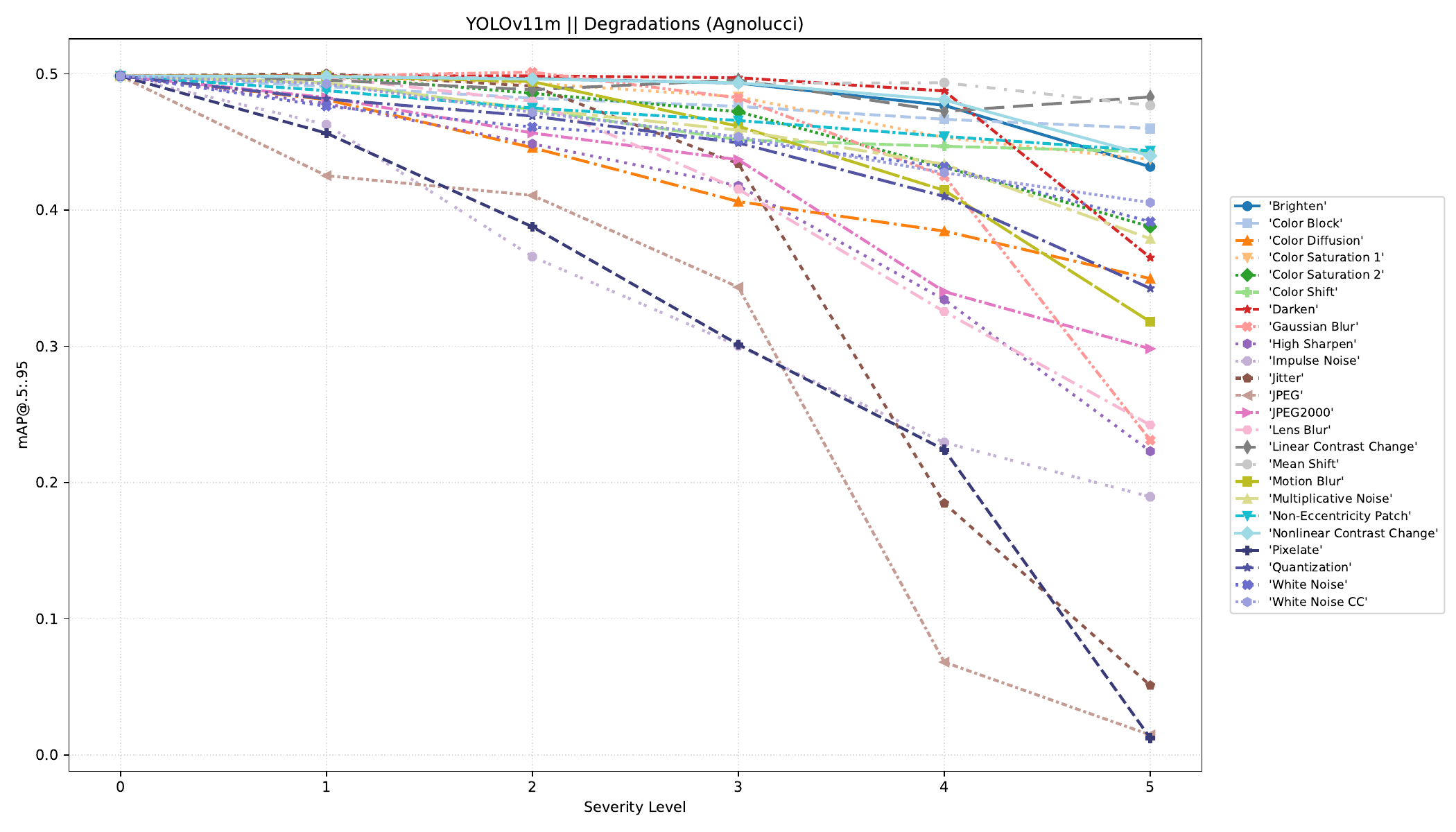}
		\caption*{Degradations from~\cite{Agnolucci_WACV_2024}}
	\end{subfigure}
	\hfill
	\begin{subfigure}{\subw}
		\centering
		\includegraphics[width=\textwidth]{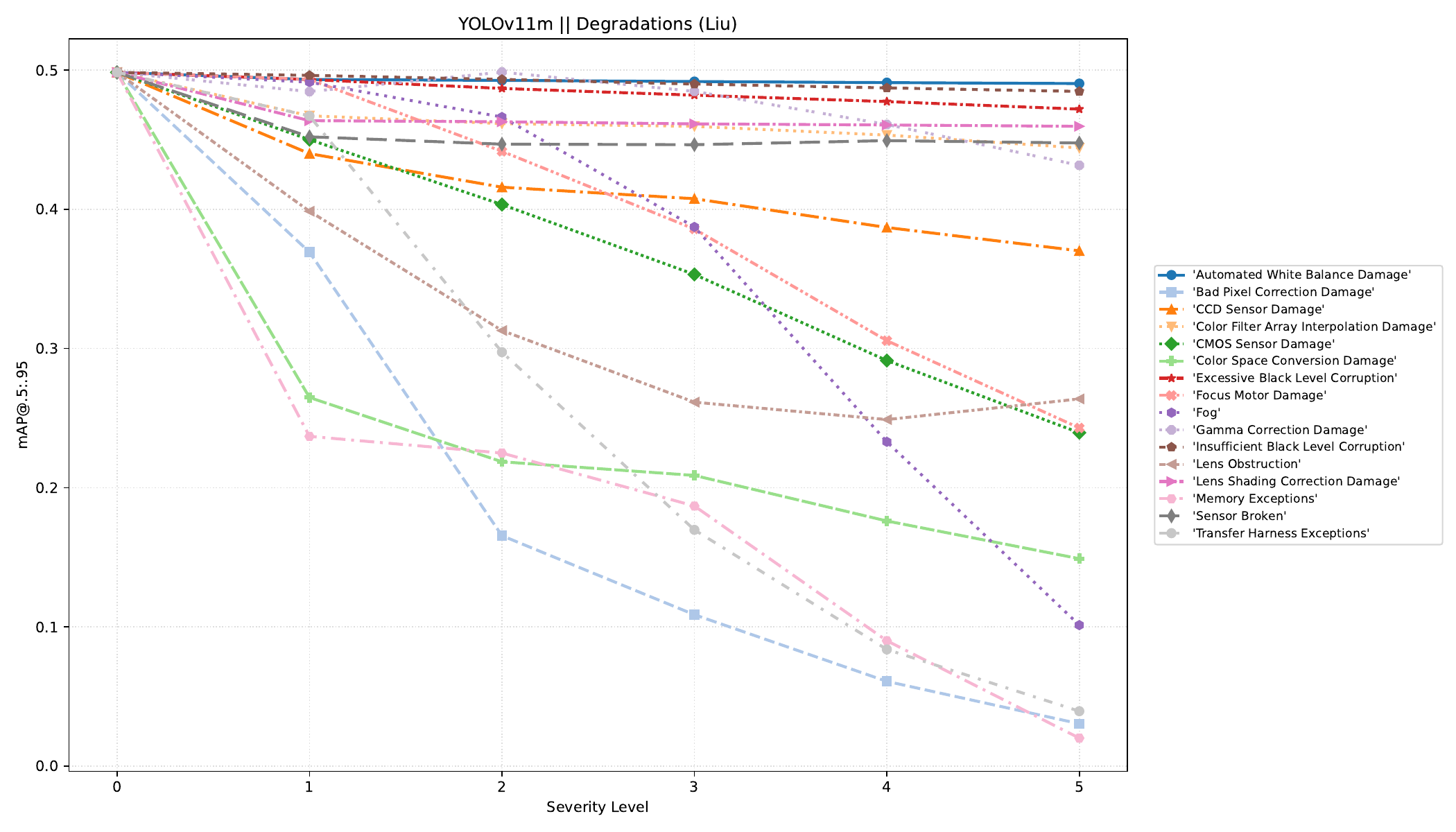}
		\caption*{Degradations from~\cite{Liu_IJCV_2024}}
	\end{subfigure}
	\caption{
		Absolute detection performance of YOLO11-m~\cite{yolo11_ultralytics} on COCO \texttt{val2017} across native severity levels (mAP@0.5:0.95).
	}
	\label{fig:map_vs_severity_all_absolute_yolov11m}
\end{figure*}

\renewcommand{\subw}{0.3\textwidth}
\begin{figure*}[h!]
	\captionsetup[subfigure]{justification=centering}
	\centering
	\begin{subfigure}{\subw}
		\centering
		\includegraphics[width=\textwidth]{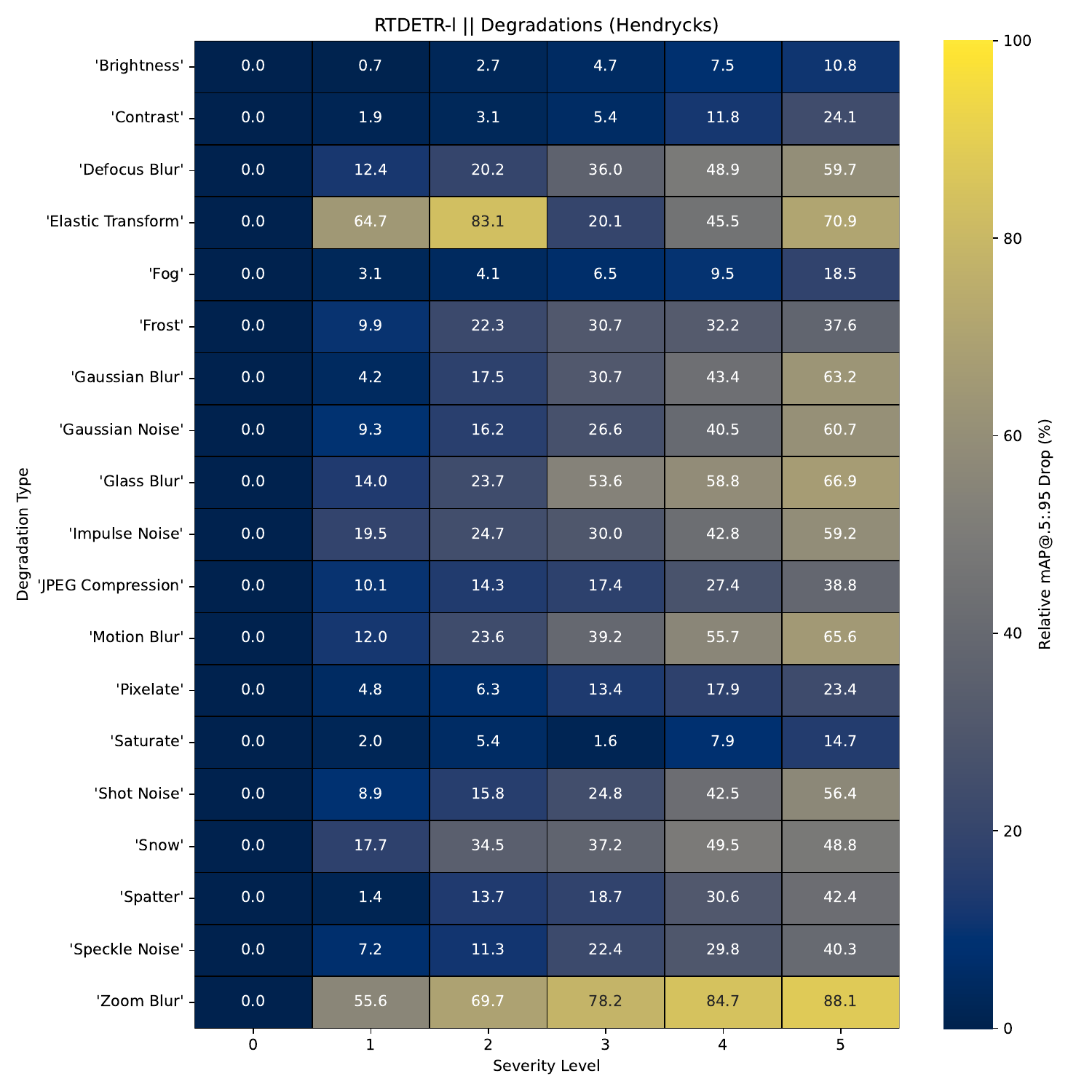}
		\caption*{Degradations from~\cite{Hendrycks_ICLR_2019}}
	\end{subfigure}
	\hfill
	\begin{subfigure}{\subw}
		\centering
		\includegraphics[width=\textwidth]{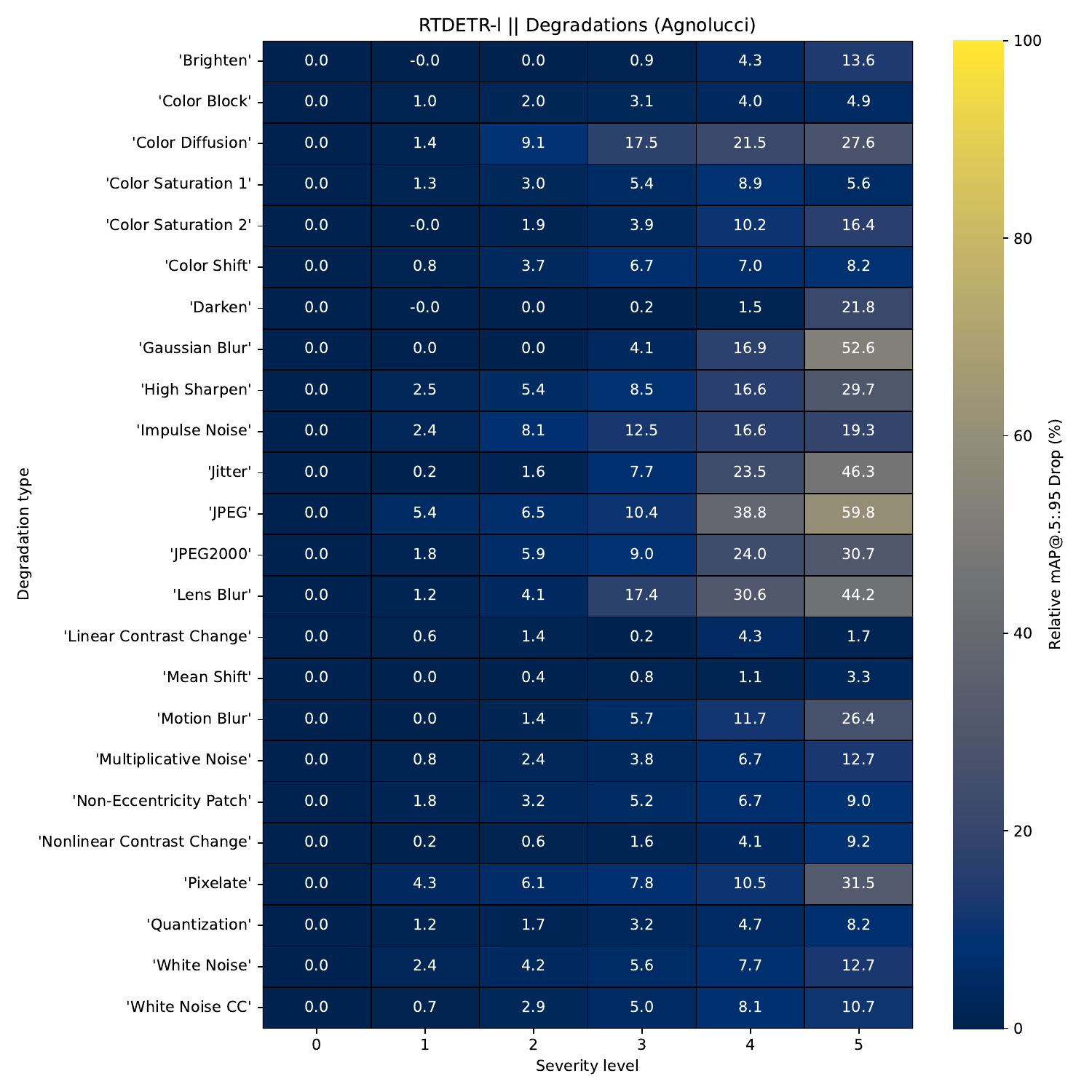}
		\caption*{Degradations from~\cite{Agnolucci_WACV_2024}}
	\end{subfigure}
	\hfill
	\begin{subfigure}{\subw}
		\centering
		\includegraphics[width=\textwidth]{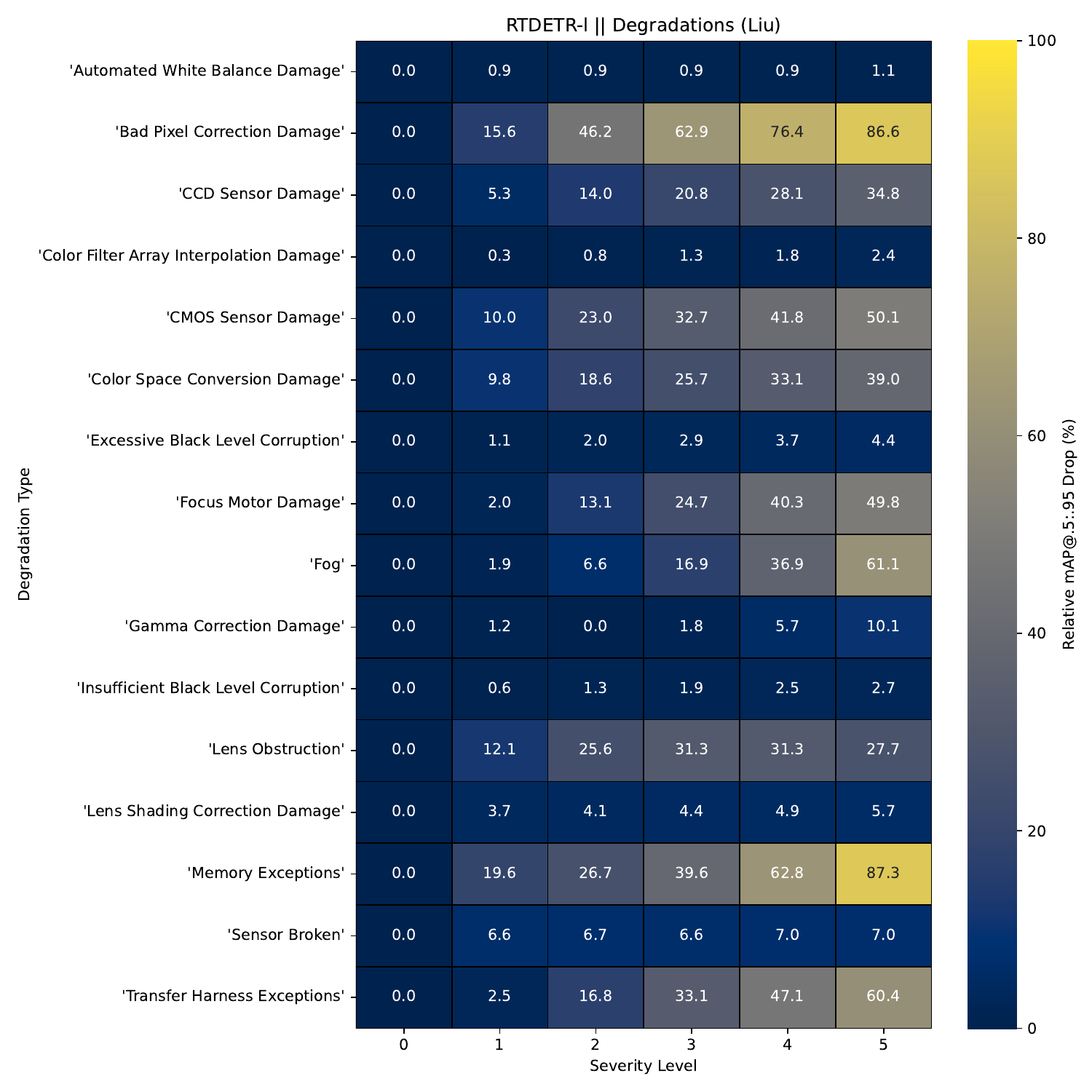}
		\caption*{Degradations from~\cite{Liu_IJCV_2024}}
	\end{subfigure}
	\caption{
		Relative detection performance drop of RT-DETR-l~\cite{Zhao_2024_CVPR} on COCO \texttt{val2017} under increasing native severity levels (mAP@0.5:0.95).
	}
	\label{fig:map_vs_severity_all_RTDETR}
\end{figure*}

\renewcommand{\subw}{0.3\textwidth}
\begin{figure*}[h!]
	\captionsetup[subfigure]{justification=centering}
	\centering
	\begin{subfigure}{\subw}
		\centering
		\includegraphics[width=\textwidth]{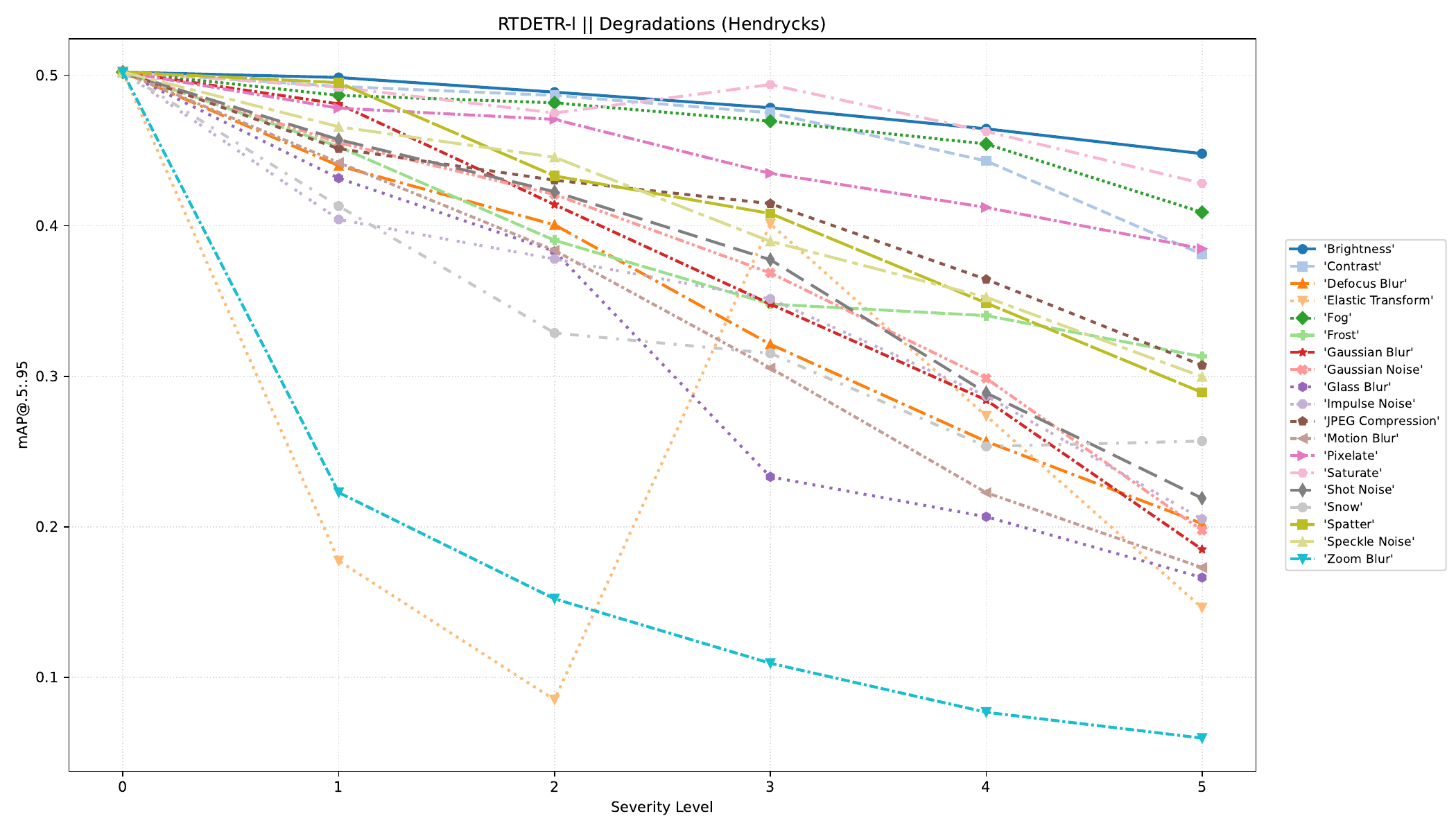}
		\caption*{Degradations from~\cite{Hendrycks_ICLR_2019}}
	\end{subfigure}
	\hfill
	\begin{subfigure}{\subw}
		\centering
		\includegraphics[width=\textwidth]{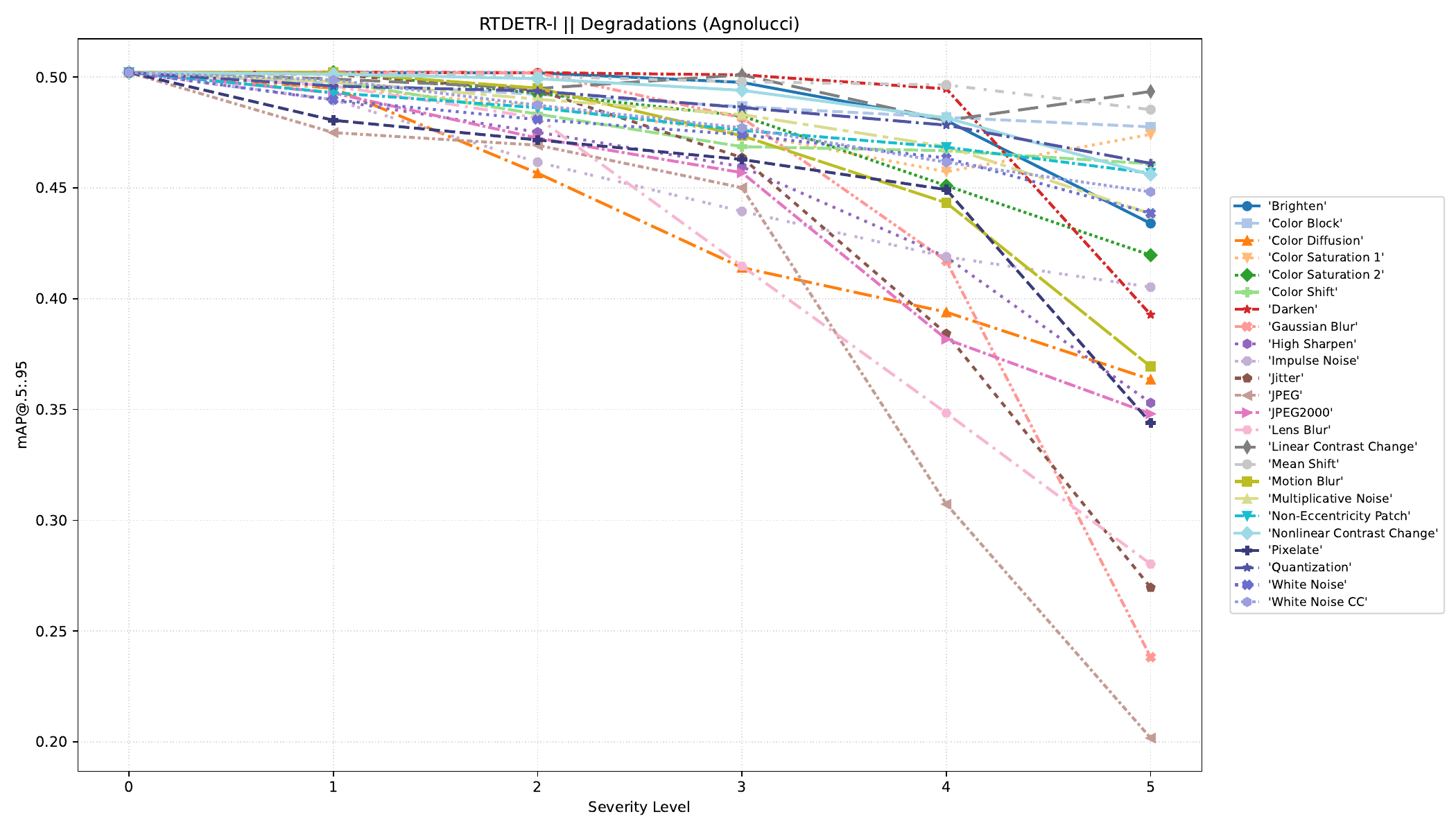}
		\caption*{Degradations from~\cite{Agnolucci_WACV_2024}}
	\end{subfigure}
	\hfill
	\begin{subfigure}{\subw}
		\centering
		\includegraphics[width=\textwidth]{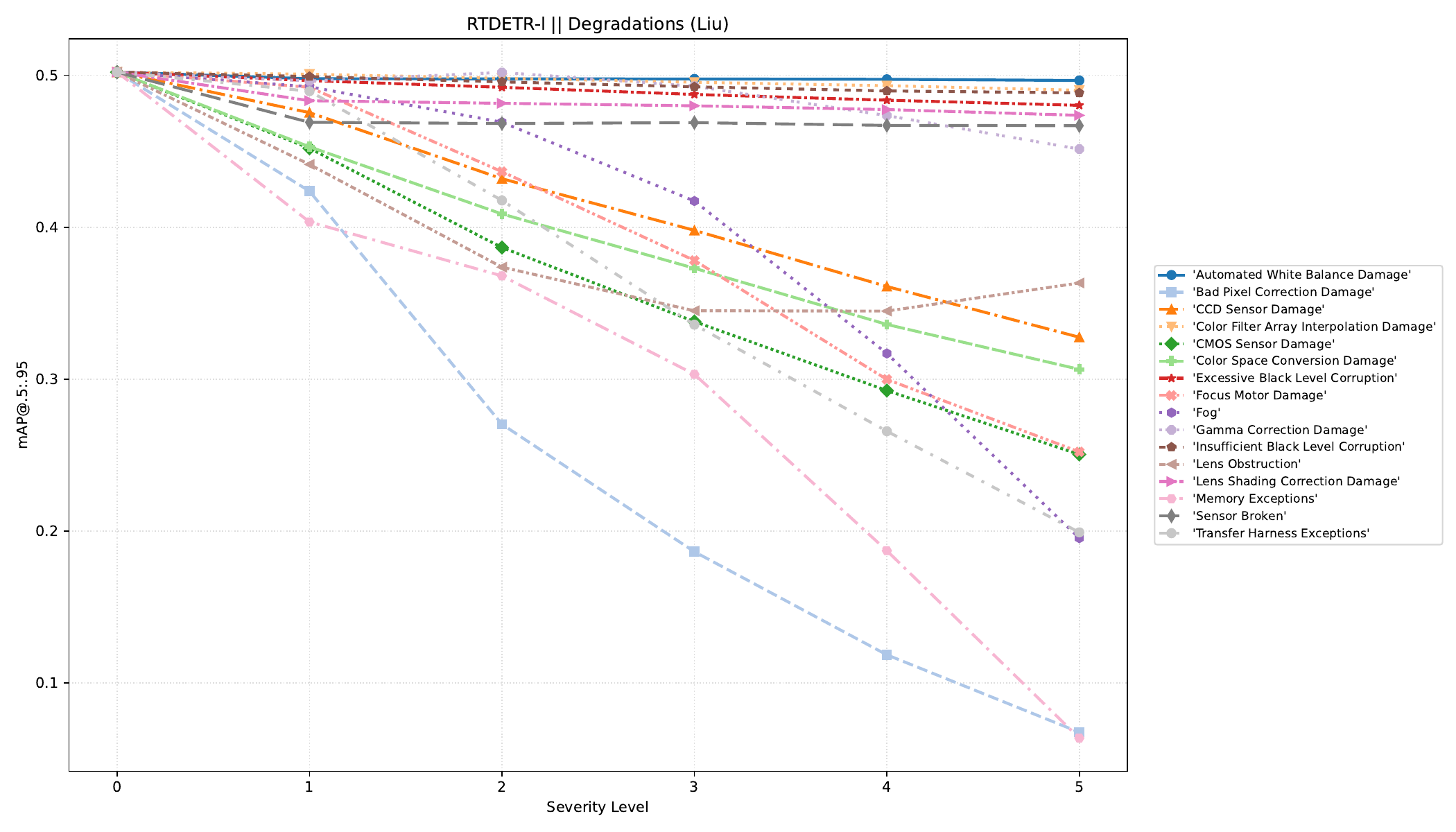}
		\caption*{Degradations from~\cite{Liu_IJCV_2024}}
	\end{subfigure}
	\caption{
		Absolute detection performance of RT-DETR-l~\cite{Zhao_2024_CVPR} on COCO \texttt{val2017} across native severity levels (mAP@0.5:0.95).
	}
	\label{fig:map_vs_severity_all_absolute_rtdetr_l}
\end{figure*}

\clearpage
\phantomsection
\section*{Supplementary Evaluation: Severity Tables}
\addcontentsline{toc}{subsection}{Supplementary Evaluation: Severity Tables}
\label{app:severity_tables}

The main paper reports a full severity table for the ImageNet-C style degradation family as a representative example. This section provides the remaining full per-backend severity tables for the IQA-motivated degradations of~\cite{Agnolucci_WACV_2024} and the real-camera or system-level degradations of~\cite{Liu_IJCV_2024}. As in the main paper, degradation strength is reported using $(1-\mathrm{SSIM})$, PSNR, and LPIPS for native severity levels 1--5.

Higher values of $(1-\mathrm{SSIM})$ and LPIPS indicate stronger degradation, whereas lower PSNR indicates stronger degradation. Entries with PSNR~=~50\,dB or zero-valued $(1-\mathrm{SSIM})$/LPIPS correspond to identity-like or metric-saturated cases under the respective metric.

\begin{table*}[h]
	\centering
	\scriptsize
	\setlength{\tabcolsep}{2pt}
	\resizebox{\textwidth}{!}{%
		\begin{tabular}{l|c:c:c:c:c|c:c:c:c:c|c:c:c:c:c|}
			\hline
			& \multicolumn{5}{c|}{(1-SSIM)}
			& \multicolumn{5}{c|}{PSNR / dB}
			& \multicolumn{5}{c|}{LPIPS} \\
			& \multicolumn{5}{c|}{Severity Level}
			& \multicolumn{5}{c|}{Severity Level}
			& \multicolumn{5}{c|}{Severity Level} \\
			Degradation Function \cite{Agnolucci_WACV_2024}
			& 1 & 2 & 3 & 4 & 5
			& 1 & 2 & 3 & 4 & 5
			& 1 & 2 & 3 & 4 & 5 \\
			\hline
			\rowcolor{Set3-08!20}
			'Gaussian Blur'
			& 0.000 & 0.009 & 0.081 & 0.202 & 0.393
			& 50.000 & 35.615 & 27.395 & 24.242 & 21.347
			& 0.000 & 0.026 & 0.161 & 0.300 & 0.491 \\
			\rowcolor{gray!20}
			'Lens Blur'
			& 0.034 & 0.101 & 0.236 & 0.332 & 0.395
			& 30.496 & 26.427 & 23.687 & 22.284 & 21.461
			& 0.082 & 0.167 & 0.293 & 0.351 & 0.408 \\
			\rowcolor{Set3-08!20}
			'Motion Blur'
			& 0.000 & 0.021 & 0.075 & 0.132 & 0.220
			& 50.000 & 32.440 & 27.712 & 25.797 & 23.790
			& 0.000 & 0.058 & 0.146 & 0.210 & 0.302 \\
			\rowcolor{gray!20}
			'White Noise'
			& 0.110 & 0.178 & 0.229 & 0.301 & 0.410
			& 30.206 & 27.255 & 25.541 & 23.397 & 20.528
			& 0.158 & 0.217 & 0.255 & 0.306 & 0.380 \\
			\rowcolor{Set3-08!20}
			'White Noise in Color Component'
			& 0.040 & 0.141 & 0.219 & 0.316 & 0.378
			& 35.509 & 28.625 & 25.694 & 22.798 & 21.126
			& 0.073 & 0.180 & 0.241 & 0.309 & 0.352 \\
			\rowcolor{gray!20}
			'Impulse Noise'
			& 0.000 & 0.000 & 0.000 & 0.000 & 0.000
			& 50.000 & 50.000 & 50.000 & 50.000 & 50.000
			& 0.000 & 0.000 & 0.000 & 0.000 & 0.000 \\
			\rowcolor{Set3-08!20}
			'Multiplicative Noise'
			& 0.006 & 0.027 & 0.050 & 0.087 & 0.168
			& 36.267 & 29.374 & 26.430 & 23.510 & 19.696
			& 0.007 & 0.029 & 0.050 & 0.081 & 0.142 \\
			\rowcolor{gray!20}
			'Brighten'
			& 0.008 & 0.026 & 0.085 & 0.205 & 0.343
			& 29.277 & 23.496 & 17.636 & 12.990 & 10.185
			& 0.004 & 0.014 & 0.054 & 0.141 & 0.240 \\
			\rowcolor{Set3-08!20}
			'Darken'
			& 0.002 & 0.008 & 0.033 & 0.149 & 0.490
			& 36.043 & 30.022 & 24.002 & 17.981 & 12.368
			& 0.001 & 0.005 & 0.020 & 0.074 & 0.232 \\
			\rowcolor{gray!20}
			'Mean Shift'
			& 0.000 & 0.047 & 0.068 & 0.106 & 0.172
			& 50.000 & 22.629 & 22.815 & 17.268 & 17.636
			& 0.000 & 0.014 & 0.019 & 0.040 & 0.056 \\
			\rowcolor{Set3-08!20}
			'Color Diffusion'
			& 0.361 & 0.426 & 0.505 & 0.542 & 0.596
			& 11.914 & 11.375 & 10.211 & 9.603 & 8.629
			& 0.222 & 0.336 & 0.419 & 0.454 & 0.508 \\
			\rowcolor{gray!20}
			'Color Shift'
			& 0.000 & 0.000 & 0.000 & 0.000 & 0.000
			& 50.000 & 50.000 & 50.000 & 50.000 & 50.000
			& 0.000 & 0.000 & 0.000 & 0.000 & 0.000 \\
			\rowcolor{Set3-08!20}
			'Color Saturation 1'
			& 0.287 & 0.414 & 0.478 & 0.541 & 0.673
			& 14.560 & 12.061 & 11.038 & 10.123 & 8.936
			& 0.209 & 0.327 & 0.405 & 0.485 & 0.509 \\
			\rowcolor{gray!20}
			'Color Saturation 2'
			& 0.000 & 0.209 & 0.337 & 0.503 & 0.592
			& 50.000 & 18.287 & 15.293 & 12.209 & 10.618
			& 0.000 & 0.136 & 0.219 & 0.326 & 0.387 \\
			\rowcolor{Set3-08!20}
			'JPEG'
			& 0.802 & 0.888 & 0.906 & 0.944 & 0.990
			& 4.914 & 4.858 & 4.953 & 2.643 & 0.489
			& 0.258 & 0.644 & 0.661 & 0.734 & 0.591 \\
			\rowcolor{gray!20}
			'JPEG2000'
			& 0.290 & 0.388 & 0.426 & 0.524 & 0.552
			& 15.281 & 14.010 & 13.638 & 13.079 & 12.976
			& 0.273 & 0.346 & 0.379 & 0.471 & 0.502 \\
			\rowcolor{Set3-08!20}
			'Jitter'
			& 0.580 & 0.580 & 0.580 & 0.580 & 0.580
			& 10.800 & 10.800 & 10.793 & 10.800 & 10.800
			& 0.518 & 0.518 & 0.518 & 0.518 & 0.519 \\
			\rowcolor{gray!20}
			'Non-Eccentricity Patch'
			& 0.000 & 0.000 & 0.000 & 0.000 & 0.000
			& 50.000 & 50.000 & 50.000 & 50.000 & 50.000
			& 0.000 & 0.000 & 0.000 & 0.000 & 0.000 \\
			\rowcolor{Set3-08!20}
			'Pixelate'
			& 0.529 & 0.552 & 0.567 & 0.585 & 0.599
			& 8.508 & 8.506 & 8.506 & 8.505 & 8.498
			& 0.196 & 0.268 & 0.333 & 0.426 & 0.547 \\
			\rowcolor{gray!20}
			'Quantization'
			& 0.000 & 0.000 & 0.000 & 0.000 & 0.000
			& 50.000 & 50.000 & 50.000 & 50.000 & 50.000
			& 0.000 & 0.000 & 0.000 & 0.000 & 0.000 \\
			\rowcolor{Set3-08!20}
			'Color Block'
			& 0.000 & 0.000 & 0.000 & 0.000 & 0.000
			& 50.000 & 50.000 & 50.000 & 50.000 & 50.000
			& 0.000 & 0.000 & 0.000 & 0.000 & 0.000 \\
			\rowcolor{gray!20}
			'High Sharpen'
			& 0.338 & 0.418 & 0.486 & 0.634 & 0.751
			& 10.512 & 9.424 & 8.692 & 7.347 & 6.170
			& 0.314 & 0.361 & 0.395 & 0.461 & 0.525 \\
			\rowcolor{Set3-08!20}
			'Linear Contrast Change'
			& 0.996 & 0.997 & 0.002 & 0.998 & 0.007
			& 0.900 & 0.898 & 37.366 & 0.895 & 31.346
			& 0.602 & 0.603 & 0.091 & 0.603 & 0.087 \\
			\rowcolor{gray!20}
			'Nonlinear Contrast Change'
			& 0.024 & 0.099 & 0.241 & 0.446 & 0.564
			& 20.336 & 14.315 & 10.794 & 8.295 & 7.272
			& 0.158 & 0.253 & 0.317 & 0.411 & 0.521 \\
			\bottomrule
		\end{tabular}
	}
	\caption{
		Measured degradation strengths for COCO images under IQA-motivated degradations from~\cite{Agnolucci_WACV_2024}. Columns report $(1-\mathrm{SSIM})$, PSNR (dB), and LPIPS for native severity levels 1--5.
	}
	\label{tab:degradation_strengths_all_metrics_agnolucci}
\end{table*}

\begin{table}[h]
	\centering
	\scriptsize
	\setlength{\tabcolsep}{2pt}
	\resizebox{\textwidth}{!}{%
		\begin{tabular}{l|c:c:c:c:c|c:c:c:c:c|c:c:c:c:c|}
			\hline
			& \multicolumn{5}{c|}{(1-SSIM)}
			& \multicolumn{5}{c|}{PSNR / dB}
			& \multicolumn{5}{c|}{LPIPS} \\
			& \multicolumn{5}{c|}{Severity Level}
			& \multicolumn{5}{c|}{Severity Level}
			& \multicolumn{5}{c|}{Severity Level} \\
			Degradation Function \cite{Liu_IJCV_2024}
			& 1 & 2 & 3 & 4 & 5
			& 1 & 2 & 3 & 4 & 5
			& 1 & 2 & 3 & 4 & 5 \\
			\hline
			\rowcolor{Set3-08!20}
			\degfunc{Fog}
			& 0.296 & 0.473 & 0.564 & 0.608 & 0.630
			& 13.026 & 10.540 & 10.097 & 10.282 & 10.673
			& 0.155 & 0.326 & 0.481 & 0.608 & 0.690 \\
			\rowcolor{gray!20}
			\degfunc{Lens Obstruction}
			& 0.179 & 0.240 & 0.271 & 0.279 & 0.273
			& 15.886 & 15.392 & 15.223 & 15.246 & 15.373
			& 0.118 & 0.189 & 0.222 & 0.232 & 0.226 \\
			\rowcolor{Set3-08!20}
			\degfunc{Focus Motor Damage}
			& 0.050 & 0.175 & 0.262 & 0.346 & 0.389
			& 29.557 & 25.152 & 23.598 & 22.350 & 21.723
			& 0.115 & 0.282 & 0.371 & 0.453 & 0.497 \\
			\rowcolor{gray!20}
			\degfunc{CCD Sensor Damage}
			& 0.067 & 0.170 & 0.245 & 0.318 & 0.379
			& 24.897 & 20.442 & 18.494 & 17.056 & 15.964
			& 0.074 & 0.174 & 0.237 & 0.291 & 0.333 \\
			\rowcolor{Set3-08!20}
			\degfunc{CMOS Sensor Damage}
			& 0.113 & 0.266 & 0.363 & 0.443 & 0.504
			& 23.054 & 18.700 & 16.762 & 15.347 & 14.337
			& 0.147 & 0.296 & 0.372 & 0.430 & 0.473 \\
			\rowcolor{gray!20}
			\degfunc{Insufficient Black Level Corruption}
			& 0.087 & 0.092 & 0.100 & 0.108 & 0.117
			& 17.585 & 17.839 & 17.548 & 16.846 & 15.882
			& 0.124 & 0.197 & 0.255 & 0.299 & 0.333 \\
			\rowcolor{Set3-08!20}
			\degfunc{Excessive Black Level Corruption}
			& 0.088 & 0.093 & 0.102 & 0.111 & 0.121
			& 17.456 & 17.611 & 17.262 & 16.548 & 15.602
			& 0.205 & 0.318 & 0.390 & 0.438 & 0.473 \\
			\rowcolor{gray!20}
			\degfunc{Lens Shading Correction Damage}
			& 0.151 & 0.177 & 0.225 & 0.290 & 0.363
			& 24.027 & 19.353 & 15.841 & 13.478 & 11.838
			& 0.151 & 0.179 & 0.209 & 0.237 & 0.263 \\
			\rowcolor{Set3-08!20}
			\degfunc{Automated White Balance Damage}
			& 0.015 & 0.013 & 0.013 & 0.013 & 0.014
			& 29.918 & 29.576 & 28.948 & 28.148 & 27.279
			& 0.066 & 0.071 & 0.080 & 0.092 & 0.104 \\
			\rowcolor{gray!20}
			\degfunc{Bad Pixel Correction Damage}
			& 0.243 & 0.606 & 0.706 & 0.775 & 0.836
			& 23.513 & 15.574 & 12.596 & 10.600 & 9.065
			& 0.251 & 0.519 & 0.593 & 0.633 & 0.658 \\
			\rowcolor{Set3-08!20}
			\degfunc{Color Filter Array Interpolation Damage}
			& 0.033 & 0.077 & 0.099 & 0.112 & 0.119
			& 36.219 & 30.628 & 28.294 & 26.793 & 25.689
			& 0.019 & 0.053 & 0.080 & 0.106 & 0.132 \\
			\rowcolor{gray!20}
			\degfunc{Gamma Correction Damage}
			& 0.177 & 0.013 & 0.244 & 0.443 & 0.571
			& 14.540 & 29.784 & 16.592 & 13.187 & 11.480
			& 0.068 & 0.005 & 0.073 & 0.152 & 0.221 \\
			\rowcolor{Set3-08!20}
			\degfunc{Color Space Conversion Damage}
			& 0.367 & 0.592 & 0.689 & 0.751 & 0.794
			& 19.706 & 13.966 & 11.577 & 10.044 & 8.914
			& 0.259 & 0.502 & 0.627 & 0.705 & 0.757 \\
			\rowcolor{gray!20}
			\degfunc{Sensor Broken}
			& 0.069 & 0.069 & 0.069 & 0.070 & 0.068
			& 19.120 & 19.054 & 19.093 & 19.073 & 19.094
			& 0.123 & 0.122 & 0.122 & 0.122 & 0.123 \\
			\rowcolor{Set3-08!20}
			\degfunc{Memory Exceptions}
			& 0.377 & 0.478 & 0.590 & 0.740 & 0.887
			& 17.843 & 14.966 & 12.022 & 9.045 & 7.119
			& 0.362 & 0.453 & 0.543 & 0.623 & 0.669 \\
			\rowcolor{gray!20}
			\degfunc{Transfer Harness Exceptions}
			& 0.031 & 0.218 & 0.409 & 0.553 & 0.659
			& 32.926 & 22.635 & 18.062 & 15.086 & 12.864
			& 0.031 & 0.194 & 0.353 & 0.484 & 0.585 \\
			\bottomrule
		\end{tabular}
	}
	\caption{
		Measured degradation strengths for COCO images under real-camera or system-level degradations from~\cite{Liu_IJCV_2024}. Columns report $(1-\mathrm{SSIM})$, PSNR (dB), and LPIPS for native severity levels 1--5.
	}
	\label{tab:degradation_strengths_all_metrics_liu}
\end{table}

These full per-backend tables complete the measurement layer used in the main paper. The following two sections summarize how the same metric-based representation can be extended beyond the default native benchmark range and reused across different degradation-application protocols.

\clearpage
\phantomsection
\section*{Supplementary Evaluation: Extrapolated Severity Levels}
\addcontentsline{toc}{subsection}{Supplementary Evaluation: Extrapolated Severity Levels}
\label{app:extrapolation}

The main paper evaluates degradations at their native backend severity levels and characterizes them through explicit metric-based measurements. In some cases, however, the standard five-level native range does not fully probe the robustness limits of modern detectors. Empirically, models may remain relatively stable over the default benchmark range but degrade sharply beyond the strongest native level. To study such regimes, the \texttt{imdeg} framework supports an extrapolation of the canonical severity axis in distortion space.

Let $L_{\deg,k}$ denote canonical severity levels derived from measured degradation strengths for a degradation type $\degfunc{deg}$. Additional target levels can be defined by fixed-step continuation beyond the largest canonical level:
\begin{equation}
	L_{\deg,K+m} = L_{\deg,K} + m \cdot \Delta_{\deg},
	\qquad
	\Delta_{\deg} = L_{\deg,k+1} - L_{\deg,k},
\end{equation}
where $K$ is the largest canonical level used in the benchmark and $m \geq 1$ indexes extrapolated levels. This construction extends severity in metric space rather than in backend parameter space.

\begin{figure*}[t]
	\captionsetup[subfigure]{justification=centering}
	\centering
	\begin{subfigure}{0.45\textwidth}
		\centering
		\includegraphics[width=\linewidth]{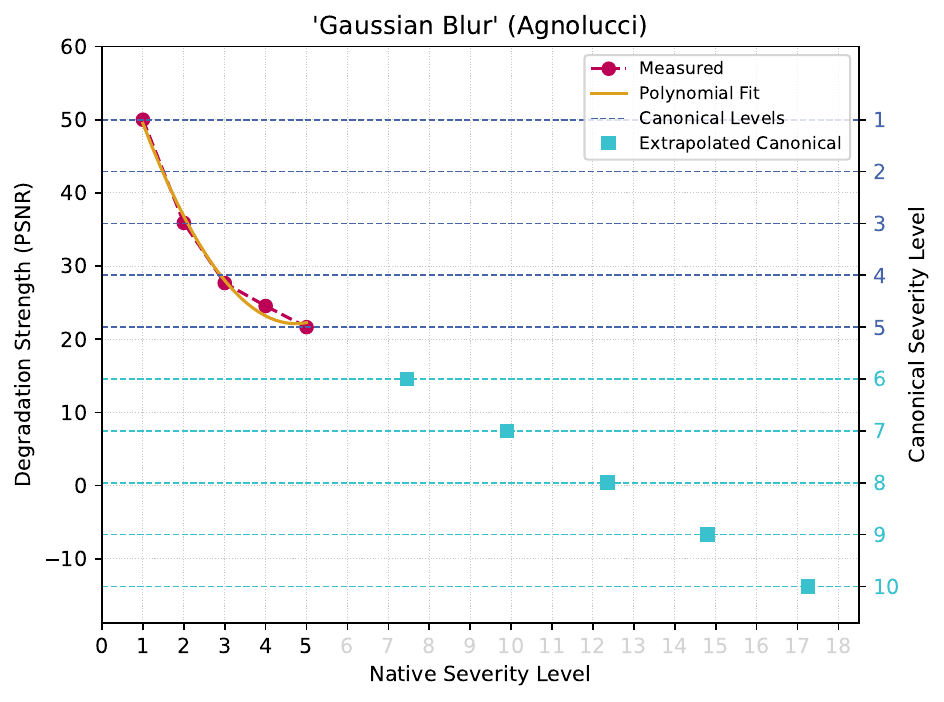}
		\caption*{PSNR-based severity extrapolation}
	\end{subfigure}
	\hfill
	\begin{subfigure}{0.45\textwidth}
		\centering
		\includegraphics[width=\linewidth]{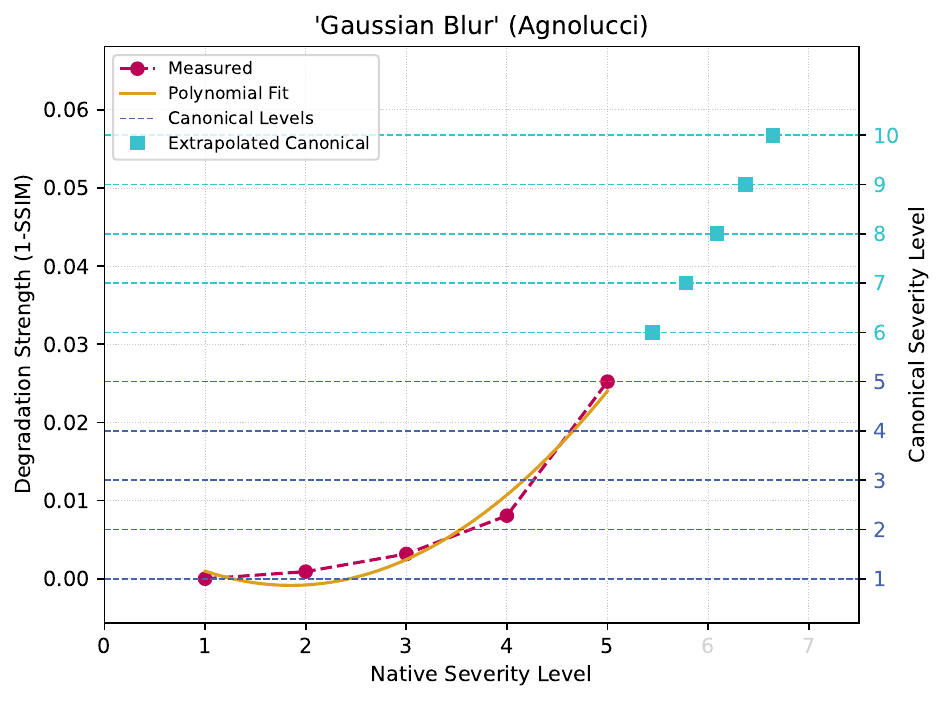}
		\caption*{$(1-\mathrm{SSIM})$-based severity extrapolation}
	\end{subfigure}
	\caption{
		Extension of canonical severity levels for \degfunc{Gaussian Blur} using fixed-step extrapolation in distortion space. Red dashed lines indicate extrapolated targets $L_{6}$--$L_{10}$. Due to backend constraints, these levels saturate in parameter space at the maximum supported native severity. Compositional application can approximate stronger effects but is used only for diagnostic visualization.
	}
	\label{fig:extrapolated_levels}
\end{figure*}

Figure~\ref{fig:extrapolated_levels} illustrates this procedure for \degfunc{Gaussian Blur} using both PSNR and $(1-\mathrm{SSIM})$. In practice, most backends expose only a small set of discrete native severity levels, typically $l \in \{1,\dots,5\}$. As a result, extrapolated targets often saturate at the strongest available native implementation. In such cases, the extrapolated level should be interpreted as indicating insufficient backend coverage of the stronger distortion regime, rather than as a failure of the measurement framework.

Extrapolated severities are therefore used only as a diagnostic analysis tool. They are useful for visualizing robustness saturation, identifying backends whose native range is too limited, and motivating future extensions of degradation parameterizations. All benchmark results reported in the main paper remain based on native severity schedules.

As an optional approximation, stronger effects can be synthesized through sequential composition of native degradations, for example by applying severity~5 followed by an additional lower-severity application. Such compositions are not part of the standard benchmark definition and should be reported separately, since they no longer correspond to the original native backend semantics.

\clearpage
\phantomsection
\section*{Supplementary Evaluation: Degradation Application Protocols}
\addcontentsline{toc}{subsection}{Supplementary Evaluation: Degradation Application Protocols}
\label{app:protocols}

The proposed framework separates \emph{degradation definition}, \emph{taxonomy annotation}, and \emph{severity measurement}. This separation makes it possible to apply the same interpretive layer across different dataset construction strategies. While the main paper focuses on the standard single-degradation setting, the \texttt{imdeg} framework supports several commonly used degradation application protocols.

\smallsec{Protocol A: Round--Robin Assignment.}
Each image receives exactly one degradation type and one severity level, assigned deterministically in a cyclic or balanced manner~\cite{Oksuz_CVPR_2023}. This protocol ensures broad coverage while avoiding the combinatorial growth of exhaustively evaluating every degradation on every image.

\smallsec{Protocol B: Cartesian Product Evaluation.}
Every image is combined with every degradation type and every severity level, forming the full Cartesian product over images, degradations, and severities. This protocol corresponds closely to the benchmark construction used in ImageNet-C and COCO-C and provides exhaustive coverage at the cost of substantially increased storage and evaluation time.

\smallsec{Protocol C: Full-Factorial Degradation Chains.}
Multiple degradations are applied sequentially in a fixed and explicitly reported order. This protocol enables controlled studies of compound effects and causal interactions, for example when blur, noise, and compression are applied in sequence. Since degradation composition is generally non-commutative, the application order must always be documented.

\smallsec{Protocol D: Random Compositional Chains.}
Following IQA-inspired augmentation settings~\cite{Agnolucci_WACV_2024,Zhao_CVPR_2023}, degradation chains can be sampled randomly from semantically distinct canonical groups. This protocol yields highly diverse degradation patterns and is particularly useful for augmentation pipelines, representation learning, or stress testing beyond fixed benchmark suites.

These protocols differ in dataset construction, but not in the underlying interpretive framework. In all cases, degradations can still be annotated by dominant cause and perceptual effect, and their native severities can still be characterized using PSNR, $(1-\mathrm{SSIM})$, and LPIPS. The taxonomy therefore remains an organizational layer, while the severity measurement layer remains descriptive, regardless of whether degradations are applied singly or in composition.

For reproducible reporting, we recommend that any application protocol specify at least the following: the degradation backend, the native severity schedule, the mapping to canonical taxonomy groups, the image selection protocol, and the distortion metrics used for severity characterization. For protocols involving sequential composition, the order of application should always be stated explicitly, since different orders can lead to materially different visual effects and measured strengths.

In the present study, all main-paper results are restricted to single-degradation evaluation at native severity levels in order to isolate severity semantics and maintain comparability with prior robustness benchmarks. The additional protocols summarized here are included to document the broader applicability of the framework and to support future extensions of \texttt{imdeg}.

\par\medskip
\phantomsection
\section*{Supplementary Evaluation: Degradation Functions}
\addcontentsline{toc}{subsection}{Supplementary Evaluation: Degradation Functions}
\label{app:degfunctions}

Visual examples of the degradations applied to a representative COCO image are shown in Figures~\ref{fig:hendrycks_grid_supp}, \ref{fig:agnolucci_grid_supp}, and \ref{fig:Liu_example_degs_supp}. These galleries provide qualitative context for the degradation families evaluated in the main paper and supplementary detector comparisons.

Figure~\ref{fig:elastic_transform} illustrates the special case of \degfunc{Elastic Transform}, where certain native severity levels can induce larger global displacements than higher ones, leading to irregular robustness behavior. Such degradations should therefore be interpreted with caution in comparisons based on native severity alone.

\renewcommand{\subw}{0.15\textwidth} 
\begin{figure*}[t]
	\captionsetup[subfigure]{justification=centering}
	\centering	
	\begin{subfigure}[t]{\subw}
		\centering
		\includegraphics[width=\linewidth]{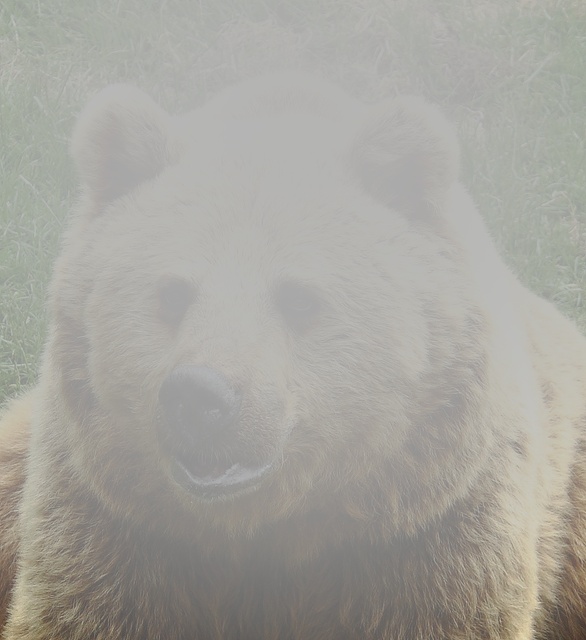}
		\caption*{'Fog'}
	\end{subfigure}\hfill
	\begin{subfigure}[t]{\subw}
		\centering
		\includegraphics[width=\linewidth]{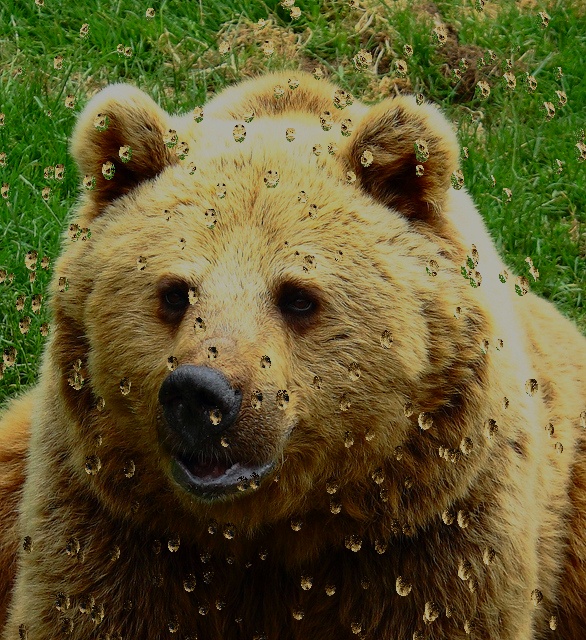}
		\caption*{'Lens Obstruction}
	\end{subfigure}\hfill
	\begin{subfigure}[t]{\subw}
		\centering
		\includegraphics[width=\linewidth]{img/fig_14/0003_Liu_IJCV_2024_focus_motor_damage_degraded_img.jpg}
		\caption*{'Focus Motor Damage'}
	\end{subfigure}\hfill
	\begin{subfigure}[t]{\subw}
		\centering
		\includegraphics[width=\linewidth]{img/fig_14/0004_Liu_IJCV_2024_ccd_sensor_damage_degraded_img.jpg}
		\caption*{'CCD Sensor Damage'}
	\end{subfigure}
	
	\vspace{0.6em}
	
	\begin{subfigure}[t]{\subw}
		\centering
		\includegraphics[width=\linewidth]{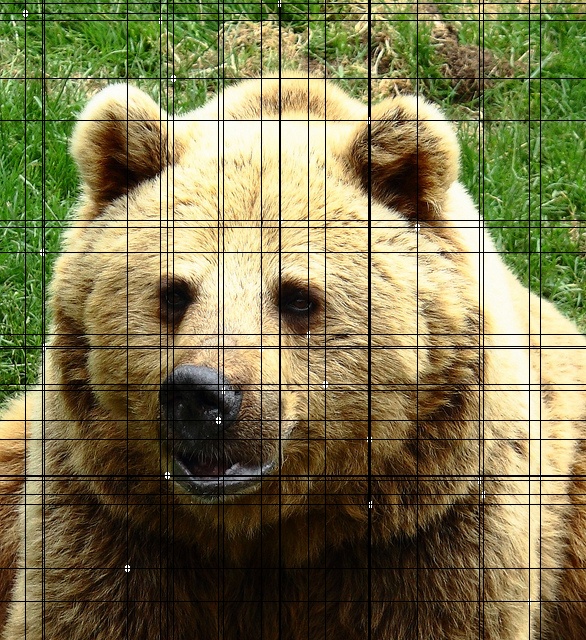}
		\caption*{'CMOS Sensor Damage'}
	\end{subfigure}\hfill
	\begin{subfigure}[t]{\subw}
		\centering
		\includegraphics[width=\linewidth]{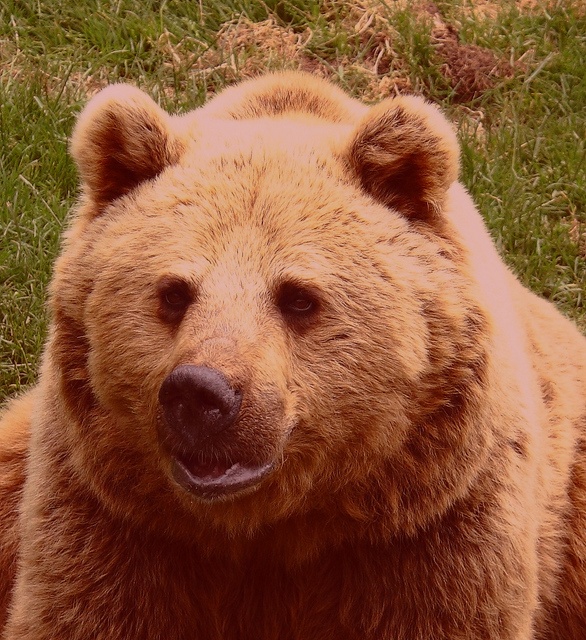}
		\caption*{'Insufficient Black Level Correction'}
	\end{subfigure}\hfill
	\begin{subfigure}[t]{\subw}
		\centering
		\includegraphics[width=\linewidth]{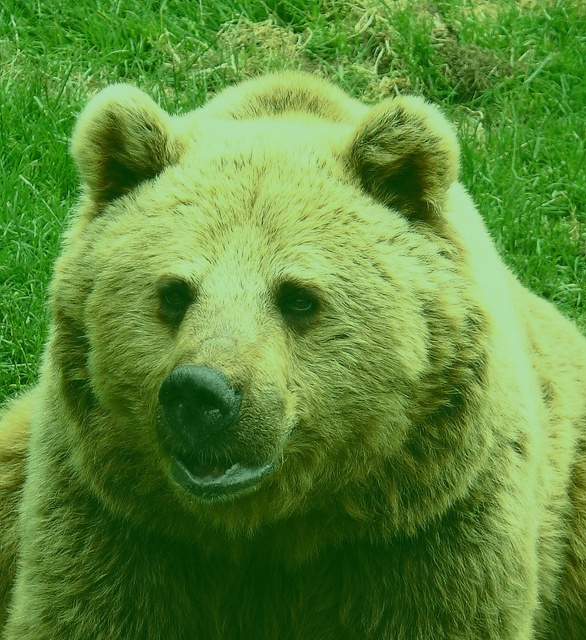}
		\caption*{'Excessive Black Level Correction'}
	\end{subfigure}\hfill
	\begin{subfigure}[t]{\subw}
		\centering
		\includegraphics[width=\linewidth]{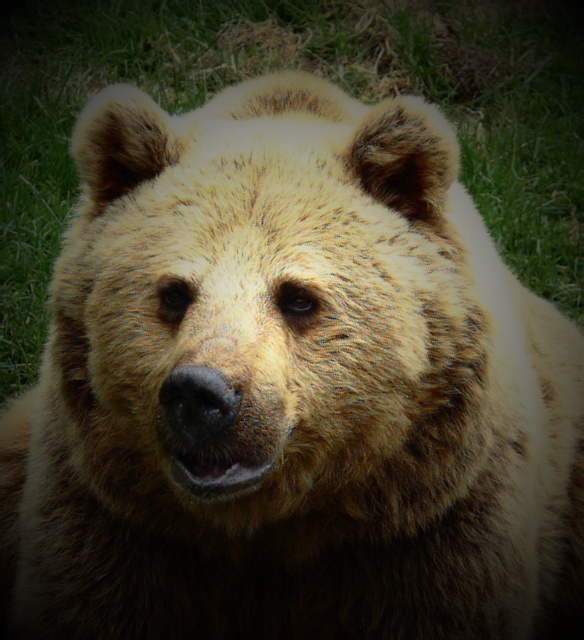}
		\caption*{'Lens Shading Correction Damage'}
	\end{subfigure}
	
	\vspace{0.6em}
	
	\begin{subfigure}[t]{\subw}
		\centering
		\includegraphics[width=\linewidth]{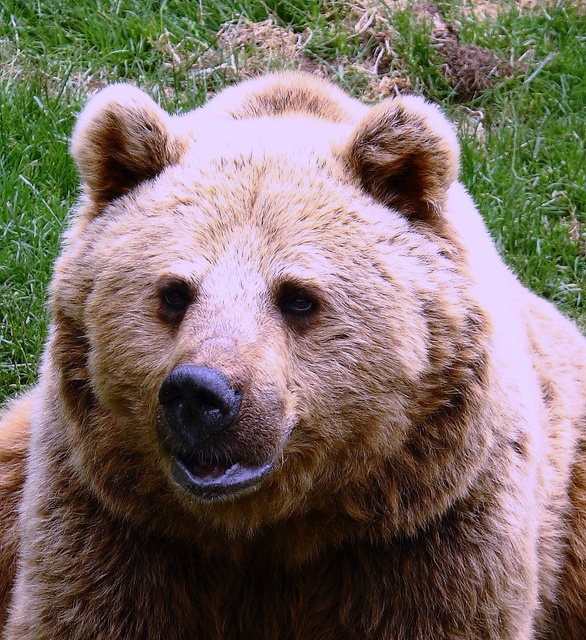}
		\caption*{'Automated White Balance Damage'}
	\end{subfigure}\hfill
	\begin{subfigure}[t]{\subw}
		\centering
		\includegraphics[width=\linewidth]{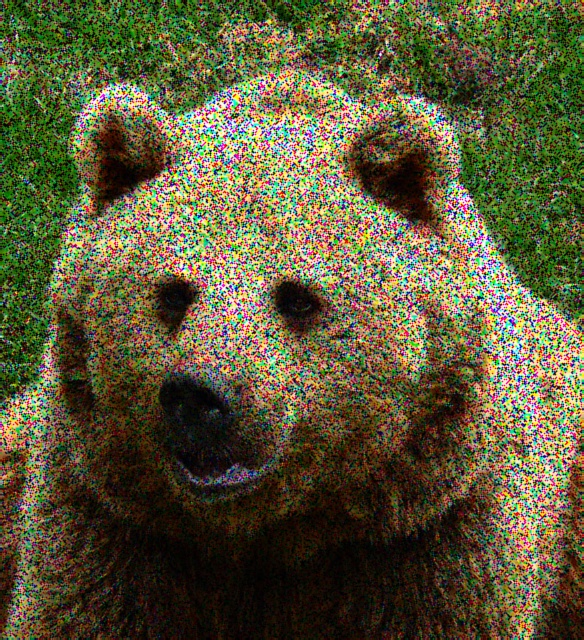}
		\caption*{'Bad Pixel Correction Damage'}
	\end{subfigure}\hfill
	\begin{subfigure}[t]{\subw}
		\centering
		\includegraphics[width=\linewidth]{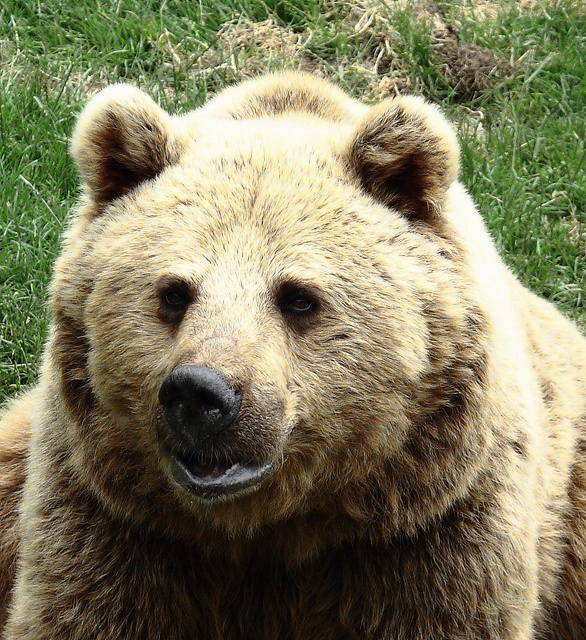}
		\caption*{'Colour Filter Array Interpolation Damage'}
	\end{subfigure}\hfill
	\begin{subfigure}[t]{\subw}
		\centering
		\includegraphics[width=\linewidth]{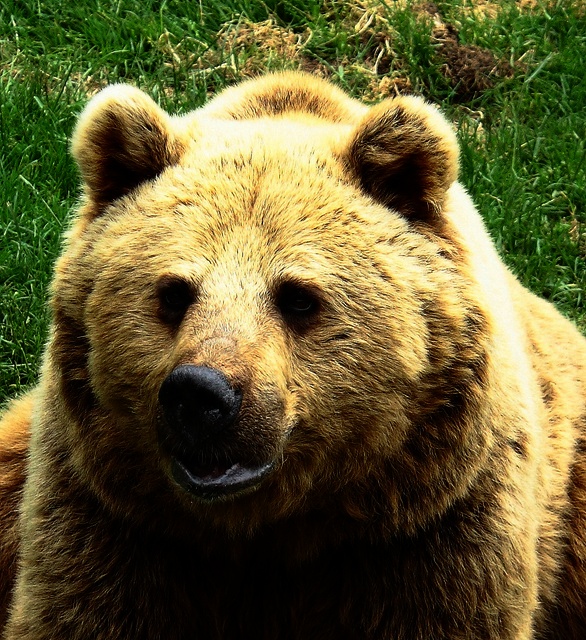}
		\caption*{'Gamma Correction Damage'}
	\end{subfigure}
	
	\vspace{0.6em}
	
	\begin{subfigure}[t]{\subw}
		\centering
		\includegraphics[width=\linewidth]{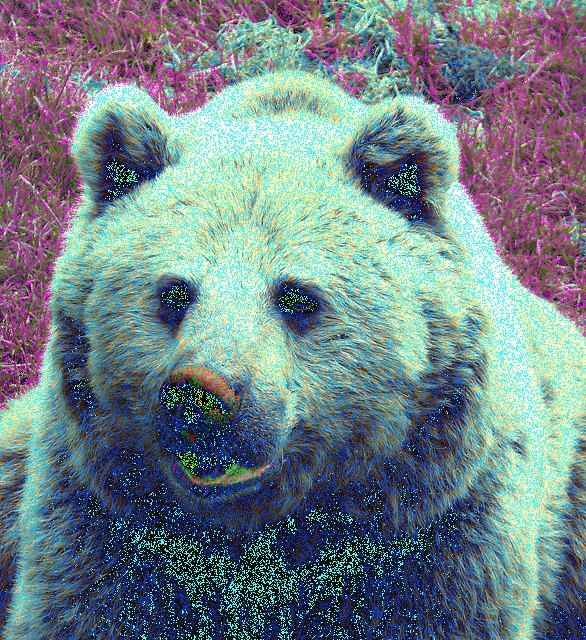}
		\caption*{'Colour Space Conversion Damage'}
	\end{subfigure}\hfill
	\begin{subfigure}[t]{\subw}
		\centering
		\includegraphics[width=\linewidth]{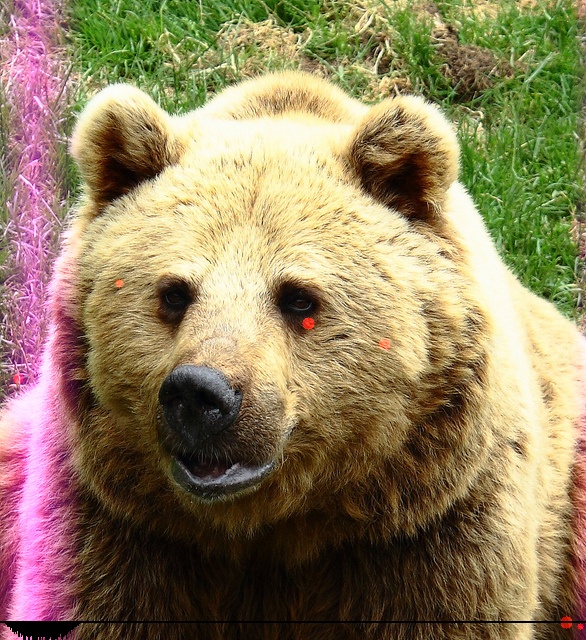}
		\caption*{'Sensor Broken'}
	\end{subfigure}\hfill
	\begin{subfigure}[t]{\subw}
		\centering
		\includegraphics[width=\linewidth]{img/fig_14/0014_Liu_IJCV_2024_memory_exceptions_degraded_img.jpg}
		\caption*{'Memory Exception'}
	\end{subfigure}\hfill
	\begin{subfigure}[t]{\subw}
		\centering
		\includegraphics[width=\linewidth]{img/fig_14/0015_Liu_IJCV_2024_transfer_harness_exceptions_degraded_img.jpg}
		\caption*{'Transfer Harness Exceptions'}
	\end{subfigure}
	\caption{
		Visual examples of degradations from~\cite{Liu_IJCV_2024} at native severity level 3 on an example COCO image~\cite{Lin_ECCV_2014}.
	}
	\label{fig:Liu_example_degs_supp}
\end{figure*}
\renewcommand{\subw}{0.15\textwidth} 
\begin{figure*}[t]
	\captionsetup[subfigure]{justification=centering}
	\centering	
	\begin{subfigure}[t]{\subw}
		\centering
		\includegraphics[width=\linewidth]{img/fig_15/0001_Agnolucci_WACV_2024_gaublur_degraded_img.jpg}
		\caption*{'Gaussian Blur'}
	\end{subfigure}\hfill
	\begin{subfigure}[t]{\subw}
		\centering
		\includegraphics[width=\linewidth]{img/fig_15/0002_Agnolucci_WACV_2024_lensblur_degraded_img.jpg}
		\caption*{'Lens Blur'}
	\end{subfigure}\hfill
	\begin{subfigure}[t]{\subw}
		\centering
		\includegraphics[width=\linewidth]{img/fig_15/0003_Agnolucci_WACV_2024_motionblur_degraded_img.jpg}
		\caption*{'Motion Blur'}
	\end{subfigure}\hfill
	\begin{subfigure}[t]{\subw}
		\centering
		\includegraphics[width=\linewidth]{img/fig_15/0004_Agnolucci_WACV_2024_whitenoise_degraded_img.jpg}
		\caption*{'White Noise'}
	\end{subfigure}\hfill
	\begin{subfigure}[t]{\subw}
		\centering
		\includegraphics[width=\linewidth]{img/fig_15/0005_Agnolucci_WACV_2024_whitenoiseCC_degraded_img.jpg}
		\caption*{'White Noise in Color Component'}
	\end{subfigure}\hfill
	\begin{subfigure}[t]{\subw}
		\centering
		\includegraphics[width=\linewidth]{img/fig_15/0006_Agnolucci_WACV_2024_impulsenoise_degraded_img.jpg}
		\caption*{'Impulse Noise'}
	\end{subfigure}
	
	\vspace{0.6em}
	
	\begin{subfigure}[t]{\subw}
		\centering
		\includegraphics[width=\linewidth]{img/fig_15/0007_Agnolucci_WACV_2024_multnoise_degraded_img.jpg}
		\caption*{'Multiplicative Noise'}
	\end{subfigure}\hfill
	\begin{subfigure}[t]{\subw}
		\centering
		\includegraphics[width=\linewidth]{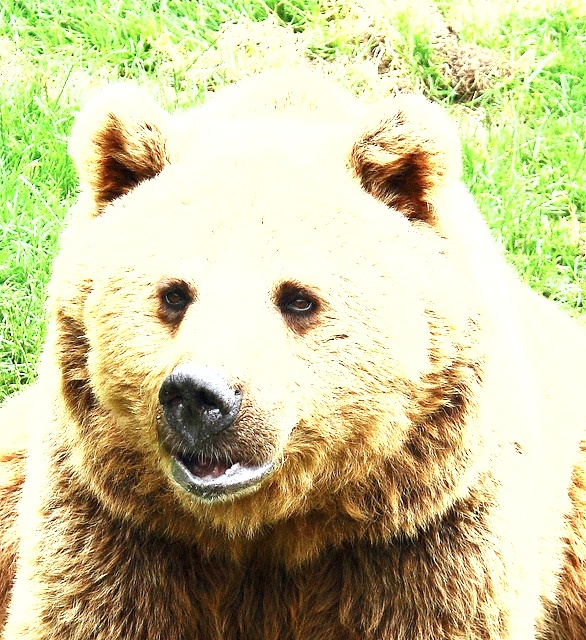}
		\caption*{'Brighten'}
	\end{subfigure}\hfill
	\begin{subfigure}[t]{\subw}
		\centering
		\includegraphics[width=\linewidth]{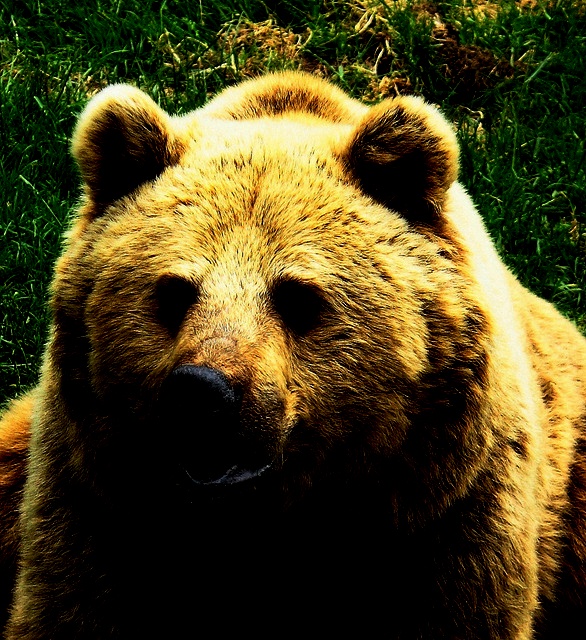}
		\caption*{'Darken'}
	\end{subfigure}\hfill
	\begin{subfigure}[t]{\subw}
		\centering
		\includegraphics[width=\linewidth]{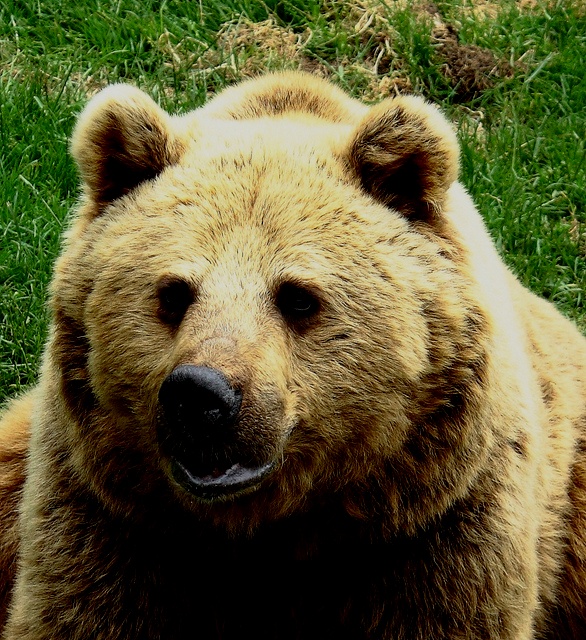}
		\caption*{'Mean Shift'}
	\end{subfigure}\hfill
	\begin{subfigure}[t]{\subw}
		\centering
		\includegraphics[width=\linewidth]{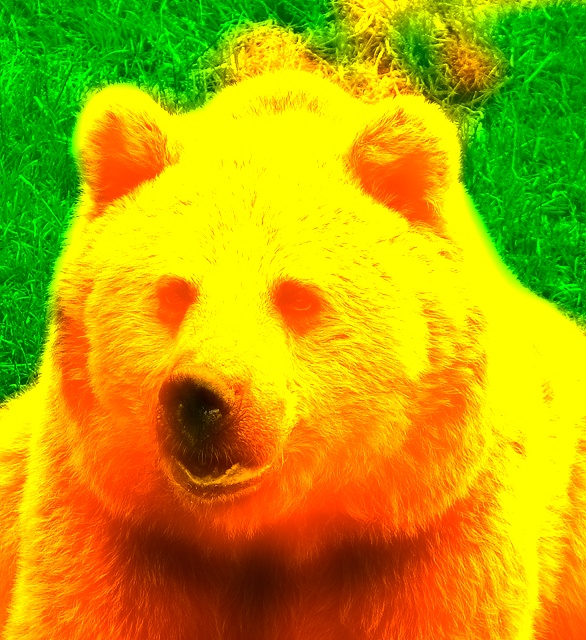}
		\caption*{'Color Diffusion'}
	\end{subfigure}\hfill
	\begin{subfigure}[t]{\subw}
		\centering
		\includegraphics[width=\linewidth]{img/fig_15/0012_Agnolucci_WACV_2024_colorshift_degraded_img.jpg}
		\caption{'Color Shift'}
	\end{subfigure}
	
	\vspace{0.6em}
	
	\begin{subfigure}[t]{\subw}
		\centering
		\includegraphics[width=\linewidth]{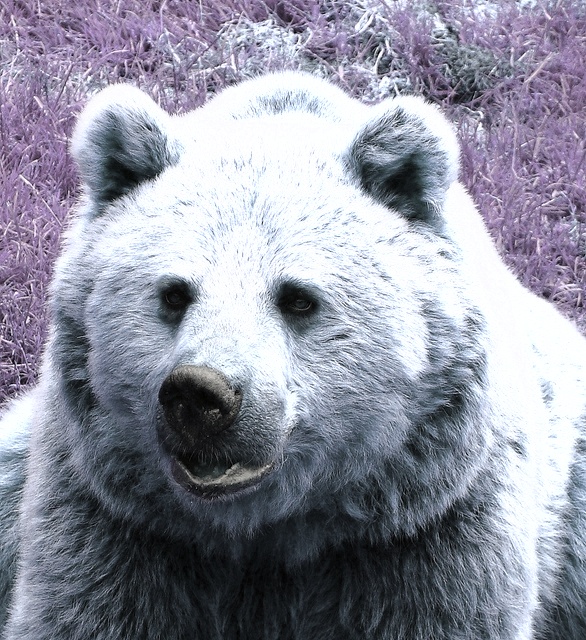}
		\caption*{'Color Saturation 1'}
	\end{subfigure}\hfill
	\begin{subfigure}[t]{\subw}
		\centering
		\includegraphics[width=\linewidth]{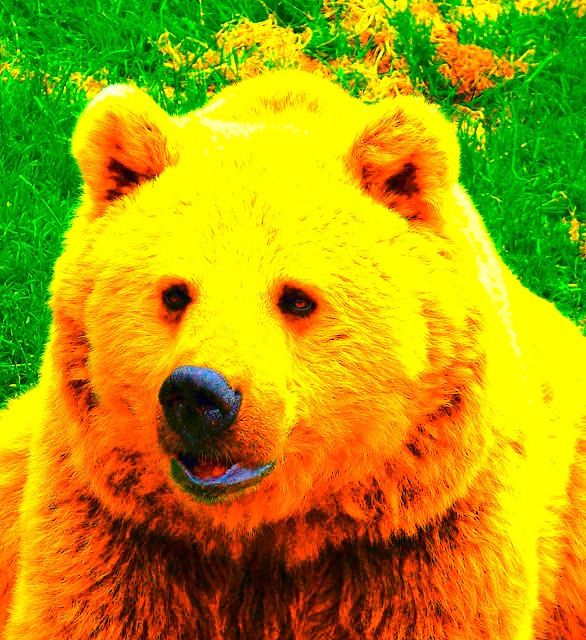}
		\caption*{'Color Saturation 2'}
	\end{subfigure}\hfill
	\begin{subfigure}[t]{\subw}
		\centering
		\includegraphics[width=\linewidth]{img/fig_15/0015_Agnolucci_WACV_2024_jpeg2000_degraded_img.jpg}
		\caption*{'JPEG2000'}
	\end{subfigure}\hfill
	\begin{subfigure}[t]{\subw}
		\centering
		\includegraphics[width=\linewidth]{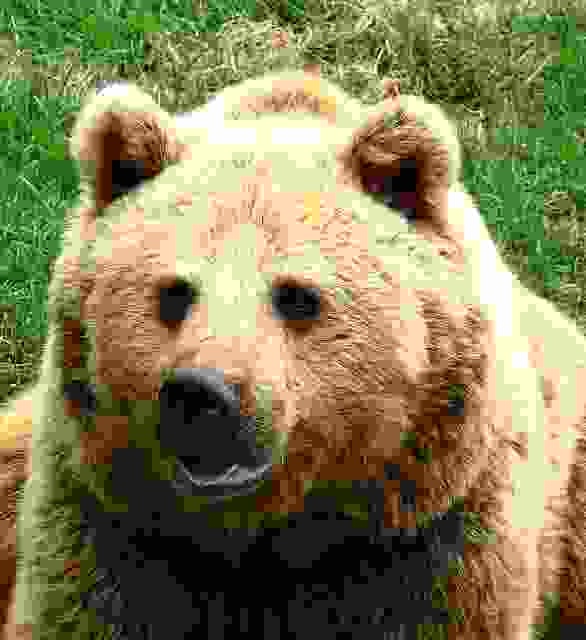}
		\caption*{'JPEG'}
	\end{subfigure}\hfill
	\begin{subfigure}[t]{\subw}
		\centering
		\includegraphics[width=\linewidth]{img/fig_15/0017_Agnolucci_WACV_2024_jitter_degraded_img.jpg}
		\caption*{'Jitter'}
	\end{subfigure}\hfill
	\begin{subfigure}[t]{\subw}
		\centering
		\includegraphics[width=\linewidth]{img/fig_15/0018_Agnolucci_WACV_2024_noneccpatch_degraded_img.jpg}
		\caption*{'Non-Eccentricity Patch'}
	\end{subfigure}
	
	\vspace{0.6em}
	
	\begin{subfigure}[t]{\subw}
		\centering
		\includegraphics[width=\linewidth]{img/fig_15/0019_Agnolucci_WACV_2024_pixelate_degraded_img.jpg}
		\caption*{'Pixelate'}
	\end{subfigure}\hfill
	\begin{subfigure}[t]{\subw}
		\centering
		\includegraphics[width=\linewidth]{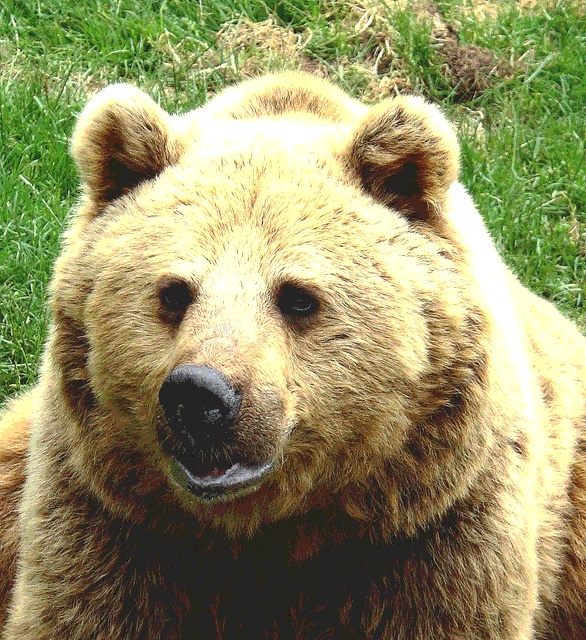}
		\caption*{'Quantization'}
	\end{subfigure}\hfill
	\begin{subfigure}[t]{\subw}
		\centering
		\includegraphics[width=\linewidth]{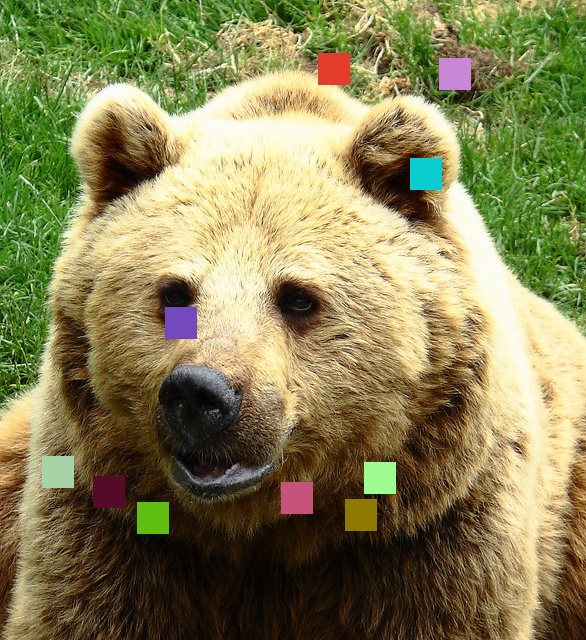}
		\caption*{'Color Block'}
	\end{subfigure}\hfill
	\begin{subfigure}[t]{\subw}
		\centering
		\includegraphics[width=\linewidth]{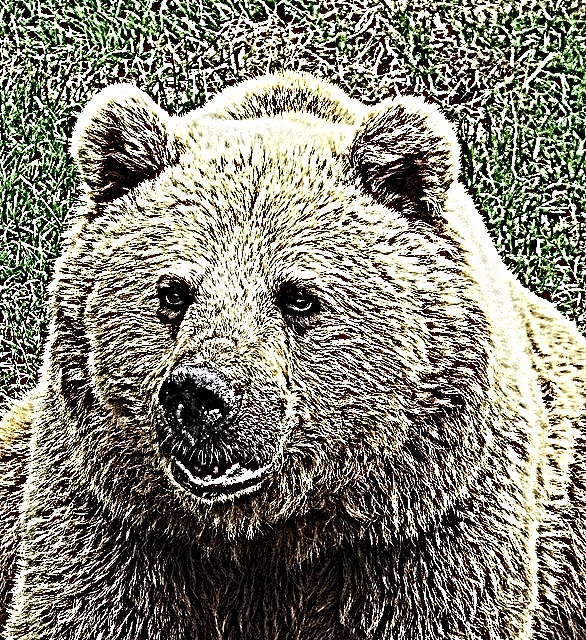}
		\caption*{'High Sharpen'}
	\end{subfigure}\hfill
	\begin{subfigure}[t]{\subw}
		\centering
		\includegraphics[width=\linewidth]{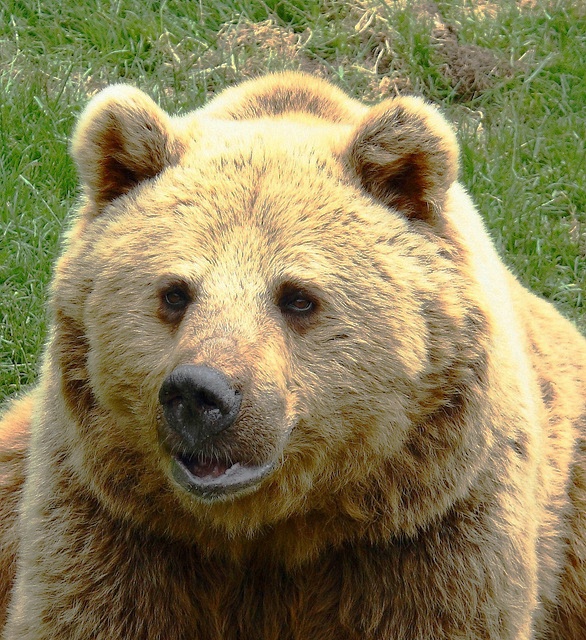}
		\caption*{'Nonlinear Contrast change}
	\end{subfigure}\hfill
	\begin{subfigure}[t]{\subw}
		\centering
		\includegraphics[width=\linewidth]{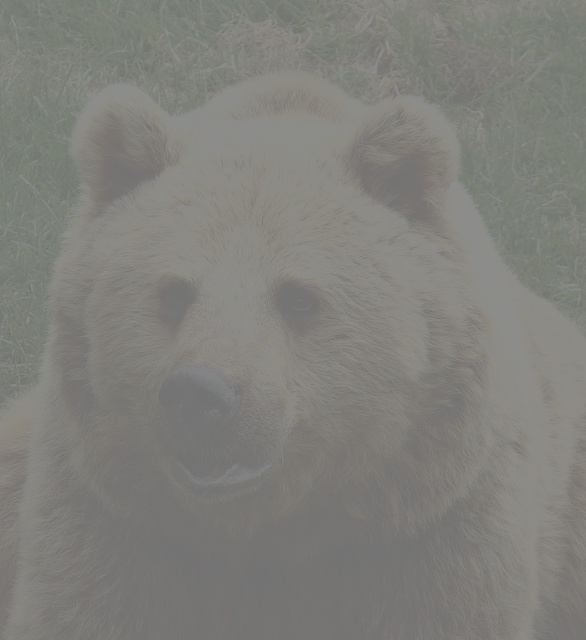}
		\caption*{'Linear Contrast Change'}
	\end{subfigure}	
	\caption{
		Visual examples of the 24 degradations from~\cite{Agnolucci_WACV_2024} (based on KADID-10k~\cite{Lin_QoMEX_2019})
		at native severity level 5 on an example COCO image~\cite{Lin_ECCV_2014}.
	}	
	\label{fig:agnolucci_grid_supp}
\end{figure*}
\renewcommand{\subw}{0.15\textwidth} 
\begin{figure*}[t]
	\captionsetup[subfigure]{justification=centering}
	\centering	
	\begin{subfigure}[t]{\subw}
		\centering
		\includegraphics[width=\linewidth]{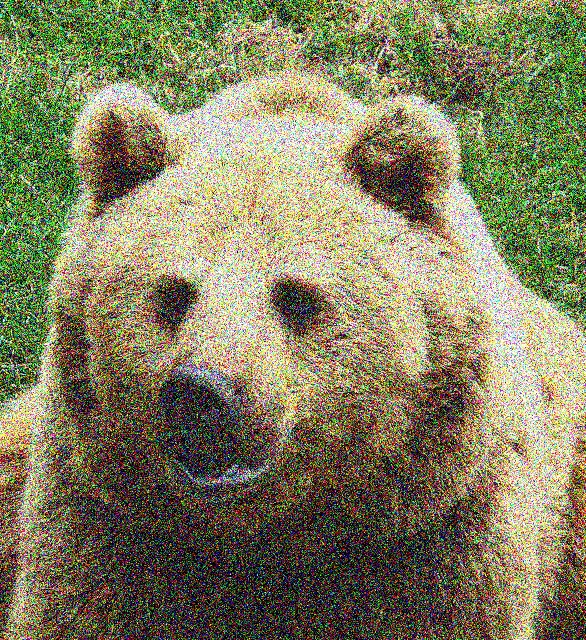}
		\caption*{'Gaussian Noise'}
	\end{subfigure}\hfill
	\begin{subfigure}[t]{\subw}
		\centering
		\includegraphics[width=\linewidth]{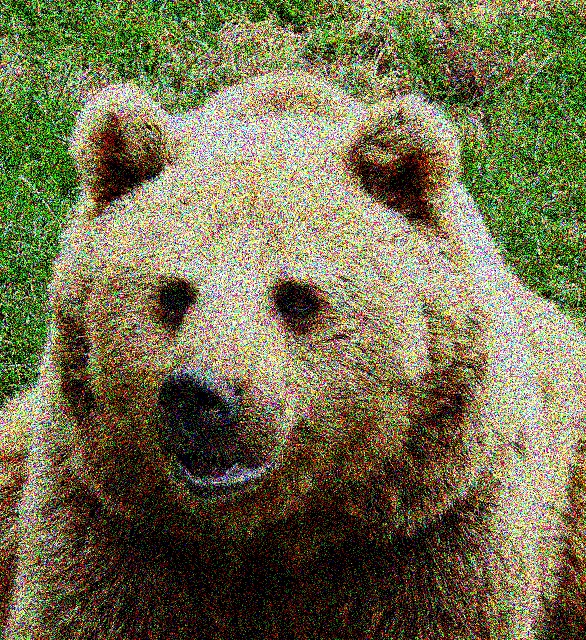}
		\caption*{'Shot Noise'}
	\end{subfigure}\hfill
	\begin{subfigure}[t]{\subw}
		\centering
		\includegraphics[width=\linewidth]{img/fig_16/0003_Hendrycks_ICLR_2019_impulse_noise_degraded_img.jpg}
		\caption*{'Impulse Noise'}
	\end{subfigure}\hfill
	\begin{subfigure}[t]{\subw}
		\centering
		\includegraphics[width=\linewidth]{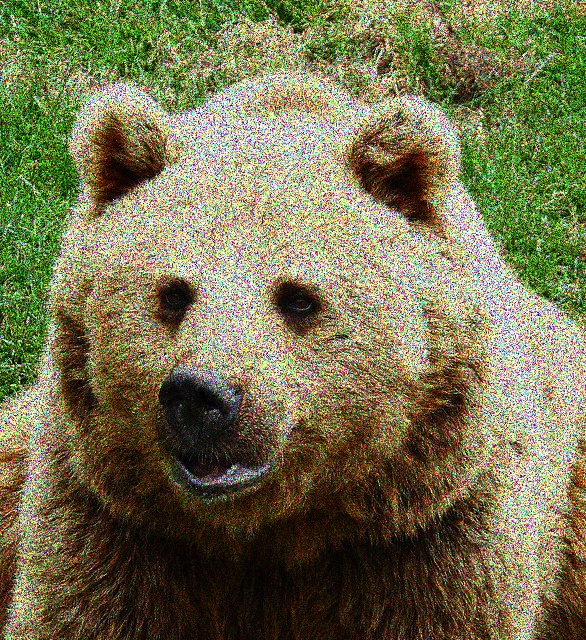}
		\caption*{'Speckle Noise'}
	\end{subfigure}
	
	\vspace{0.6em}
	
	\begin{subfigure}[t]{\subw}
		\centering
		\includegraphics[width=\linewidth]{img/fig_16/0005_Hendrycks_ICLR_2019_defocus_blur_degraded_img.jpg}
		\caption*{'Defocus Blur'}
	\end{subfigure}\hfill
	\begin{subfigure}[t]{\subw}
		\centering
		\includegraphics[width=\linewidth]{img/fig_16/0006_Hendrycks_ICLR_2019_gaussian_blur_degraded_img.jpg}
		\caption*{'Gaussian Blur'}
	\end{subfigure}\hfill
	\begin{subfigure}[t]{\subw}
		\centering
		\includegraphics[width=\linewidth]{img/fig_16/0007_Hendrycks_ICLR_2019_glass_blur_degraded_img.jpg}
		\caption*{'Glass Blur'}
	\end{subfigure}\hfill
	\begin{subfigure}[t]{\subw}
		\centering
		\includegraphics[width=\linewidth]{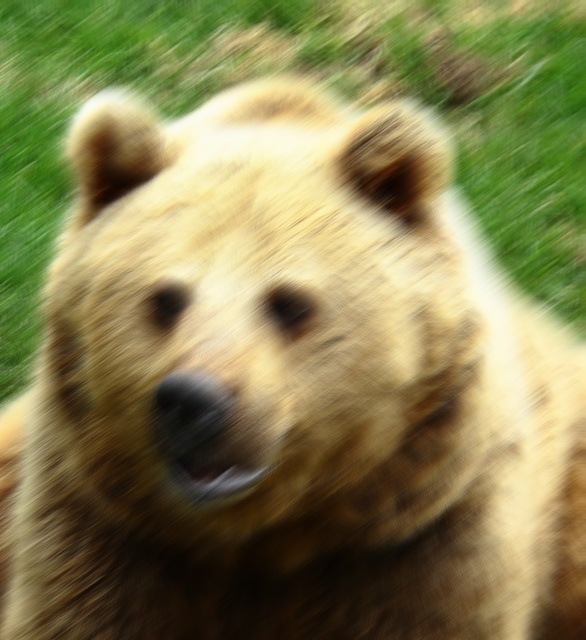}
		\caption*{'Motion Blur'}
	\end{subfigure}\hfill
	\begin{subfigure}[t]{\subw}
		\centering
		\includegraphics[width=\linewidth]{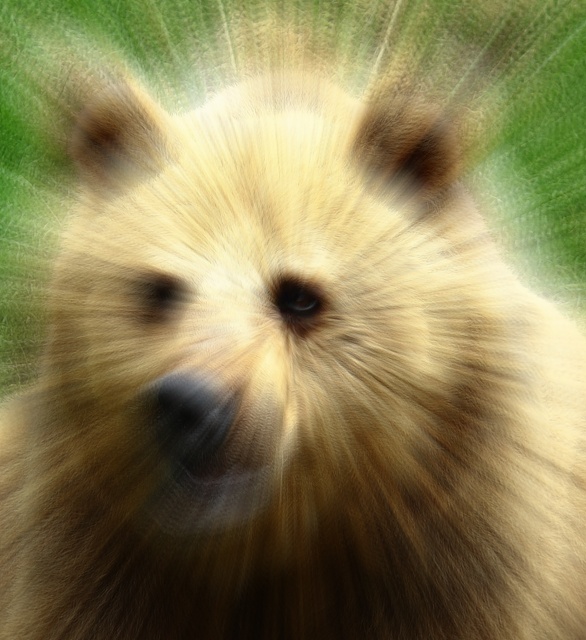}
		\caption*{'Zoom Blur'}
	\end{subfigure}
	
	\vspace{0.6em}
	
	\begin{subfigure}[t]{\subw}
		\centering
		\includegraphics[width=\linewidth]{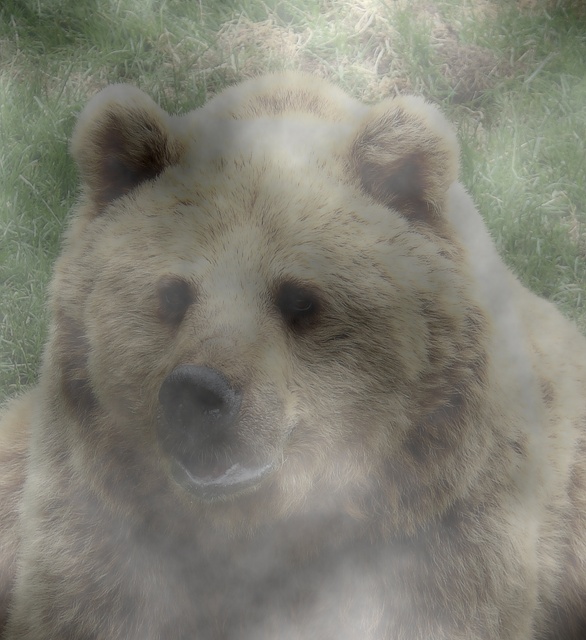}
		\caption*{'Fog'}
	\end{subfigure}\hfill
	\begin{subfigure}[t]{\subw}
		\centering
		\includegraphics[width=\linewidth]{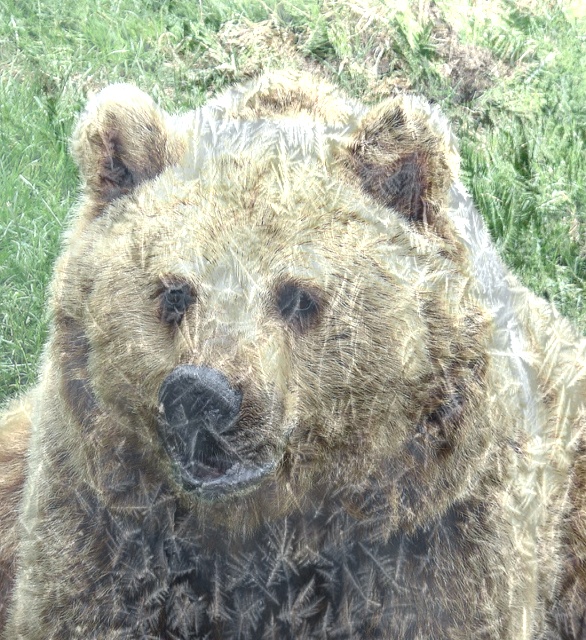}
		\caption*{'Frost'}
	\end{subfigure}\hfill
	\begin{subfigure}[t]{\subw}
		\centering
		\includegraphics[width=\linewidth]{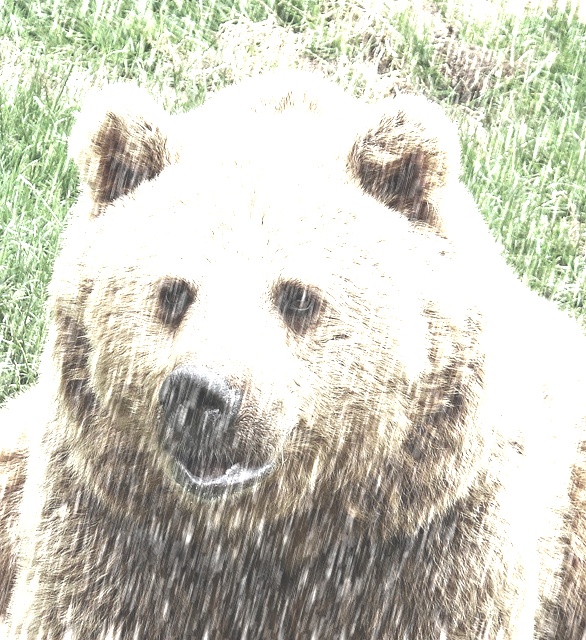}
		\caption{'Snow'}
	\end{subfigure}\hfill
	\begin{subfigure}[t]{\subw}
		\centering
		\includegraphics[width=\linewidth]{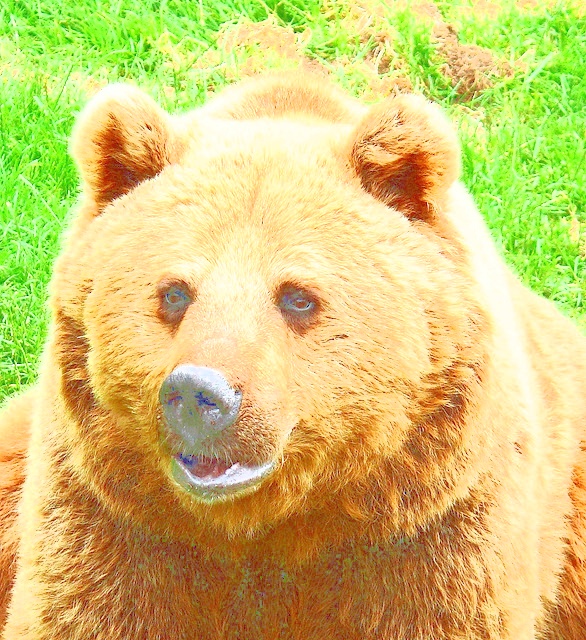}
		\caption*{'Brightness'}
	\end{subfigure}
	
	\vspace{0.6em}
	
	\begin{subfigure}[t]{\subw}
		\centering
		\includegraphics[width=\linewidth]{img/fig_16/0013_Hendrycks_ICLR_2019_spatter_degraded_img.jpg}
		\caption*{'Spatter'}
	\end{subfigure}\hfill
	\begin{subfigure}[t]{\subw}
		\centering
		\includegraphics[width=\linewidth]{img/fig_16/0015_Hendrycks_ICLR_2019_contrast_degraded_img.jpg}
		\caption*{'Contrast'}
	\end{subfigure}\hfill
	\begin{subfigure}[t]{\subw}
		\centering
		\includegraphics[width=\linewidth]{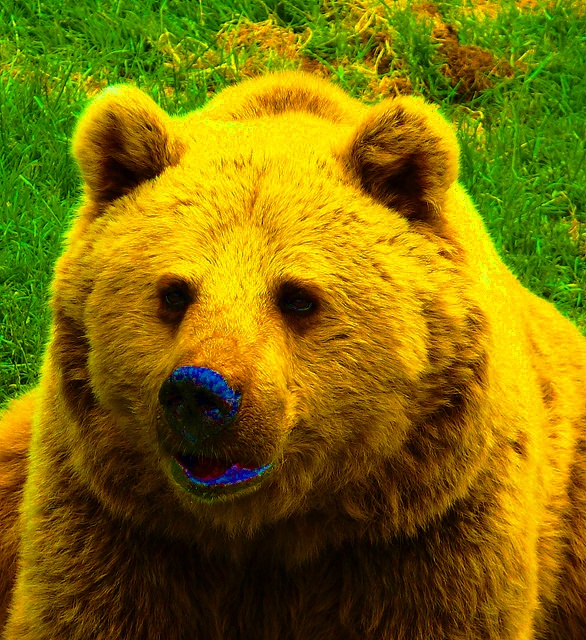}
		\caption*{'Saturate'}
	\end{subfigure}\hfill
	\begin{subfigure}[t]{\subw}
		\centering
		\includegraphics[width=\linewidth]{img/fig_16/0017_Hendrycks_ICLR_2019_jpeg_compression_degraded_img.jpg}
		\caption*{'JPEG'}
	\end{subfigure}\hfill
	\begin{subfigure}[t]{\subw}
		\centering
		\includegraphics[width=\linewidth]{img/fig_16/0018_Hendrycks_ICLR_2019_pixelate_degraded_img.jpg}
		\caption*{'Pixelate'}
	\end{subfigure}\hfill
	\begin{subfigure}[t]{\subw}
		\centering
		\includegraphics[width=\linewidth]{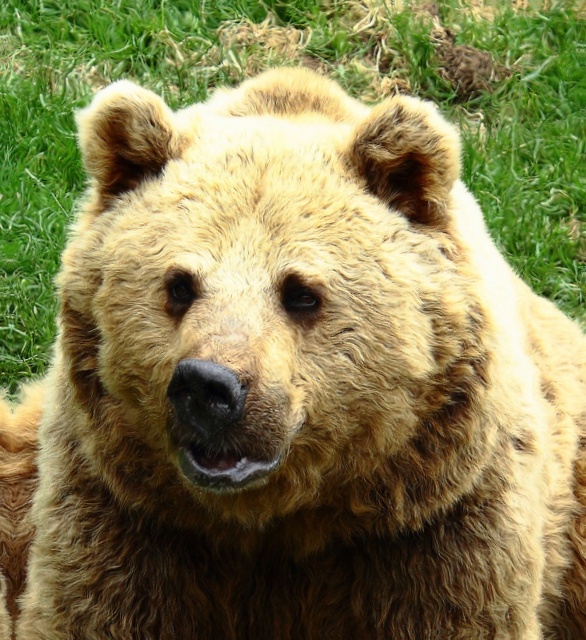}
		\caption*{'Elastic Transform'}
	\end{subfigure}
	\caption{Visualizations of the 19 degradation function in \cite{Michaelis_NeurIPSW_2019} based on \cite{Hendrycks_ICLR_2019} for a severity level of 5 on an example image from the COCO~\cite{Lin_ECCV_2014} dataset.}
	\label{fig:hendrycks_grid_supp}
\end{figure*}

\renewcommand{\subw}{0.15\textwidth} 
\begin{figure*}[t]
	\captionsetup[subfigure]{justification=centering}
	\centering	
	
	\begin{subfigure}[t]{\subw}
		\centering
		\includegraphics[width=\linewidth]{img/fig_17/000000000139_c0.png}
		\caption*{Original}
	\end{subfigure}\hfill
	\begin{subfigure}[t]{\subw}
		\centering
		\includegraphics[width=\linewidth]{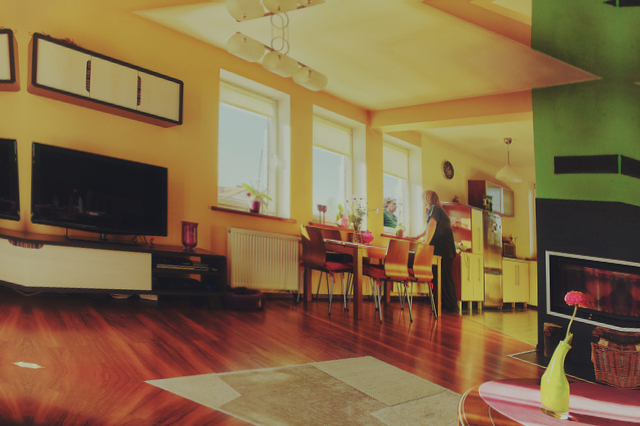}
		\caption*{Severity Level 1}
	\end{subfigure}\hfill
	\begin{subfigure}[t]{\subw}
		\centering
		\includegraphics[width=\linewidth]{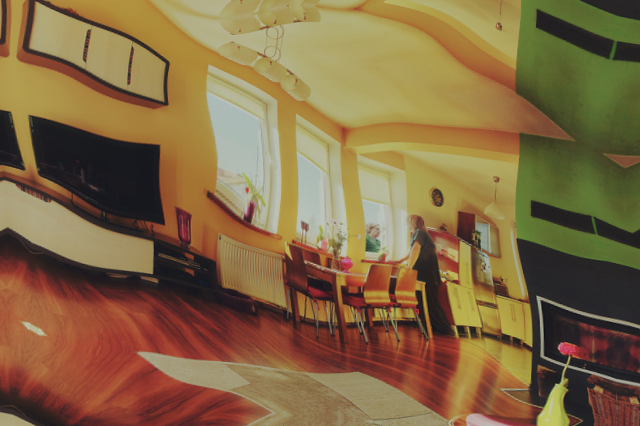}
		\caption*{Severity Level 2}
	\end{subfigure}\hfill
	\begin{subfigure}[t]{\subw}
		\centering
		\includegraphics[width=\linewidth]{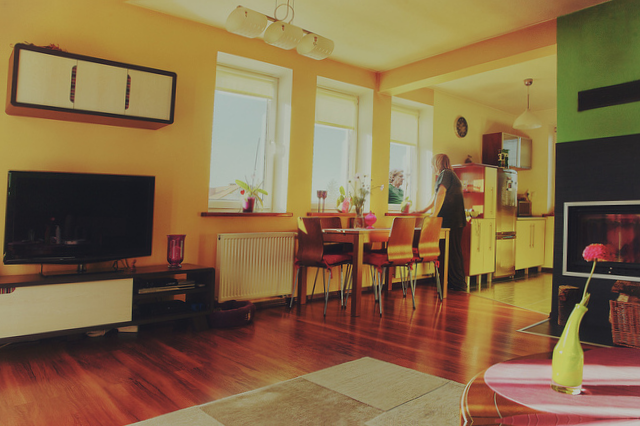}
		\caption*{Severity Level 3}
	\end{subfigure}\hfill
	\begin{subfigure}[t]{\subw}
		\centering
		\includegraphics[width=\linewidth]{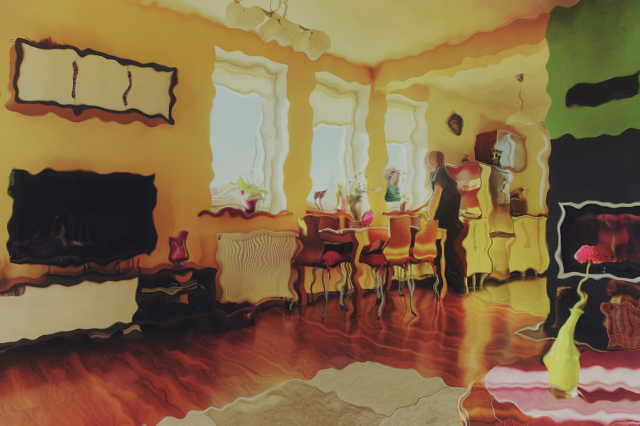}
		\caption*{Severity Level 4}
	\end{subfigure}
	\begin{subfigure}[t]{\subw}
		\centering
		\includegraphics[width=\linewidth]{img/fig_17/000000000139_c5.png}
		\caption*{Severity Level 5}
	\end{subfigure}	
	\caption{
		Visual examples of the \degfunc{Elastic Transform} degradation from~\cite{Agnolucci_WACV_2024} across native severity levels (1--5) on an example COCO image~\cite{Lin_ECCV_2014}.
	}	
	\label{fig:elastic_transform}
\end{figure*}

\clearpage 
\phantomsection
\section*{Supplementary Taxonomy: Mapping}
\addcontentsline{toc}{subsection}{Supplementary Taxonomy: Mapping}
\label{app:mapping}
Mappings of all degradation functions of the realized backend papers to the canonical taxonomy are summarized in Tables~\ref{tab:hendrycks_mapping_all}, \ref{tab:arniqa_mapping_all}, and \ref{tab:liu_mapping_all}. 
\begin{table}[h]
	\centering
	\scriptsize
	\setlength{\tabcolsep}{2pt}
	\begin{tabularx}{\linewidth}{l X l c c}
		\toprule
		Original Category & Degradation Function & Canonical Group ($c$, $e$) & c & e \\
		\midrule
		\multirow{3}{*}{Noise}
		& \degfunc{Gaussian Noise} & G1 Noise & S & N \\
		& \degfunc{Shot Noise}     & G1 Noise & S & N \\
		& \degfunc{Impulse Noise}  & G1 Noise & S & N \\
		\midrule
		\multirow{4}{*}{Blur}
		& \degfunc{Defocus Blur}   & G2 Blur  & S & B \\
		& \degfunc{Gaussian Blur}   & G2 Blur & S & B \\
		& \degfunc{Glass Blur}     & G2 Blur  & S & B \\
		& \degfunc{Zoom Blur}      & G2 Blur  & S & B \\
		\midrule
		\multirow{4}{*}{Weather}
		& \degfunc{Fog}            & G8 Weather/Medium          & E & WX \\
		& \degfunc{Snow}           & G8 Weather/Medium          & E & WX \\
		& \degfunc{Frost}          & G8 Weather/Medium          & E & WX \\
		& \degfunc{Brightness}     & G6 Illumination/Exposure   & E & IL \\
		\midrule
		\multirow{4}{*}{Digital}
		& \degfunc{Contrast}           & G10 Sharpness/Contrast/Texture & R & TX \\
		& \degfunc{JPEG}               & G4 Compression                 & R & CP \\
		& \degfunc{Pixelate}           & G3 Resolution                  & R & RZ \\
		& \degfunc{Elastic Transform}  & G7 Geometry                    & R & GD \\
		\midrule
		\multirow{4}{*}{Validation}
		& \degfunc{Speckle Noise}  & G1 Noise & S & N \\
		& \degfunc{Gaussian Blur}  & G2 Blur  & S & B \\
		& \degfunc{Spatter}        & G8 Weather & E & WX \\
		& \degfunc{Saturate}       & G5 Color/White Balance & R & CL \\
		\bottomrule
	\end{tabularx}
	\caption{Mapping of 19 degradation functions as in \cite{Michaelis_NeurIPSW_2019}, \cite{Oksuz_CVPR_2023}, \cite{Medeiros_arxiv_2025} based on \cite{Hendrycks_ICLR_2019} to the canonical taxonomy.}
	\label{tab:hendrycks_mapping_all}
\end{table}

\begin{table}[h]
	\centering
	\scriptsize
	\setlength{\tabcolsep}{2pt}
	\begin{tabularx}{\linewidth}{l X l c c}
		\toprule
		Original Category & Degradation Function & Canonical Group ($c$, $e$) & c & e \\
		\midrule
		\multirow{3}{*}{Blur}
		& \degfunc{Gaussian Blur} & G2 Blur & S & B \\
		& \degfunc{Lens Blur}     & G2 Blur & S & B \\
		& \degfunc{Motion Blur}   & G2 Blur & S & B \\
		\midrule
		\multirow{4}{*}{Noise}
		& \degfunc{White Noise}                    & G1 Noise & S/R & N \\
		& \degfunc{White Noise in Color Component} & G1 Noise & S/R & N \\
		& \degfunc{Impulse Noise}                  & G1 Noise & S   & N \\
		& \degfunc{Multiplicative Noise}           & G1 Noise & S   & N \\
		\midrule
		\multirow{3}{*}{Brightness Change}
		& \degfunc{Brighten}   & G6 Illumination/Exposure & E/R & IL \\
		& \degfunc{Darken}     & G6 Illumination/Exposure & E/R & IL \\
		& \degfunc{Mean Shift} & G6 Illumination/Exposure & R   & IL \\
		\midrule
		\multirow{4}{*}{Color Distortions}
		& \degfunc{Color Diffusion}     & G5 Color/White Balance & R & CL \\
		& \degfunc{Color Shift}         & G5 Color/White Balance & R & CL \\
		& \degfunc{Color Saturation 1}  & G5 Color/White Balance & R & CL \\
		& \degfunc{Color Saturation 2}  & G5 Color/White Balance & R & CL \\
		\midrule
		\multirow{2}{*}{Compression}
		& \degfunc{JPEG}     & G4 Compression & R & CP \\
		& \degfunc{JPEG2000} & G4 Compression & R & CP \\
		\midrule
		\multirow{5}{*}{Spatial Distortions}
		& \degfunc{Jitter}                  & G7 Geometry                & R & GD \\
		& \degfunc{Non-Eccentricity Patch}  & G9 Occlusion/Obstruction   & R & OC \\
		& \degfunc{Pixelate}                & G3 Resolution/Sampling     & R & RZ \\
		& \degfunc{Quantization}            & G4 Compression             & R & CP \\
		& \degfunc{Color Block}             & G9 Occlusion/Obstruction   & R & OC \\
		\midrule
		\multirow{3}{*}{Sharpness \& Contrast}
		& \degfunc{High Sharpen}               & G10 Sharpness/Contrast/Texture & R & TX \\
		& \degfunc{Linear Contrast Change}     & G10 Sharpness/Contrast/Texture & R & TX \\
		& \degfunc{Nonlinear Contrast Change}  & G10 Sharpness/Contrast/Texture & R & TX \\
		\bottomrule
	\end{tabularx}
	\caption{Mapping of 24 degradation functions in \cite{Agnolucci_WACV_2024} based on \cite{Lin_QoMEX_2019} to the canonical taxonomy.}
	\label{tab:arniqa_mapping_all}
\end{table}
\begin{table}[h]
	\centering
	\scriptsize
	\setlength{\tabcolsep}{2pt}
	\begin{tabularx}{\linewidth}{l X l c c}
		\toprule
		Original Category & Degradation Function & Canonical Group ($c$, $e$) & c & e \\
		\midrule
		\multirow{5}{*}{\textit{Camera Damage}}
		& \degfunc{Fog}                 & G8 Weather/Medium         & E & WX \\
		& \degfunc{Lens Obstruction}    & G9 Occlusion/Obstruction  & S & OC \\
		& \degfunc{Focus Motor Damage}  & G2 Blur                   & S & B \\
		& \degfunc{CCD Sensor Damage}   & G1 Noise                  & S & N \\
		& \degfunc{CMOS Sensor Damage}  & G1 Noise                  & S & N \\
		\midrule
		\multirow{8}{*}{\textit{ISP Damage}}
		& \degfunc{Insufficient Black Level Corruption} & G6 Illumination/Exposure & R & IL \\
		& \degfunc{Excessive Black Level Corruption}    & G6 Illumination/Exposure & R & IL \\
		& \degfunc{Lens Shading Correction Damage}      & G6 Illumination/Exposure & R & IL \\
		& \degfunc{Automated White Balance Damage}      & G5 Color/White Balance   & R & CL \\
		& \degfunc{Bad Pixel Correction Damage}         & G1 Noise                 & R & N \\
		& \degfunc{Color Filter Array Interpolation Damage} & G10 Sharpness/Contrast/Texture & R & TX \\
		& \degfunc{Gamma Correction Damage}             & G10 Sharpness/Contrast/Texture & R & TX \\
		& \degfunc{Color Space Conversion Damage}       & G5 Color/White Balance   & R & CL \\
		\midrule
		\multirow{3}{*}{\textit{Board-Level Damage}}
		& \degfunc{Synchronization Exceptions} & G12 Temporal/Video  & T & TV \\
		& \degfunc{Sensor Broken}          & G11 System/Transfer & T & TV \\
		& \degfunc{Memory Exceptions}          & G11 System/Transfer & T & TV \\
		& \degfunc{Transfer Harness Exceptions} & G11 System/Transfer & T & TV \\
		\bottomrule
	\end{tabularx}
	\caption{Mapping of \cite{Liu_IJCV_2024} degradations to our canonical taxonomy. The original groups are camera, ISP, and board-level damage.}
	\label{tab:liu_mapping_all}
\end{table}

\clearpage 
\phantomsection
\section*{Supplementary Taxonomy: Temporal/Video Subtypes (G12)}
\addcontentsline{toc}{subsection}{Supplementary Taxonomy: Temporal/Video Subtypes (G12)}
\label{app:tv-subtypes}
While most degradations in this work can be modeled as per-image transformations, a substantial class of real-world failures is inherently \emph{temporal}. These effects emerge from time-varying scene conditions (e.g., illumination flicker), sensor readout and control loops (e.g., rolling shutter wobble, autofocus hunting), ISP and video processing (e.g., temporal denoising ghosting, stabilization jitter), or transfer and synchronization faults (e.g., dropped frames, timing jitter, GOP loss). 

To capture this variability without overloading the main taxonomy, we introduce a set of \emph{TV subtypes} within canonical group $G12$ (perceptual effect \textbf{TV}). Table~\ref{tab:taxonomy_tv_subtypes} summarizes the subtype codes used in the \texttt{imdeg} taxonomy, along with a \emph{default causal source} indicating the imaging-pipeline stage that most commonly dominates the effect: Environment ($E$), Sensor/Optics ($S$), ISP/Renderer/Codec ($R$), or Transfer/System ($T$). The default source is intended as a practical annotation prior rather than a strict physical attribution, since real systems may exhibit mixed causes (e.g., codec-induced GOP loss versus transfer-induced packet loss).

In this paper, these temporal subtypes primarily serve as a \emph{taxonomy and annotation layer}. Unless otherwise stated, empirical evaluations operate on frame-wise image degradations, and the temporal taxonomy is included to support consistent naming, reporting, and future extensions to sequence-based benchmarks (e.g., tracking or video detection), where temporal metrics and temporally consistent ground truth become necessary.

\begin{table}[h]
	\centering
	\caption{Canonical corruption groups $G1$--$G12$ as combinations of causal source and phenomenological effect.}
	\label{tab:taxonomy_groups}
	\resizebox{\textwidth}{!}{%
		\begin{tabular}{lll p{7cm}}
			\toprule
			Group & Primary source & Primary effect & Short description \\
			\midrule
			G1  Noise                         & S (Optics/Sensor) & N (Noise) &
			Stochastic or defective sensor/ISP signals (Gaussian, shot, impulse, speckle, bad pixels). \\
			G2  Blur                          & S (Optics/Sensor) & B (Blur) &
			Per-frame focus, motion, zoom or glass-scattering blur. \\
			G3  Resolution / Sampling         & R (ISP/Rendering) & RZ (Resolution) &
			Down/upsampling, scaling, pixelation and other sampling-related resolution changes. \\
			G4  Compression / Quantization    & R (ISP/Rendering) & CP (Compression) &
			Lossy codec artefacts and quantization (JPEG, H.264/HEVC, banding, blocking, bit loss). \\
			G5  Color / White balance         & R (ISP/Rendering) & CL (Color/WB) &
			White balance, color space conversion and saturation changes. \\
			G6  Brightness / Exposure         & R (ISP/Rendering) & IL (Illumination) &
			Brightness, gamma, black-level, vignetting and low-light processing. \\
			G7  Geometry / Spatial            & R (ISP/Rendering) & GD (Geometry) &
			Per-frame geometric distortions (elastic, rotate, translate, scale, tilt, demosaicing artefacts). \\
			G8  Weather / Medium              & E (Environment)   & WX (Weather) &
			Fog, snow, frost, spatter and other medium/weather-induced degradations. \\
			G9  Occlusion / Obstruction       & E (Environment)   & OC (Occlusion) &
			Lens obstruction, dirt, droplets, or other occluding objects. \\
			G10 Sharpness / Contrast / Texture& R (ISP/Rendering) & TX (Sharpness/Contrast) &
			Global or local sharpening and contrast operators, tone mapping, structure enhancement. \\
			G11 System / Transfer / Board     & T (System/Transfer) & -- &
			System- and board-level failures in the data path (sync, memory, transfer harness). \\
			G12 Temporal / Video              & variable (E/S/R/T) & TV (Temporal) &
			Purely temporal artefacts such as flicker, jitter, wobble, ghosting, frame drops/repeats. \\
			\bottomrule
		\end{tabular}
	}
\end{table}

\begin{table}[h]
	\centering
	\caption{Temporal/video subtypes used for the $TV$ effect (group $G12$).}
	\label{tab:taxonomy_tv_subtypes}
	\resizebox{\textwidth}{!}{%
		\begin{tabular}{lllp{7cm}}
			\toprule
			Subtype code & Default source & Name & Description \\
			\midrule
			tv\_flicker      & E / R & Temporal flicker &
			Luminance or chroma flicker over time (mains flicker, AE/gamma pumping). \\
			tv\_awb\_osc     & R     & AWB oscillation &
			Temporal oscillation of white balance / color temperature. \\
			tv\_rs\_wobble   & S     & Rolling-shutter wobble &
			Temporal skew or wobble due to rolling-shutter readout and motion. \\
			tv\_af\_hunting  & S     & Autofocus hunting &
			Temporal focus breathing and hunting between focus distances. \\
			tv\_ois\_jitter  & S     & OIS jitter &
			High-frequency jitter or instability caused by optical image stabilization. \\
			tv\_ghosting     & R     & Temporal ghosting &
			Ghosting due to temporal denoising, deinterlacing or motion-compensated filtering. \\
			tv\_stab\_jitter & R     & Stabilization jitter &
			Residual jitter or drift caused by digital video stabilization. \\
			tv\_vfr          & R     & Variable frame-rate artefacts &
			Temporal artefacts due to variable frame rate and resampling. \\
			tv\_drop\_repeat & T     & Frame drop/repeat &
			Missing or repeated frames, often caused by system/transfer issues. \\
			tv\_desync       & T     & Timing / AV desynchronization &
			Timestamp jitter, AV desync or irregular frame timing. \\
			tv\_gop\_loss    & T     & GOP / slice loss &
			Loss of keyframes or slices causing temporal glitches in decoded video. \\
			tv\_seq\_transforms & R  & Sequential transforms &
			Small step-wise transforms across a sequence (translate/rotate/scale/tilt as in ImageNet-P). \\
			\bottomrule
		\end{tabular}
	}
\end{table}

\clearpage 

\phantomsection
\section*{Supplementary Related Work: Corruption Benchmarks for Object Detection}
\addcontentsline{toc}{subsection}{Supplementary Related Work: Corruption Benchmarks for Object Detection}
\label{app:addcorruptionbenchmarks}
This section provides a detailed, documentary overview of corruption benchmarks and dataset variants used for object detection robustness evaluation.
While the main paper focuses on conceptual differences in degradation definitions, grouping schemes, and severity semantics, this supplementary overview records the breadth of existing benchmarks for completeness and reference.
It also includes extensions to 3D perception, infrared detection, and composed or perceptually dissimilar corruptions that are mentioned only briefly in the main text.

\smallsec{Corruption Benchmarks}
Synthetic corruption benchmarks are designed to evaluate the vulnerability of vision models under controlled variations of input conditions.
\cite{Hendrycks_ICLR_2019}\footnote{\cite{Hendrycks_ICLR_2019}: \url{https://github.com/hendrycks/robustness}} introduced fifteen algorithmic degradation functions at five predefined severity levels for image classification, resulting in ImageNet-C (ImageNet~\cite{Deng_CVPR_2009}) and CIFAR10-C (CIFAR10~\cite{Krizhevsky_2009}). Building on this corruption set, several works extended, modified, or re-parameterized these degradations for object detection, most prominently for COCO~\cite{Lin_ECCV_2014}.

\cite{Michaelis_NeurIPSW_2019}\footnote{\cite{Michaelis_NeurIPSW_2019}: \url{https://github.com/bethgelab/robust-detection-benchmark}}
proposed COCO-C by exhaustively applying all ImageNet-C corruptions at all severity levels to every COCO image, forming a Cartesian product over
images, corruption types, and severity levels. Their implementation also includes adaptations to handle variable image sizes, expanding the original fifteen corruptions to nineteen. In addition to COCO-C, they introduced PASCAL-C (PASCAL VOC~\cite{Everingham_IJCV_2010}) and Cityscapes-C
(Cityscapes~\cite{Cordts_CVPR_2016}). 

In contrast, \cite{Oksuz_CVPR_2023}\footnote{\cite{Oksuz_CVPR_2023}: 	\url{https://github.com/asharakeh/probdet}} used the extended ImageNet-C corruption set but applied a round--robin assignment where each image receives exactly one corruption at one severity. This design leads to statistical properties very different from the Cartesian sampling of \cite{Michaelis_NeurIPSW_2019}. They additionally introduced Obj45K-C and BDD45K-C, modified versions of Object365~\cite{Shao_CVPR_2019} and BDD100K~\cite{Yu_CVPR_2020}.

Using an entirely different corruption family, \cite{Liu_IJCV_2024}\footnote{\cite{Liu_IJCV_2024}: \url{https://sites.google.com/view/real-worldbenchmark/real-world-benchmarking}} introduced a COCO-C and BDD100K-C variant based on real-camera damage, broadening the design space beyond algorithmic distortions.

Beyond 2D detection, robustness benchmarks have also been developed for 3D perception. \cite{Dong_CVPR_2023}\footnote{\cite{Dong_CVPR_2023}:	\url{https://github.com/thu-ml/3D_Corruptions_AD}} evaluated 3D object detectors under geometric, structural, and appearance-based corruptions, introducing 27 degradation types and the datasets KITTI-C, nuScenes-C, and Waymo-C (KITTI~\cite{Geiger2013IJRR}, nuScenes~\cite{nuscenes}, Waymo~\cite{Sun_2020_CVPR}). To generate geometry-consistent corruptions, \cite{Kar_CVPR_20223}\footnote{\cite{Kar_CVPR_20223}: \url{https://3dcommoncorruptions.epfl.ch/}} proposed a set of 20 corruptions derived from scene geometry rather than purely 2D image operators.

For infrared object detection, \cite{Medeiros_arxiv_2025}\footnote{\cite{Medeiros_arxiv_2025}: \url{https://github.com/heitorrapela/wiseod}} introduced Flir-C and LLVIP-C by applying ImageNet-C–style corruptions to FLIR (aligned)~\cite{ZHANG_ICIP_2020,flirDataset} and LLVIP~\cite{jia2021llvip}. 

Another COCO-C variant is introduced by \cite{Beghdadi_ICIP_2022}, which combined global with additional local transformations. 
To investigate the effect of dissimilarities during test-time, \cite{Mintun_NEURIPS_2021}\footnote{\cite{Mintun_NEURIPS_2021}: \url{https://github.com/facebookresearch/augmentation-corruption}} introduced CIFAR-\={c} and ImageNet-\={C} relying for the comparison on 10 perceptually dissimilar corruptions to those of \cite{Hendrycks_ICLR_2019}. To increase data variety, \cite{press2022ccc} uses corruption pairs to generate COCO-CCC.

Since corruptions induce distributional shifts that affect model performance, out-of-distribution (OOD) datasets are sometimes discussed alongside robustness benchmarks. However, OOD datasets mainly target conceptual or domain-level shifts (e.g., stylization, drawings, cartoons, novel objects). For completeness, \cite{mao2023coco}\footnote{\cite{mao2023coco}: \url{https://github.com/alibaba/easyrobust}} introduced COCO-O, containing images with diverging content such as sketches and cartoons. For a deeper overview of OOD detection, we refer to the survey of \cite{Yang_IJCV_2024}.

In addition, adversarial perturbations also constitute a class of image degradations designed to fool model predictions. Due to their fundamentally different nature and objectives, we exclude adversarial robustness from our scope and refer to \cite{Wang_Veldhuis_arxiv_2023} for a detailed categorization in the context of robustness.

\smallsec{Image Quality Assessment (IQA)}
Perceptual image quality assessment (IQA) studies degradations from the perspective of human visual perception rather than downstream task performance. IQA methods are typically categorized as full-reference, reduced-reference, or no-reference, depending on whether a pristine reference image is available.

Classical full-reference metrics quantify perceptual similarity using structural or feature comparisons, such as SSIM~\cite{Wang_TIP_2004}, MS-SSIM~\cite{Wang_2003}, and FSIM~\cite{Zhang_2011}. In contrast, no-reference models often rely on natural scene statistics to detect departures from statistical regularity (e.g., BRISQUE~\cite{Mittal_TIP_2012}, NIQE~\cite{Mittal_SPL_2013}, PIQE~\cite{Venkatanath_NCC_2015}, and DIIVINE~\cite{Moorthy_TIP_2011}).

More recent approaches leverage deep learning to approximate perceptual similarity. These include models trained to regress human opinion scores (e.g., MUSIQ~\cite{Ke_ICCV_2021}, ARNIQA~\cite{Agnolucci_WACV_2024}), as well as those that compute distances in deep feature spaces (e.g., LPIPS~\cite{Zhang_CVPR_2018}, DISTS~\cite{DingTPAMI_2022}). For broader overviews, see recent surveys~\cite{Ma_arxiv_2025,Wang_IQA_arxiv_2023}.

These metrics are benchmarked on datasets with subjective human ratings (mean opinion score (MOS) and variants), including KADID-10k~\cite{Lin_QoMEX_2019}, TID2013~\cite{Ponomarenko_IPIC_2015}, UHD-IQA~\cite{Hosu_ECCVW_2024}, and GFIQA-20k~\cite{Su_ToM_2023}. Crucially, distortion levels in IQA are defined by perceptual separability (often approximating just-noticeable or just-objectionable differences), rather than by fixed algorithmic parameters. Thus, a constant step in severity generally corresponds to a comparable perceived degradation, though not necessarily to a uniform change in parameter space across images or resolutions.

Although IQA and robustness benchmarks pursue different objectives, they employ overlapping degradation types such as blur, noise, compression artifacts, and color distortions. A key distinction lies in severity semantics: IQA datasets define severity levels based on perceptual separability, whereas robustness benchmarks rely on fixed algorithmic or physical parameters. As a result, equal severity indices may correspond to very different perceptual distortion magnitudes across datasets.

\smallsec{Imaging System Modeling}
Imaging system modeling grounds degradations in the physics of image acquisition. Rather than applying abstract transformations, this line of work models the image as the output of a pipeline comprising environment, optics, sensor, and image signal processing (ISP), with parameters dependent on hardware, scene content, and operating conditions.

Real-camera degradation studies manipulate physical or system-level parameters to induce artifacts such as noise, blur, exposure errors, temporal instabilities, or ISP failures. These approaches enable continuous and physically interpretable variation of degradation strength, often as a function of distance, illumination, exposure time, or temperature.
Historically, system performance was evaluated through human observer experiments such as Triangle Orientation Discrimination (TOD)~\cite{Bijl_SPIE_1998} or Minimum Resolvable Contrast Difference (MRCD)~\cite{Lloyd_book_1975}. More recent work explores algorithmic surrogates, including CNN-based detectors and learned estimators trained on real or simulated imagery~\cite{Wegner_SPIE_2023,Wegner_SPIE_2025}.

Compared to synthetic corruption benchmarks and IQA datasets, real-camera studies emphasize causal provenance but often lack standardized grouping or severity conventions that facilitate comparison across datasets or tasks. As a result, their integration into existing robustness evaluation pipelines remains limited, despite their physical realism.

\par\medskip
\phantomsection
\section*{Supplementary Related Work: Weather Datasets}
\addcontentsline{toc}{subsection}{Supplementary Related Work: Weather Datasets}
\label{app:weatherdatasets}

Adverse weather conditions such as fog, haze, rain, or snow constitute a particularly illustrative class of image degradations.
Weather-related degradations appear across all three research communities discussed in this work—synthetic corruption benchmarks, perceptual image quality assessment, and physically grounded imaging system modeling—but with fundamentally different motivations, parameterizations, and interpretations of severity.

In synthetic corruption benchmarks (e.g., ImageNet-C or COCO-C), weather effects such as \degfunc{Fog} are implemented as algorithmic transformations designed to stress-test model robustness.
These operators typically rely on simplified approximations of atmospheric scattering or contrast attenuation with fixed parameter schedules.
Severity levels are defined through backend-specific parameter choices and applied uniformly across images, prioritizing controlled variation over physical realism.

In perceptual image quality assessment (IQA) datasets, visually similar degradations appear under different assumptions.
Fog-, haze-, or contrast-related distortions are included because they systematically affect human-perceived image quality.
Severity levels are chosen to span perceptually meaningful quality ranges, often approximating just-noticeable or just-objectionable differences.
As a result, equal steps in severity aim to correspond to comparable perceptual changes, rather than to fixed physical parameters or imaging conditions.

In contrast, weather effects in autonomous driving, remote sensing, and aerial imagery are commonly modeled using physically grounded simulation pipelines.
Here, depth, scene geometry, illumination, and radiative transport determine the appearance and spatial structure of fog or haze. Representative examples include Foggy Cityscapes \cite{Sakaridis_IJCV_2028} and drone-view haze benchmarks such as HazyDet \cite{feng2025HazyDet}.
These datasets aim to approximate real atmospheric conditions and enable continuous, physically interpretable variation of degradation strength as a function of distance, density, or visibility.

From the perspective of the proposed taxonomy, weather-related degradations therefore differ primarily in their \emph{causal source}—algorithmic transformation, perceptual distortion, or physical atmospheric modeling—while often sharing similar \emph{perceptual effects}, such as reduced contrast, blur, or color desaturation.
Although these variants may appear visually related, their formulations, parameterizations, and severity semantics differ substantially across communities.

By organizing degradations according to causal source and perceptual effect, the proposed taxonomy places synthetic fog operators, IQA haze distortions, and physically based atmospheric scattering models within a coherent conceptual structure.
This makes their similarities explicit while preserving critical distinctions in underlying mechanisms, parameterization, and severity interpretation, and illustrates how degradations with similar visual appearance can be meaningfully compared without conflating their origins.

\par\medskip
\phantomsection
\section*{Supplementary Material: Third-Party Licensing}
\addcontentsline{toc}{subsection}{Third-Party Licensing}
\label{app:thirdpartylicenses}
Our \texttt{imdeg} implementation integrates and/or re-implements degradations described in prior work. The reference implementations of Hendrycks \cite{Hendrycks_ICLR_2019} and Öksüz \cite{Oksuz_CVPR_2023} are released under Apache-2.0, and Michaelis et al.\ \cite{Michaelis_NeurIPSW_2019} under the MIT license. ARNIQA \cite{Agnolucci_WACV_2024} is released under Apache-2.0. For Liu et al.\ \cite{Liu_IJCV_2024}, the dataset is distributed under CC BY 4.0, while the associated repository did not specify a software license at the time of writing; accordingly, our library does not incorporate unlicensed code and relies only on paper-described behavior and independently written implementations. A list of licenses for the mentioned third‑party degradation libraries or datasets is shown in Table~\ref{tab:thirdpartylicenses}.

\begin{table*}[h]
	\centering
	\caption{Third-party code and resources referenced via footnoted repository links in Related Work. Licenses should be verified against each repository's \texttt{LICENSE} file at the commit used in our implementation.}
	\label{tab:thirdparty_licenses}
	\setlength{\tabcolsep}{6pt}
	\renewcommand{\arraystretch}{1.15}
	\resizebox{\textwidth}{!}{%
		\begin{tabular}{c c p{3.0cm} p{4.5cm}}
			\toprule
			Paper / Dataset & Repository (URL) & Code license & Notes \\
			\midrule
			\cite{Hendrycks_ICLR_2019} (ImageNet-C/P) &
			\repo{hendrycks/robustness}{https://github.com/hendrycks/robustness} &
			Apache-2.0 &
			License reported by repository metadata. \\
			\midrule
			\cite{Mintun_NEURIPS_2021} (Augmentation--Corruption) &
			\repo{augmentation-corruption}{https://github.com/facebookresearch/augmentation-corruption}   &
			MIT &
			Repository states MIT license. \\
			\midrule
			\cite{Michaelis_NeurIPSW_2019} (COCO-C variant) &
			\repo{robust-detection-benchmark}{https://github.com/bethgelab/robust-detection-benchmark}  &
			MIT &
			Repository states MIT license. \\
			\midrule
			\cite{Oksuz_CVPR_2023} (COCO-C variant) &
			\repo{asharakeh/probdet}{https://github.com/asharakeh/probdet} &
			Apache-2.0   & Repository states Apache-2.0 license.
			\\
			\midrule
			\cite{Liu_IJCV_2024} (Real-camera corruptions) &
			\repo{real-world-benchmarking}{https://sites.google.com/view/real-worldbenchmark/real-world-benchmarking} &
			\emph{n/a} &
			dataset license CC BY 4.0; repository/license info not clearly stated on site \\
			\midrule
			\cite{Dong_CVPR_2023} (3D\_Corruptions\_AD) &
			\repo{3D Corruptions AD}{https://github.com/thu-ml/3D\_Corruptions\_AD} &
			MIT  &
			Repository states MIT license. \\
			\midrule
			\cite{Kar_CVPR_20223} (3DCommonCorruptions) &
			\repo{3DCommonCorruptions}{https://github.com/EPFL-VILAB/3DCommonCorruptions} &
			Attribution-NonCommercial 4.0 International &
			Repository states Attribution-NonCommercial 4.0 International license.. \\
			\midrule
			\cite{Medeiros_arxiv_2025}	\ (WiSE-OD) &
			\repo{heitorrapela/wiseod}{https://github.com/heitorrapela/wiseod} &
			\emph{n/a} &
			no license information \\
			\midrule
			\cite{mao2023coco} (EasyRobust / COCO-O) &
			\repo{easyrobust}{https://github.com/alibaba/easyrobust} &
			Apache-2.0   & Repository states Apache-2.0 license.
			\\
			\bottomrule
		\end{tabular}
	}
	\label{tab:thirdpartylicenses}
\end{table*}

\par\medskip
\phantomsection
\section*{Supplementary Material: Example Usage (\texttt{imdeg} Library): Calibration and Canonical Degradation}
\addcontentsline{toc}{subsection}{Supplementary Material: Example Usage (\texttt{imgdeg} Library): Calibration and Canonical Degradation}
\label{app:usage}
The \texttt{imdeg} library separates \emph{severity calibration} from
\emph{degradation application}. Calibration maps backend-native severity
levels to measured distortion strength, while degradation application
uses this mapping to apply canonical severity levels that are comparable
across backends.

\subsection{Calibration of a Degradation Function}
\label{subsec:usage_calibration}

Calibration estimates the degradation-strength curve for a given
degradation operator using a small set of reference images.
The result defines a canonical severity axis in metric space.

\begin{verbatim}
	from PIL import Image
	from torchvision.transforms.functional import to_tensor
	from pathlib import Path
	
	from imdeg.calibration import calibrate_distortion
	
	# Load calibration images
	paths = list(Path("calib_images").glob("*.jpg"))[:50]
	images = [to_tensor(Image.open(p).convert("RGB")) for p in paths]
	
	# Calibrate Gaussian blur using SSIM
	calibration = calibrate_distortion(
	images=images,
	paper="Agnolucci_WACV_2024",
	term="gaublur",
	metric="1-ssim",
	)
	
	canonical_strengths = calibration["canonical_strengths"]
\end{verbatim}

The resulting calibration object stores the measured degradation strengths
for each native severity level as well as the derived canonical reference
levels.

\subsection{Applying a Degradation at Canonical Severity}
\label{subsec:usage_application}

Once calibration is available, degradations can be applied using
\texttt{canonical} mode. The requested severity index refers to the
canonical severity axis rather than to backend-native parameters.

\begin{verbatim}
	from PIL import Image
	from torchvision.transforms.functional import to_tensor, to_pil_image
	
	from imdeg import apply_degradation
	
	# Load image to be degraded
	x = to_tensor(Image.open("example.jpg").convert("RGB"))
	
	# Apply Gaussian blur at canonical severity k=3
	y = apply_degradation(
	image=x,
	paper="Agnolucci_WACV_2024",
	term="gaublur",
	severity=3,
	mode="canonical",
	calibration=calibration,
	canonical_strengths=canonical_strengths,
	)
	
	to_pil_image(y).save("gaublur_canonical_k3.png")
\end{verbatim}

In this mode, backend-native severity levels remain unchanged.
Canonical severity indices are mapped to native parameters via
nearest-neighbor matching in degradation-strength space, ensuring
backend-agnostic and interpretable severity semantics.

\phantomsection
\section*{The Use of Large Language Models (LLMs)}
Large Language Models (LLMs) were used solely for writing refinement, coding assistance, literature search, and summarization. All conceptual development, experimental design, implementation, analysis, and initial writing of the paper were carried out by the authors.

\clearpage

\end{document}